\documentclass{ieeeaccess}
\usepackage{cite}
\usepackage{amsmath,amssymb,amsfonts}
\usepackage{algorithmic}
\usepackage{graphicx}
\usepackage{textcomp}
\usepackage{bbm}
\usepackage{multirow}
\usepackage{booktabs}
\usepackage{gensymb}
\usepackage{pifont}

\newcommand{\cmark}{\ding{51}}
\newcommand{\xmark}{\ding{55}}
\DeclareMathOperator*{\median}{median}

\def\BibTeX{{\rm B\kern-.05em{\sc i\kern-.025em b}\kern-.08em
    T\kern-.1667em\lower.7ex\hbox{E}\kern-.125emX}}
\begin{document}
\history{Date of publication xxxx 00, 0000, date of current version xxxx 00, 0000.}
\doi{}

\title{A Study on Self-Supervised Pretraining for Vision Problems in Gastrointestinal Endoscopy}
\author{\uppercase{Edward Sanderson and Bogdan J. Matuszewski}.
\IEEEmembership{Member, IEEE}}
\address{Computer Vision and Machine Learning (CVML) Group, University of Central Lancashire, Preston, UK}
\tfootnote{This work was supported by the Science and Technology Facilities Council [grant number ST/S005404/1]
}

\markboth
{Sanderson \headeretal: A Study on Self-Supervised Pretraining for Vision Problems in Gastrointestinal Endoscopy}
{Sanderson \headeretal: A Study on Self-Supervised Pretraining for Vision Problems in Gastrointestinal Endoscopy}

\corresp{Corresponding author: Edward Sanderson (e-mail: esanderson4@uclan.ac.uk).}

\begin{abstract}
Solutions to vision tasks in gastrointestinal endoscopy (GIE) conventionally use image encoders pretrained in a supervised manner with ImageNet-1k as backbones. However, the use of modern self-supervised pretraining algorithms and a recent dataset of 100k unlabelled GIE images (Hyperkvasir-unlabelled) may allow for improvements. In this work, we study the fine-tuned performance of models with ResNet50 and ViT-B backbones pretrained in self-supervised and supervised manners with ImageNet-1k and Hyperkvasir-unlabelled (self-supervised only) in a range of GIE vision tasks. In addition to identifying the most suitable pretraining pipeline and backbone architecture for each task, out of those considered, our results suggest three general principles. Firstly, that self-supervised pretraining generally produces more suitable backbones for GIE vision tasks than supervised pretraining. Secondly, that self-supervised pretraining with ImageNet-1k is typically more suitable than pretraining with Hyperkvasir-unlabelled, with the notable exception of monocular depth estimation in colonoscopy. Thirdly, that ViT-Bs are more suitable in polyp segmentation and monocular depth estimation in colonoscopy, ResNet50s are more suitable in polyp detection, and both architectures perform similarly in anatomical landmark recognition and pathological finding characterisation. We hope this work draws attention to the complexity of pretraining for GIE vision tasks, informs this development of more suitable approaches than the convention, and inspires further research on this topic to help advance this development. Code available: \underline{github.com/ESandML/SSL4GIE}
\end{abstract}


\titlepgskip=-15pt

\maketitle

\section{Introduction}
\label{sec:introduction}
\PARstart{G}{astrointestinal} endoscopy (GIE) is a procedure for screening and treating various digestive disorders that involves the insertion of a thin, flexible tube with a camera and light at the end, known as an endoscope, into either the mouth (gastroscopy) or anus (colonoscopy or sigmoidoscopy) of the patient. The endoscope is then traversed through the gastrointestinal tract as it transmits images of the inner lining to a monitor, where the endoscopist can inspect them for abnormalities and perform any necessary interventions. However, this poses several challenges for the endoscopist, such as the high volume and complexity of visual information, the variability and subtlety of the lesions, and the need for real-time decision making\cite{endo_challenges}.

To help overcome these challenges, computer vision has been identified as offering a promising set of tools for assisting endoscopists with various aspects of data analysis. Such aspects may be framed as traditional computer vision tasks such as image classification, object detection, semantic segmentation, and monocular depth estimation, among others, where the current state-of-the-art solutions for these tasks use deep learning models trained on large amounts of data.

\subsection{Related work}
While large datasets suitable for training models to perform image classification with \textit{everyday} images exist; most notably the publicly available ImageNet-1k\cite{imagenet}, but also the privately held JFT-300M \cite{jft300m1,jft300m2} and JFT-3B\cite{jft3b}; the datasets available for other computer vision tasks and distributions of images, particularly GIE images\cite{giedatasets}, are notably smaller. It has become clear that the amount of data a model is trained on has a strong influence on its performance\cite{unreasonable}, and efforts have therefore been taken to identify ways in which the largest available datasets can be leveraged in the training of models for tasks which these large datasets do not include suitable annotations for, and which may involve images of a dissimilar distribution. A now well-established approach\cite{imagenettransfer} is to train (\textit{pretrain}) an image classifier from random initialisation with the ImageNet-1k dataset (1.2M everday images), remove the classification layer and add any decoder components required for the intended (\textit{downstream}) task to the then pretrained image encoder, and train (\textit{fine-tune}) the resulting model with a  dataset which does include suitable annotations for the downstream task. Encoders used in this manner are often referred to as \textit{backbones}.

The approach of pretraining backbones on image classification with ImageNet-1k may however be limiting for two main reasons. Firstly, the model will learn to make high-level abstractions during pretraining, and since this pretraining is task-specific, these abstractions may not generalise well and may need to be \textit{unlearned} during fine-tuning. For example, the ground truth class of many images in ImageNet-1k refers to objects in the foreground and training a model to classify images on this basis may lead to the model learning to pay less attention to the background, which could contain information that is useful for the downstream task. Secondly, image classification datasets require annotations which can be expensive to produce, limiting the degree to which we can leverage more data in pretraining. This is particularly true of GIE images\cite{giedatasets}, which are especially expensive to annotate, and the use of which in pretraining may be beneficial when the downstream task involves such images.

With the aim of addressing these limitations, a significant amount of research into \textit{self-supervised} pretraining has been undertaken in recent years, leading to a range of popular algorithms\cite{simclr,dino,dinov2,mae,barlow,mocov3,beit,vicreg,swav,byol,simsiam}. Self-supervised pretraining algorithms set \textit{task-agnostic} objectives that require models to predict targets extracted from the input data, which can allow for the learning of \textit{generalisable} high-level feature recognition. Additionally, since this paradigm of learning does not require annotations, it provides the potential for leveraging a much larger amount of data and/or data of a more similar distribution to that involved in the downstream task.

A significant amount of research into self-supervised pretraining with everyday images\cite{simclr,dino,dinov2,mae,barlow,mocov3,beit,vicreg,swav,byol,simsiam}, as well as several modalities of medical images\cite{medssl1,medssl2,medssl3,medssl4,medssl5,medssl6,medssl7,medssl8}, has now been undertaken. However, it is still the convention in GIE to employ backbones that have been pretrained in a supervised manner with ImageNet-1k. A set of 99,417 unlabelled GIE images (Hyperkvasir-unlabelled) was however included in the recently released Hyperkvasir dataset\cite{hyperkvasir} which, while much smaller than ImageNet-1k, is significantly larger than other datasets of GIE images. This data should allow for the self-supervised pretraining of GIE-specific backbones, which may be better suited to some tasks in GIE than the described convention. Additionally, self-supervised pretraining with datasets of everyday images, e.g. ImageNet-1k, may also provide opportunities for improvements.

\subsection{Contributions}

This paper presents a study on pretraining encoders for use as backbones in solutions to vision tasks in GIE. We consider twelve encoders, each of a ResNet50\cite{resnet} or ViT-B\cite{vit} architecture and pretrained with one of six pipelines, including two self-supervised pretraining algorithms per architecture, each used separately with both ImageNet-1k and Hyperkvasir-unlabelled, as well as baselines of supervised pretraining with ImageNet-1k and random initialisation (not pretrained). We use state-of-the-art methods for adapting and fine-tuning each encoder for a range of vision tasks in GIE, namely: anatomical landmark recognition, pathological finding characterisation, polyp detection, polyp segmentation, and monocular depth estimation in colonoscopy; and we compare the resulting models on the basis of their fine-tuned performance using well-established metrics. The overall workflow of our experimentation is illustrated in Fig. \ref{fig:pipeline}.

\Figure[t!](topskip=0pt, botskip=0pt, midskip=0pt){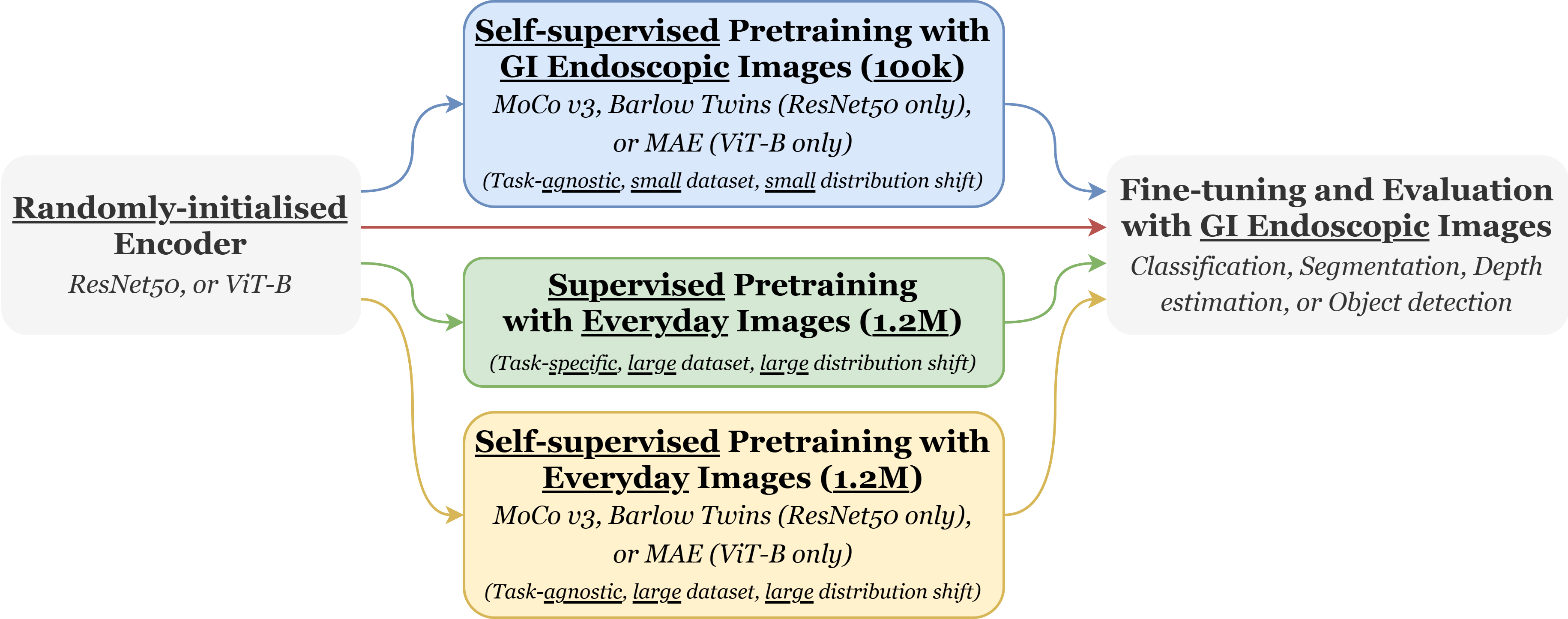}
{The overall workflow of our experimentation.\label{fig:pipeline}}

In addition to identifying which architecture and pretraining pipeline (algorithm and data) is most suitable for each task, our results suggest that self-supervised pretraining with ImageNet-1k consistently allows for better performance than supervised pretraining with ImageNet-1k, across all considered tasks and architectures. We also demonstrate that self-supervised pretraining with ImageNet-1k is typically more suitable than self-supervised pretraining with Hyperkvasir-unlabelled, with the notable exception of monocular depth estimation in colonoscopy where the similarity of the pretraining data to the downstream data appears to be more critical than the amount of pretraining data. Additionally, we find that ViT-B backbones are typically more suitable for polyp segmentation and monocular depth estimation in colonoscopy, that ResNet50 backbones are more suitable for polyp detection, and that both architectures perform similarly in anatomical landmark recognition and pathological finding characterisation.

While a number of studies have experimented with self-supervised pretraining for certain GIE vision tasks before\cite{ssl4gie1,ssl4gie2,ssl4gie3,ssl4gie4,ssl4gie5}, only two\cite{ssl4gie4,ssl4gie5} have compared self-supervised pretraining against the convention of supervised pretraining with ImageNet-1k. Additionally, in their experiments with GIE vision tasks, these works either compared self-supervised pretraining against supervised pretraining of a different architecture with the same dataset, or the same architecture with a different dataset. Our work is therefore the first to compare self-supervised pretraining against supervised pretraining for the same encoder architecture and pretraining data, in terms of fine-tuned performance on GIE vision tasks. Additionally, we consider a much wider scope of self-supervised pretraining algorithms and GIE vision tasks than these previous works, each of which focuses on a single task, and are the first that we know of to experiment with self-supervised pretraining for polyp detection and monocular depth estimation in colonoscopy. Beyond the value of these results in isolation, this wide scope allows us to expose the general principles revealed by our analysis.

\section{Investigated self-supervised pretraining algorithms}
Self-supervised algorithms for pretraining image encoders for use as backbones can be grouped into four families\cite{cookbook}:
\begin{itemize}
    \item \textit{\textbf{Deep metric learning (DML)}}-based self-supervised pretraining algorithms train an encoder to describe \textit{semantically} similar images with \textit{quantifiably} similar representations, and semantically dissimilar images with quantifiably dissimilar representations. This is typically achieved by creating \textit{positive} pairs, which are distorted variants of the same image, and \textit{negative} pairs, which are distorted variants of different images, and training the encoder with a \textit{contrastive} loss that is minimised through a reduction in the distance or angle between the representations of positive pairs, and an increase in the distance or angle between the representations of negative pairs.
    \item \textit{\textbf{Self-distillation}}-based self-supervised pretraining algorithms train an encoder to describe a variant of an image with a representation that allows for a representation of a different variant of the image, produced by another encoder, to be predicted. As a means of avoiding \textit{collapse}, which occurs when both encoders learn to output the same representation for all images, the second encoder is typically an \textit{exponential moving average} of the encoder being optimised, though collapse can be avoided through a Siamese network with a stop-gradient on one branch \cite{simsiam}.
    \item \textit{\textbf{Canonical correlation analysis (CCA)}}-based self-supervised pretraining algorithms train an encoder to describe an image in such a way that each feature of its representation is informative of a \textit{distinct} attribute of the image. This is typically achieved with a loss function that encourages the encoder to maintain a certain amount of \textit{variance} for each feature in the representation, while establishing \textit{uncorrelatedness} between features.
    \item \textit{\textbf{Masked image modelling (MIM)}}-based self-supervised pretraining algorithms aim to reproduce the success of masked language modelling (MLM) pretraining algorithms, first introduced for pretraining the transformer-based text encoder BERT\cite{bert}, in the domain of vision. MIM algorithms are therefore typically used with ViT architectures, which are also inspired by BERT, where the image is split into patches that are treated as a sequence of \textit{visual} tokens akin to the sequence of word tokens used to represent input text for BERT. In both MLM and MIM, input tokens are randomly \textit{masked}, and a model is trained to \textit{reconstruct} these tokens based on the information contained in the remaining tokens.
\end{itemize}

The rest of this section presents the selection of algorithms considered in our experimentation, which we ensured spanned these four families of self-supervised algorithms. We illustrate and provide a definition of the key details of each algorithm, where we use $f_\theta$ to denote the image encoder being optimised for use as a backbone, and explain how we obtained and used encoders pretrained using each algorithm with either ImageNet-1k or Hyperkvasir-unlabelled. Note that any training performed as part of this work was done on an ASUS ESC8000-G4 GPU server with 6$\times$ NVIDIA RTX A6000 48GB GPUs. Due to the number of GPUs and amount of memory, the batch sizes used are in multiples of 6 and, for pretraining, are the maximum that we could allow for.

\subsection{MoCo v3}

\Figure[t!](topskip=0pt, botskip=0pt, midskip=0pt){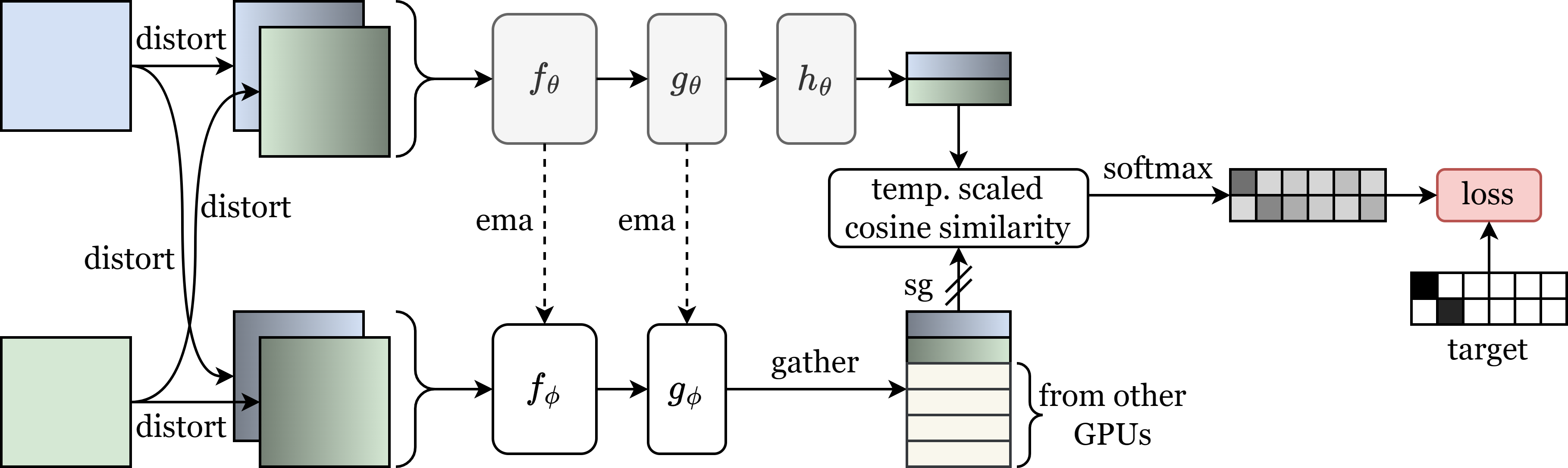}
{Visualisation of the MoCo v3 algorithm. Shown for a per-GPU batch size of 2, and 3 GPUs. We use $g_\theta$ to denote the projector, $h_\theta$ to denote the predictor, $\phi$ to denote the momentum parameters that are computed with an exponential moving average (denoted $\mathrm{ema}$) of the online parameters $\theta$, and $\mathrm{sg}$ is a stop-gradient.\label{fig:mocov3}}

MoCo v3\cite{mocov3}, illustrated in Fig. \ref{fig:mocov3}, is the latest iteration of the \textit{momentum contrast (MoCo)} algorithm, which started as an example of DML. While the distinguishing feature of all iterations of MoCo is the momentum encoder and projector $g_\phi\circ f_\phi$, which is used to compute a representation for one image variant in each pair, rather than using the online encoder and projector $g_\theta\circ f_\theta$ to compute both representations as is more conventional in DML, e.g. SimCLR\cite{simclr}, MoCo v3 incorporates a prediction head $h_\theta$. The resulting algorithm can be framed as either a DML algorithm that incorporates the principle of self-distillation, or a self-distillation algorithm which uses a contrastive loss. As such, we consider MoCo v3 as a representative of both the DML and self-distillation families.

We define a batch of positive pairs of image variants on a single GPU as $\{(\mathbf{x}_{i,1},\mathbf{x}_{i,2})\}_{i=1}^{N_b}$. We then define the representations used by MoCo v3 as:

\begin{equation}    \mathbf{q}_{i,j}=h_\theta\left(g_\theta\left(f_\theta\left(\mathbf{x}_{i,j}\right)\right)\right), \quad i=1,\ldots,N_b \quad \text{and} \quad j=1,2
\end{equation}

\begin{equation}    \mathbf{k}_{i,j}=g_\phi\left(f_\phi\left(\mathbf{x}_{i,j}\right)\right), \quad i=1,\ldots,N_{Gb} \quad \text{and} \quad j=1,2
\end{equation}

\noindent where $N_{Gb}=N_GN_b$, where $N_G$ is the number of GPUs, and the representations $\{\mathbf{k}_{i,j}\}_{i=N_b+1,j=1}^{N_{Gb},2}$ are gathered from the other GPUs (see Fig. \ref{fig:mocov3}), where they are computed in the same manner as $\{\mathbf{k}_{i,j}\}_{i=1,j=1}^{N_b,2}$ on different image variants, i.e. $\{(\mathbf{x}_{i,1},\mathbf{x}_{i,2})\}_{i=N_b+1}^{N_{Gb}}$. The loss function used by MoCo v3 for a batch on a single GPU can then be defined:

\begin{multline}
    \mathcal{L}_{MC3}\left(\{\mathbf{q}_{i,1}\}_{i=1}^{N_b},\{\mathbf{k}_{i,1}\}_{i=1}^{N_{Gb}},\{\mathbf{q}_{i,2}\}_{i=1}^{N_b},\{\mathbf{k}_{i,2}\}_{i=1}^{N_{Gb}}\right)=\\\frac{2\tau}{N_b}\sum_{i=1}^{N_b}\biggl[\mathcal{L}_{INCE}\left(\mathbf{q}_{i,1},\{\mathbf{k}_{j,2}\}_{j=1}^{N_{Gb}}\right)+\\ \mathcal{L}_{INCE}\left(\mathbf{q}_{i,2},\{\mathbf{k}_{j,1}\}_{j=1}^{N_{Gb}} \right) \biggr]\label{moco}
\end{multline}

\noindent where $\tau$ is the temperature parameter, a constant positive scalar, and $\mathcal{L}_{INCE}$ is the InfoNCE loss\cite{infonce}, which is defined:

\begin{equation}
    \mathcal{L}_{INCE}(\mathbf{q}_i,\{\mathbf{k}_j\}_{j=1}^N)=-\log \left(\frac{e^{\mathrm{CoSim}(\mathbf{q}_i,\mathbf{k}_i)/\tau}}{\sum_{j=1}^{N} e^{\mathrm{CoSim}(\mathbf{q}_i,\mathbf{k}_j)/\tau}}\right )\label{infonce}
\end{equation}

\noindent where $\mathrm{CoSim}$ is the cosine similarity:

\begin{equation}
    \mathrm{CoSim}(\mathbf{a},\mathbf{b})=\frac{\mathbf{a}^\top \mathbf{b}}{\Vert \mathbf{a}\Vert \Vert \mathbf{b} \Vert}
\end{equation}

Note that Fig. \ref{fig:mocov3} can be seen as illustrating

\begin{equation}
\frac{2\tau}{N_b} \sum_{i=1}^{N_b}\mathcal{L}_{INCE}\left(\mathbf{q}_{i,1},\{\mathbf{k}_{j,2}\}_{j=1}^{N_GN_b}\right)
\end{equation}

\noindent whereas $\mathcal{L}_{MC3}$ makes this symmetrical.

The algorithm has been designed to work effectively for optimising both ResNet and ViT architectures of $f_\theta$. For ViT architectures, the patch embedding layer is frozen as a random linear projection for stability reasons, and the \textit{class token} \texttt{[cls]} is taken as the output of $f_\theta$. The projectors $g_\theta$ and $g_\phi$, and the predictor $h_\theta$, are defined as multilayer perceptrons (MLPs) composed of fully connected layers, batch normalisation and ReLU activations.

We consider the use of MoCo v3 for pretraining both ResNet50 and ViT-B architectures, and use the \texttt{torchvision} implementation of ResNet50 and the ViT-B implementation from the official MoCo v3 codebase. For the encoders pretrained using MoCo v3 with ImageNet-1k, we use the weights provided by the authors. We then used the implementation of MoCo v3 in the official codebase to pretrain encoders with Hyperkvasir-unlabelled, modifying the code only for loading Hyperkvasir-unlabelled and to change the batch size from 4096 to 1536/768 (ResNet50/ViT-B). When fine-tuning the ViT-B models, we unfreeze the patch embedding layer.

\subsection{Barlow Twins}
\Figure[t!](topskip=0pt, botskip=0pt, midskip=0pt){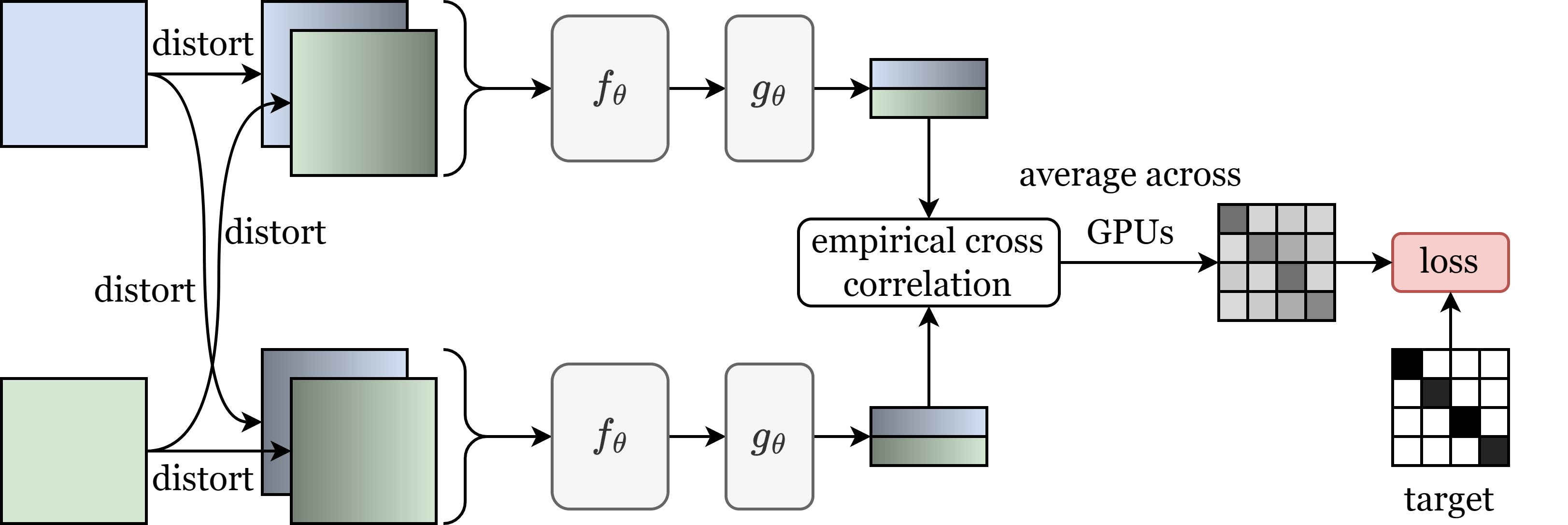}
{Visualisation of the Barlow Twins algorithm. Shown for a per-GPU batch size of 2, and representations of dimensionality 4. We use $g_\theta$ to denote the projector.\label{fig:barlow}}
Barlow Twins\cite{barlow}, illustrated in Fig. \ref{fig:barlow}, is an example of a CCA algorithm. Barlow Twins trains a model to maintain a certain amount of variance for each feature and to establish uncorrelatedness between features with a loss function that encourages an \textit{identity} empirical cross-correlation matrix between representations of two distorted variants of the same image. Other examples of CCA differ mainly in the loss function. For example, the loss function used by VicReg\cite{vicreg} encourages the variance of features to be maintained, and uncorrelatedness between features to be established, on representations of \textit{individual} variants of an image directly, as well as minimising the Euclidean distance between representations of \textit{two} variants of the same image.

We define a batch of positive pairs of image variants on a single GPU as $\{(\mathbf{x}_{i,1},\mathbf{x}_{i,2})\}_{i=1}^{N_b}$. We then define the representations used by Barlow Twins as:

\begin{equation}    \mathbf{z}_{i,j}=g_\theta\left(f_\theta\left(\mathbf{x}_{i,j}\right)\right), \quad i=1,\ldots,N_b \quad \text{and} \quad j=1,2
\end{equation}

\noindent which may also be written as $\left(z_{i,j,k}\right)_{k=1}^d=\mathbf{z}_{i,j}$. These representations are normalised to give:

\begin{equation}
    \hat{z}_{i,j,k}=\frac{z_{i,j,k}-\frac{1}{N_b}\sum_{m=1}^{N_b}z_{m,j,k}}{\sqrt{\frac{1}{N_b}\sum_{n=1}^{N_b}\left(z_{n,j,k}-\frac{1}{N_b}\sum_{m=1}^{N_b}z_{m,j,k}\right)^2}}
\end{equation}

The elements of the empirical cross correlation matrix $(c_{k,l})_{k=1,l=1}^{d,d}$ can then be defined:

\begin{equation}
    c_{k,l}=\frac{1}{N_b}\sum_{i=1}^{N_b}\hat{z}_{i,1,k}\hat{z}_{i,2,l}
\end{equation}

\noindent which is averaged across GPUs, the result of which we denote $(\Bar{c}_{k,l})_{k=1,l=1}^{d,d}$. Finally, the Barlow Twins loss can be defined:

\begin{equation}
\mathcal{L}_{BT}\left((\bar{c}_{k,l})_{k=1,l=1}^{d,d}\right)=\sum_{k=1}^{d}(1-\bar{c}_{k,k})^2 + \lambda \sum_{k=1}^{d}\sum_{l=1}^{d}\mathbbm{1}_{[k\neq l]}\bar{c}_{k,l}^2
\end{equation}

\noindent where $\lambda$ is a constant positive scalar and $\mathbbm{1}$ is an indicator function.

The algorithm has been designed to work effectively for ResNet architectures of $f_\theta$. The projector $g_\theta$ is defined as an MLP composed of fully connected layers, batch normalisation and ReLU activations.

We consider the use of Barlow Twins for pretraining ResNet50 architectures, for which we use the \texttt{torchvision} implementation. For the ResNet50 pretrained using Barlow Twins with ImageNet-1k, we use the weights provided by the authors. We then used the implementation of Barlow Twins in the official codebase to pretrain a ResNet50 with Hyperkvasir-unlabelled, modifying the code only for loading Hyperkvasir-unlabelled and to change the batch size from 2048 to 1536.

\subsection{MAE}
Masked autoencoders (MAE)\cite{mae}, illustrated in Fig. \ref{fig:mae}, is a particularly popular example of the MIM family. It differs from other popular MIM algorithms on two main fronts. First, examples such as BEiT\cite{beit}, PeCo\cite{peco}, and SimMIM\cite{simmim} use an \textit{arbitrary} token in place of the masked tokens in the input to $f_\theta$, where MAE simply omits them. Notably, this is only possible with ViTs due to the use of position embeddings that inform a model of the specific patch of an image that a token corresponds to \textit{explicitly}. For reconstruction, this does however require the insertion of an arbitrary token at each position of a masked token in the output of $f_\theta$, and the processing of the resulting sequence of tokens by a decoder $g_\theta$, which has a smaller ViT architecture. Secondly, BEiT and PeCo use the discrete variational autoencoder introduced as a component of DALL-E\cite{dalle} to quantise all possible image patches into a finite set of visual tokens akin to a \textit{vocabulary} of words, rather than directly using the patches as visual tokens. This allows the reconstruction to be framed as classifying which token in this finite set the masked token should be, closely following BERT. MAE however takes a more conventional approach to image reconstruction and frames it as a regression problem.

\Figure[t!](topskip=0pt, botskip=0pt, midskip=0pt){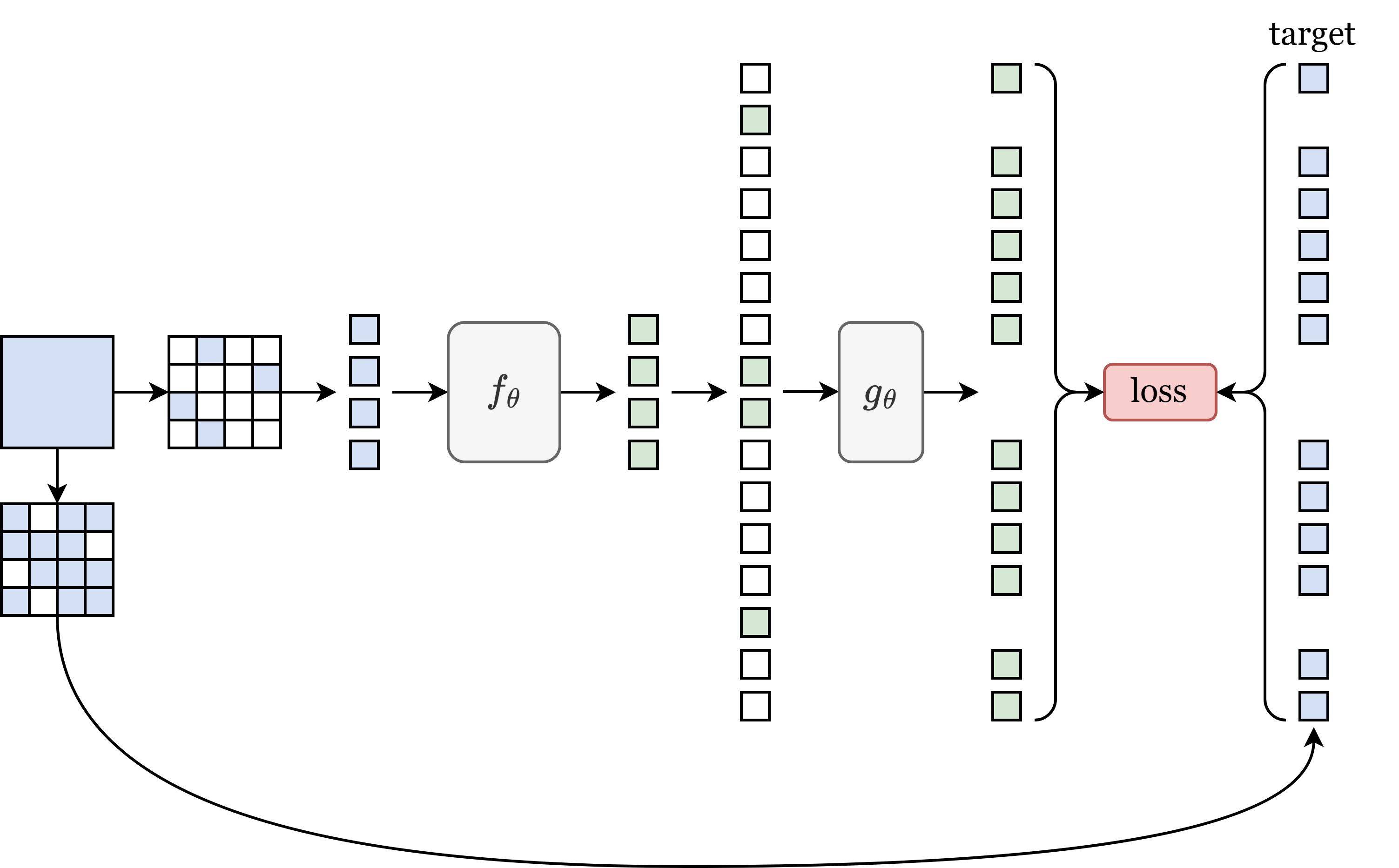}
{Visualisation of the MAE algorithm. Shown for a ViT encoder that treats an image as a $4\times4$ grid of patch tokens, with 75\% masking.\label{fig:mae}}

As is typical for a ViT, an image is first divided into a sequence of flattened non-overlapping patches that are projected by a patch embedding layer and translated by a position embedding to produce the sequence of visual tokens $(\mathbf{x}_i)_{i=1}^{N_p}$ that are to be concatenated with the \texttt{[cls]} token and fed into the first block. Before concatenating with the \texttt{[cls]} token however, MAE generates a set of uniformly distributed random values $\{\alpha_i \sim \mathcal{U}(0,1)\}_{i=1}^{N_p}$ and computes the permutation $\sigma$ which sorts the set into reverse order, i.e. $\alpha_{\sigma(i)} \ge \alpha_{\sigma(i+1)}$ for $i=1,\ldots,N_p-1$. For a proportion of masking $\gamma\in[0,1]$, selected to ensure that $\gamma N_p - \lfloor \gamma N_p\rfloor=0$, the sequence passed forward is then $(\tilde{\mathbf{x}}_{i})_{i=1}^{(1-\gamma)N_p}=(\mathbf{x}_{\sigma(i)})_{i=1}^{(1-\gamma)N_p}$. In contrast to MIM algorithms that replace rather than omit the masked tokens from the input to $f_\theta$, it is important in MAE that the same number of tokens in each input are masked, i.e. $\gamma N_p$ is constant, to allow for batching. If the sequence of visual tokens, i.e. omitting the \texttt{[cls]} token, in the output of $f_\theta$ is denoted $(\tilde{\mathbf{z}}_{i})_{i=1}^{(1-\gamma)N_p}$, we then create the sequence $(\mathbf{z}_{i})_{i=1}^{N_p}$, where:

 \begin{equation}
     \mathbf{z}_{i}=
         \begin{cases}
        \tilde{\mathbf{z}}_{\sigma^{-1}(i)} & \text{if } 1\le \sigma^{-1}(i)\le(1-\gamma)N_p\\
        \mathbf{m} & \text{if } (1-\gamma)N_p+1 \le \sigma^{-1}(i) \le N_p
    \end{cases}
 \end{equation}

\noindent where $\mathbf{m}$ is a learnt arbitrary token. The tokens in $(\mathbf{z}_{i})_{i=1}^{N_p}$ are then translated by another position embedding and fed through the decoder blocks with the \texttt{[cls]} token. The output of the decoder blocks is then fed through a prediction head and the \texttt{[cls]} token is removed, leaving the sequence of reconstructed flattened patches for the entire image $(\hat{\mathbf{y}}_{i})_{i=1}^{N_p}$. Denoting the sequence of ground truth flattened patches $(\mathbf{y}_{i})_{i=1}^{N_p}$, in which the features have been zero-centred and scaled to unit variance for each patch independently, the loss function is defined:

\begin{multline}
\mathcal{L}_{MAE}\left((\hat{\mathbf{y}}_{i})_{i=1}^{N_p}\right)=\\ \frac{1}{\gamma N_p d_p}\sum_{i=1}^{N_p}\mathbbm{1}_{[(1-\gamma)N_p+1 \le \sigma^{-1}(i) \le N_p]}\Vert \hat{\mathbf{y}}_i-\mathbf{y}_i\Vert^2
\end{multline}

\noindent where $d_p$ is the dimensionality of a patch. The loss is then averaged over all images in the batch on a single GPU, and the update to the model is averaged over GPUs, as is typical in distributed supervised learning.

As mentioned, MAE has been designed for pretraining ViT architectures specifically. A notable distinction between the use of ViT in MAE and in MoCo v3 is that the loss is computed on the processed visual tokens in MAE, whereas it is computed on the processed \texttt{[cls]} token in MoCo v3.

We consider the use of MAE for pretraining ViT-B architectures, for which we use the implementation from the official MAE codebase. For the ViT-B pretrained using MAE with ImageNet-1k, we use the weights provided by the authors. We then used the implementation of MAE in the official codebase to pretrain a ViT-B with Hyperkvasir-unlabelled, modifying the code only for loading Hyperkvasir-unlabelled and to change the batch size from 4096 to 768.

\section{Baselines}
For each of the considered encoder architectures, ResNet50 and ViT-B, we consider two baselines to compare the discussed self-supervised pretraining pipelines against. Most importantly, we consider supervised pretraining with ImageNet-1k, representing the conventional approach for pretraining image encoders for use as backbones in solutions to GIE vision tasks. We then consider no pretraining, i.e. finetuning from random initialisation. We use the \texttt{torchvision} implementation and weights for ResNet50, and the \texttt{timm} implementation and weights for ViT-B.

We note that we do not directly compare against the state-of-the-art methods for each task. While our primary aim is to study the relative effectiveness of different pretraining pipelines, which such comparisons would not be suitable for due to the need for consistency in all other details, we believe that this would still be informative. However, we cannot compare against previously reported results due to the lack of standardisation in the benchmarks, with different works using different splits and different evaluation methodologies, and re-implementing these methods to allow for a direct comparison would be too time-consuming. To the best of our knowledge, the state-of-the-art for each task uses either a convolutional neural network or some derivative of ViT that has been pretrained in a supervised manner with ImageNet-1k as a backbone, and as such we consider models with a ResNet50 or ViT-B backbone that has been pretrained in a supervised manner with ImageNet-1k as representative of the state-of-the-art.

\section{Image classification}
Image classification is the problem of determining which, out of a predefined set of classes, a given image should be assigned to. In the context of GIE, the predefined set of classes may cover, for example, possible anatomical landmarks, pathological findings, or categories of polyps. In this section, we detail and present our evaluation of the fine-tuned performance of backbones in two of these image classification tasks, namely anatomical landmark recognition and pathological finding characterisation.

\subsection{Data}
The data used in our image classification experiments is taken from the Hyperkvasir-labelled dataset\cite{hyperkvasir}, which does not share any instances with Hyperkvasir-unlabelled. We specifically used the anatomical landmarks and pathological findings subsets, which we treated the classification of as two separate problems. For each subset, we combined the data for the upper and lower gastrointestinal tract, and applied a random 80\%/10\%/10\% training/validation/test split, where the validation data is used to determine whether to save the weights after each epoch of training on the training data, and the test data is reserved for evaluating the model after fine-tuning. The number of instances of each class, in total and in each split, are given in Table \ref{tab:class splits}.

\begin{table*}[ht]
\caption{\label{tab:class splits}Number of instances of each class, in total and in each split.}
\centering
\begin{tabular}{ccccccc}
\toprule
Classification Task           & Tract & Class & Total & Train. & Val. & Test. \\ \hline
\multirow{6}{*}{Anatomical landmark recognition} & \multirow{3}{*}{Lower} & Cecum           & 1009  &  794 &  109 &  106\\
                          &                                       & Ileum      & 9 & 6  & 2  & 1   \\
                          &                                       & Retroflex-rectum      & 391 & 308  & 32  & 51                                                      \\ \cmidrule{2-7} 
                          & \multirow{3}{*}{Upper}          & Pylorus           & 999  &  816 &  87 &  96\\
                          &                                       & Retroflex-stomach      & 764 & 610  & 82  & 72   \\
                          &                                       & Z-line      & 932 & 750  & 98  & 84                                                                        \\ \cmidrule{1-7}
\multirow{12}{*}{Pathological finding characterisation} & \multirow{8}{*}{Lower} & Hemorrhoids           & 6  &  3 &  3 &  0\\
                          &                                       & Polyps      & 1028 & 825  & 100  & 103   \\
                          &                                       & Ulcerative colitis grade 0-1      & 35 & 28  & 4  & 3                                                      \\ &                                      & Ulcerative colitis grade 1      & 201 & 160  & 17  & 24                                                      \\ &                                       & Ulcerative colitis grade 1-2      & 11 & 7  & 2  & 2                                                      \\  &                                      & Ulcerative colitis grade 2      & 443 & 347  & 53  & 43                                                      \\  &                                      & Ulcerative colitis grade 2-3      & 28 & 23  & 2  & 3                                                      \\   &                                     & Ulcerative colitis grade 3      & 133 & 109  & 13  & 11                                                      \\ \cmidrule{2-7} 
                          & \multirow{4}{*}{Upper}          & Barrett's short-segment          & 53  &  44 &  5 &  4\\
                          &                                       & Barrett's      & 41 & 32  & 4  & 5   \\
                          &                                       & Esophagitis A      & 403 & 333  & 33  & 37   \\
                          &                                       & Esophagitis B-D      & 260 & 203  & 28  & 29                                                                        \\ \bottomrule
\end{tabular}
\end{table*}

\subsection{Decoders}
In image classification, it is typical to simply add a linear classifier to the final representation computed by an encoder to allow for prediction. Following convention, we implement this as a fully connected layer that maps the final representation to a vector of logits, one for each possible class, which is softmax normalised prior to computation of the loss. For the ViT-B models, we use the output \texttt{[cls]} token as the final representation.

\subsection{Fine-tuning procedure}

We separately train each model to perform both anatomical landmark recognition and pathological finding characterisation through the same procedure. We use the common fine-tuning procedure hyperparameters given in Table \ref{tab:common} and pre-process the training images using the pipeline detailed in Table \ref{tab:preprocessing}. The loss is then computed using a cross entropy loss function which, due to the significant class imbalance in the data, is weighted with a value of $N_D/N_iN_c$ for the $i^{th}$ class, where $N_D$ is the total number of images in the dataset, $N_i$ is the number of images in a particular class, and $N_c$ is the number of classes. Note that these numbers are for the entire dataset, rather than the training set. This weighting ensures that the total sum of weights across all instances is $N_D$, for consistency with unweighted cross entropy. We use the macro F1-score (mF1)\footnote{Two different formulations of mF1 can be found in the literature --- \eqref{mF1} is the more robust\cite{macrovmacro} arithmetic mean of individual F1 scores.} as the validation metric:

\begin{equation}
    \text{mF1} = \frac{1}{N_c}\sum_{i=1}^{N_c}\frac{2\mathrm{TP}_i + \epsilon}{2\mathrm{TP}_i+\mathrm{FP}_i+\mathrm{FN}_i+\epsilon}\label{mF1}
\end{equation}

\noindent where $\mathrm{TP}_i$ is the number of true positives for the $i^{th}$ class, $\mathrm{FP}_i$ is the number of false positives, $\mathrm{FN}_i$ is the number of false negatives, and $\epsilon=1e-8$. The transformations applied to the validation images include the same resizing and normalisation applied to the training images. Finally, the model is trained on this basis for 50 epochs, with the parameters saved after each epoch that leads to an improvement in mF1 on the validation set, with any batch normalisation synchronised across GPUs.

\begin{table*}[ht]
\caption{\label{tab:preprocessing}Pre-processing of training images, which is performed online during training. 1) pads to $\mathrm{max}(h,w)\times\mathrm{max}(h,w)$ for original height $h$ and width $w$. 2) resizes to $224\times224$ using bicubic interpolation with anti-aliasing\cite{antialiasing}. 3) applies colour jitter with brightness factor sampled uniformly from $[0.6,0.4]$, contrast factor sampled uniformly from $[0.5, 1.5]$, saturation factor sampled uniformly from $[0.75, 1.25]$, and hue factor sampled uniformly from $[0.99, 1.01]$. 4) applies Gaussian blur with a 25 × 25 kernel with a standard deviation sampled uniformly from $[0.001, 2]$. 5) applies a rotation of 90\degree with a probability of 0.5. 6) applies a horizontal flip with a probability of 0.5. 7) applies a vertical flip with a probability of 0.5. 8) applies a rotation of an angle sampled uniformly from $[-180\degree, 180\degree]$. 9) applies an affine transform with, horizontal translation sampled uniformly from $[-28, 28]$, vertical translation sampled uniformly from $[-28, 28]$, scaling with factor sampled uniformly from $[0.5,1.5]$, and shearing of an angle sampled uniformly from $[-22.5\degree, 22.5\degree]$. 10) applies normalisation using the ImageNet-1k pixel mean and standard deviation, for consistency with all pretraining pipelines.}
\centering
\begin{tabular}{ccccc}
\toprule
Operation            & Image classification                      & Object detection & Semantic segmentation & Monocular depth estimation                                                   \\ \hline 
1) Pad to square & \xmark & \xmark & \xmark & \cmark \\
2) Resize & \cmark & \xmark & \cmark & \cmark \\
3) Colour jitter & \cmark & \cmark & \cmark & \cmark \\
4) Gaussian blur & \cmark & \cmark & \cmark & \xmark \\
5) Discrete rotation & \xmark & \cmark & \xmark & \xmark \\
6) Horizontal flip  & \cmark & \cmark & \cmark & \cmark \\
7) Vertical flip  & \cmark & \cmark & \cmark & \cmark \\
8) Continuous rotation & \cmark & \xmark & \cmark & \xmark \\
9) Affine transform & \xmark & \xmark & \cmark & \xmark \\
10) Normalisation  & \cmark & \cmark & \cmark & \cmark
\\ \bottomrule
\end{tabular}
\end{table*}

\begin{table}[ht]
\caption{\label{tab:common}Common fine-tuning procedure hyperparameters.}
\centering
\begin{tabular}{cc}
\toprule
Hyperparameter & Value \\ \hline
Batch size & 48 \\ 
Optimiser & AdamW\cite{adamw} \\ 
\begin{tabular}{c}Initial learning\\rate\end{tabular} & 1e-4 \\ 
\begin{tabular}{c}Learning rate\\schedule\end{tabular} & \begin{tabular}{c}Halve when validation\\performance does not\\ improve over 10 epochs,\\until reaching 1e-6\end{tabular}
\\ \bottomrule
\end{tabular}
\end{table}

\subsection{Evaluation}
We evaluate the resulting image classification models using the corresponding test data, which is pre-processed in the same manner as the validation data, with four metrics, namely mF1 (as defined in \eqref{mF1}), mPrecision, mRecall, and Accuracy:

\begin{equation}
    \text{mPrecision}= \frac{1}{N_c}\sum_{i=1}^{N_c}\frac{\mathrm{TP}_i+\epsilon}{\mathrm{TP}_i+\mathrm{FP}_i+\epsilon}
\end{equation}

\begin{equation}
    \text{mRecall}= \frac{1}{N_c}\sum_{i=1}^{N_c}\frac{\mathrm{TP}_i+\epsilon}{\mathrm{TP}_i+\mathrm{FN}_i+\epsilon}
\end{equation}

\begin{equation}
    \text{Accuracy}=\frac{\sum_{i=1}^{N_c}\mathrm{TP}_i}{N_D}
\end{equation}

\noindent where $\epsilon=1e-8$.

For all metrics, a higher value indicates better performance. The results for anatomical landmark recognition are presented in Table \ref{tab:anatomical} and the results for pathological finding characterisation are presented in Table \ref{tab:pathological}.

\begin{table*}[ht]
\centering
\caption{\label{tab:anatomical}Performance in anatomical landmark recognition. The best results for each architecture are highlighted as bold, and the best results overall are underlined.}
\begin{tabular}{ccccccc}
\toprule
Backbone arch.            & Pretraining data                      & Pretraining algo. & mF1 & mPrecision & mRecall & Accuracy                                                   \\ \hline
\multirow{6}{*}{ResNet50} & \multirow{2}{*}{Hyperkvasir-unlabel.} & MoCo v3           & 0.823  &  0.989 &  0.823 &  0.988\\
                          &                                       & Barlow Twins      & 0.824 & 0.989  & 0.826  & 0.990                                                        \\ \cmidrule{2-7} 
                          & \multirow{3}{*}{ImageNet-1k}          & MoCo v3           & \textbf{\underline{0.828}}  & \textbf{\underline{0.993}}  &  \textbf{\underline{0.829}} &  \textbf{\underline{0.993}}                                                        \\ 
                          &                                       & Barlow Twins      & 0.826 &  0.991 & 0.828  &  0.990                                                        \\ \cmidrule{3-7}
                          &                                       & Supervised        & 0.826 &  0.827 &  0.825 & 0.988                                                        \\ \cmidrule{2-7} 
                          & None                                  & None              &  0.793  &  0.957 &  0.795 &  0.956                                                        \\ \cmidrule{1-7}
\multirow{6}{*}{ViT-B}    & \multirow{2}{*}{Hyperkvasir-unlabel.} & MoCo v3           & 0.818 & 0.983  &  0.819 & 0.983                                                        \\
                          &                                       & MAE               & 0.823  & 0.990  & 0.823  & 0.988 \\ \cmidrule{2-7} 
                          & \multirow{3}{*}{ImageNet-1k}          & MoCo v3           & \textbf{\underline{0.828}} &  \textbf{\underline{0.993}} & \textbf{\underline{0.829}}  &  \textbf{\underline{0.993}}                                                        \\
                          &                                       & MAE               & 0.823 &  0.989 & 0.823  &  0.988                                                        \\ \cmidrule{3-7}
                          &                                       & Supervised        & 0.815  &  0.979 &  0.817 & 0.980                                                        \\ \cmidrule{2-7} 
                          & None                                  & None              & 0.782 &  0.955 & 0.778  &  0.944                                                        \\ \bottomrule
\end{tabular}
\end{table*}

\begin{table*}[ht]
\caption{\label{tab:pathological}Performance in pathological finding characterisation. The best results for each architecture are highlighted as bold, and the best results overall are underlined.}
\centering
\begin{tabular}{ccccccc}
\toprule
Backbone arch.            & Pretraining data                      & Pretraining algo. & mF1 & mPrecision & mRecall & Accuracy                                                   \\ \hline
\multirow{6}{*}{ResNet50} & \multirow{2}{*}{Hyperkvasir-unlabel.} & MoCo v3           & 0.542 & 0.531  &  0.495 &  0.746\\
                          &                                       & Barlow Twins      & 0.613  &  0.527 &  \textbf{0.561} &  0.765                                                        \\ \cmidrule{2-7} 
                          & \multirow{3}{*}{ImageNet-1k}          & MoCo v3           & 0.595 &  0.515 & 0.516  &  0.758                                                        \\ 
                          &                                       & Barlow Twins      & \textbf{0.628}  &  \textbf{0.682} &  0.545 & 0.765                                                        \\ \cmidrule{3-7}
                          &                                       & Supervised        & 0.587 & 0.645  & 0.492  &  \textbf{0.777}                                                        \\ \cmidrule{2-7} 
                          & None                                  & None              &  0.491 &  0.584 &  0.437 &  0.621                                                        \\ \cmidrule{1-7}
\multirow{6}{*}{ViT-B}    & \multirow{2}{*}{Hyperkvasir-unlabel.} & MoCo v3           & 0.589 & 0.690  &  0.526 &  0.777                                                        \\
                          &                                       & MAE               & 0.576 &  0.563 & 0.526  &  0.723 \\ \cmidrule{2-7} 
                          & \multirow{3}{*}{ImageNet-1k}          & MoCo v3           & 0.596  & 0.608  &  0.527 &  0.731                                                        \\
                          &                                       & MAE               & \textbf{\underline{0.652}}  &  \textbf{\underline{0.723}} &  \textbf{\underline{0.596}} & \textbf{\underline{0.780}}                                                        \\ \cmidrule{3-7}
                          &                                       & Supervised        & 0.596  &  0.605 &  0.523 &  0.758                                                        \\ \cmidrule{2-7} 
                          & None                                  & None              & 0.434  &  0.679 & 0.366  &  0.621                                                        \\ \bottomrule
\end{tabular}
\end{table*}

\section{Object detection}
Object detection is the problem of recognising and locating any objects of interest in an image. In the context of GIE, the objects of interest may be polyps, tools, artefacts, or disease. In this section, we detail and present our evaluation of the fine-tuned performance of backbones in polyp detection specifically.

\subsection{Data}
The data used in our object detection experiments is taken from the Kvasir-SEG dataset\cite{kvasirseg}, which does not share any instances with Hyperkvasir-unlabelled. The dataset includes 1000 GIE images, each of which shows at least one polyp and is paired with both a set of bounding boxes, specifying the location and the horizontal and vertical dimensions of any polyps in the image, and a binary segmentation map indicating which pixels correspond to a polyp and which don't. While the segmentation maps were used in our semantic segmentation experiments, here we use the sets of bounding boxes. We applied a random 80\%/10\%/10\% training/validation/test split, where the validation data is used to determine whether to save the weights after each epoch of training on the training data, and the test data is reserved for evaluating the model after fine-tuning.

\subsection{Decoders}
For our object detection experiments, we implemented the listed backbones within a Faster R-CNN pipeline\cite{fasterrcnn} with feature pyramid network (FPN)\cite{fpn}, which we used the \texttt{torchvision} implementation of. We used the existing implementation of the pipeline with a ResNet50 backbone, specifying that all layers of the backbone should be trainable. For the ViT-B models, based on previous analyses of using ViT backbones in object detection\cite{vitmrcnn,vitdet}, we first modified the encoders to efficiently process larger image sizes\footnote{All considered pretraining was done with images resized to $224\times 224$, whereas object detection typically involves larger images, e.g. $1024\times 1024$.} by bilinearly interpolating the position embeddings and using \textit{non-overlapping} window self-attention in all but the 3rd, 6th, 9th, and 12th blocks. Window attention, also known as \textit{restricted} attention\cite{transformer}, independently applies attention to subsets of the sequence of visual tokens, where each subset corresponds to the tokens in a square window of the equivalent feature map, with no overlapping windows. We used 256 tokens in each subset, corresponding to a $16\times16$ window of a feature map. We then modified the Faster R-CNN pipeline to use the resulting encoders as backbones with a ViTDet FPN\cite{vitdet}.

\subsection{Fine-tuning procedure}

In the fine-tuning of both model architectures, we use the common fine-tuning procedure hyperparameters given in Table \ref{tab:common} and pre-process the training images using the pipeline detailed in Table \ref{tab:preprocessing}. For the ResNet50 models, using the default pre-processing pipeline for the Faster R-CNN implementation, the images in a batch are then each resized with bilinear interpolation to a scale of $\mathrm{min}(800/\mathrm{min}(h,w),1333/\mathrm{max}(h,w))$ of the original size $h\times w$, and then padded to $H\times W$, where $H$ is the maximum height of the resized images and $W$ is the maximum width of the resized images across the batch. For the ViT-B models, inspired by a previous analysis\cite{vitmrcnn}, the images are padded to $1024\times1024$ --- since several images in the dataset have a height or width larger than 1024, these images are downsampled to half the resolution using bicubic interpolation with anti-aliasing before padding. Transformations are also applied to the bounding boxes in accordance with any spatial transformations applied to the image. The usual multi-task loss function for the Faster R-CNN pipeline is used to compute the loss, and we use AP@[.5:.95] as the validation metric for predicted bounding boxes that have a confidence score $\ge$0.05:

\begin{equation}
    \text{AP@[.5:.95]}=\frac{1}{10}\sum_{t\in T}\text{AP@}t
\end{equation}

\noindent where $T=\left\{0.5,0.55,\ldots,0.95\right\}$ is the set of intersection over union (IoU) thresholds and AP@$t$ is the average precision at the $t^{th}$ IoU threshold. We compute AP@$t$ by first ranking all predicted bounding boxes with respect to the confidence score, from high to low. We then step through the predicted bounding boxes in rank order and assign the prediction to the true positives if it has an IoU with a target bounding box for the same image that is greater than the IoU threshold, and otherwise assign it to the false positives. At each rank, we then compute the precision and recall using the cumulative number of true positives and false positives and the total number of false negatives. We then determine a strictly monotonically increasing sequence of recall values $(r_i)_{i=1}^{N_r}$, with $r_{1}=0$, $r_{N_r}=1$, and $(r_i)_{i=2}^{N_r-1}$ being the recall values (excluding 0 and 1) for ranks where false positives and resulting drops in the precision occur, and AP@$t$ is then:

\begin{equation}
    \text{AP@}t=\sum_{i=2}^{N_r}\left(r_i-r_{i-1}\right)p(r_i)
\end{equation}

\noindent where $p(r_i)$ is the maximum precision value out of those which correspond to $r_i$, for $i=2,\ldots,N_r$. The transformations applied to the validation images include the same resizing and/or padding and normalisation applied to the training images, and a batch size of 1 is used to ensure the evaluation of a ResNet50 model on a particular instance is not influenced by other images in a batch (through the padding to $H\times W$). Finally, the model is trained on this basis for 200 epochs, with the parameters saved after each epoch that leads to an improvement in AP@[.5:.95] on the validation set, with any batch normalisation synchronised across GPUs.

\subsection{Evaluation}
\begin{figure*}[ht]
\makebox[\textwidth][c]{\begin{tabular}{cccccc}
  RN-HK-MC & RN-HK-BT  & RN-IN-MC  & RN-IN-BT  & RN-IN-SL  & RN-NA-NA  \\ 
 
 \includegraphics[width=2.5cm]{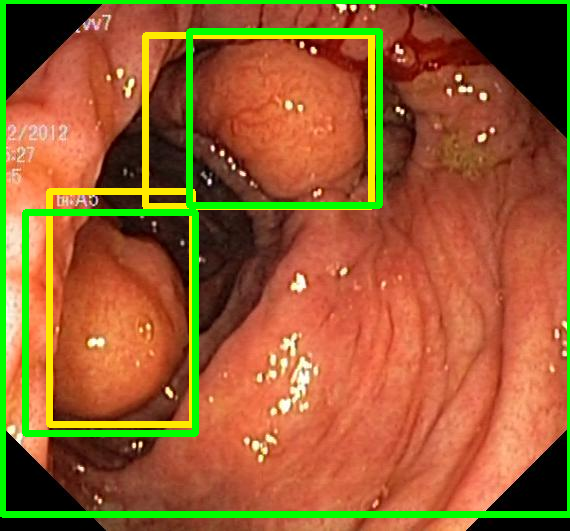}& \includegraphics[width=2.5cm]{test4_resnet50-Hyperkvasir_barlowtwins_init-frozen_False.png} & \includegraphics[width=2.5cm]{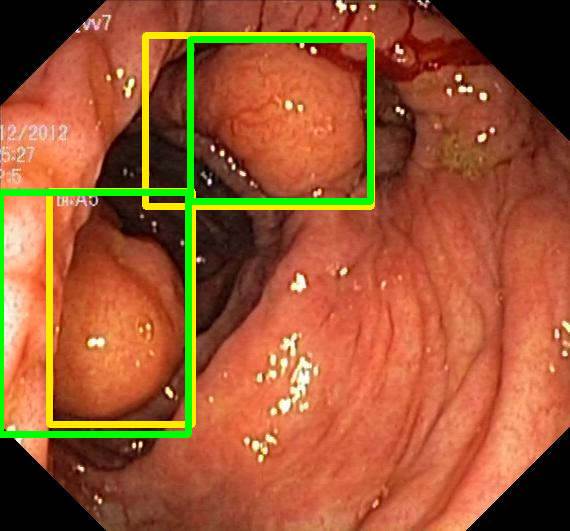} & \includegraphics[width=2.5cm]{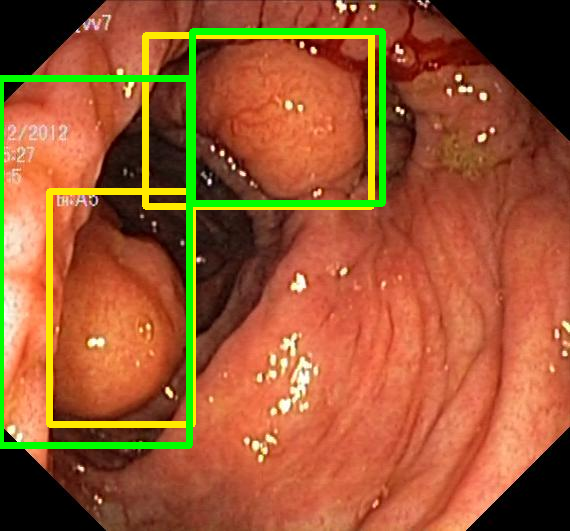} & \includegraphics[width=2.5cm]{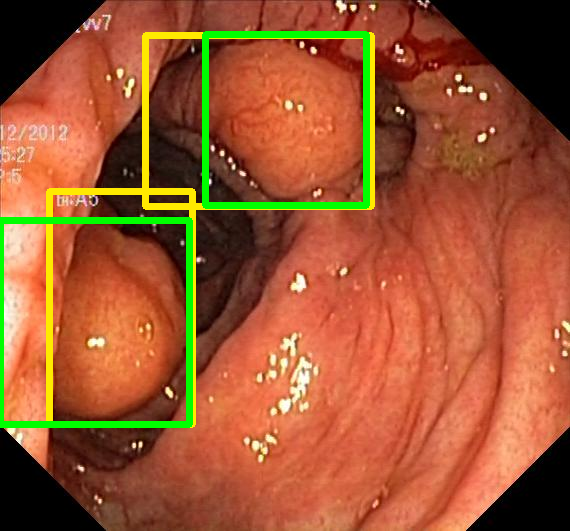} & \includegraphics[width=2.5cm]{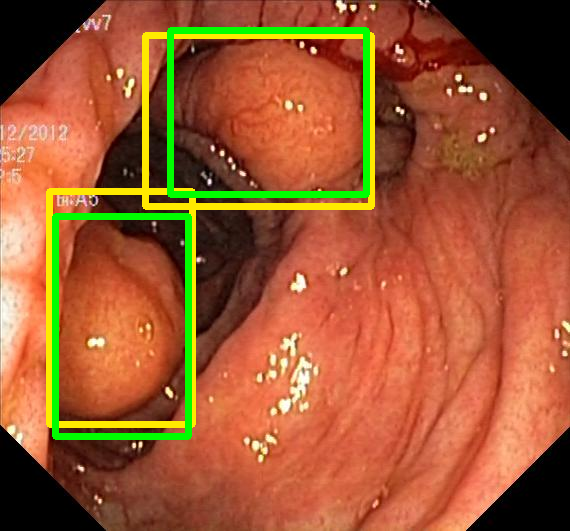} \\
   \includegraphics[width=2.5cm]{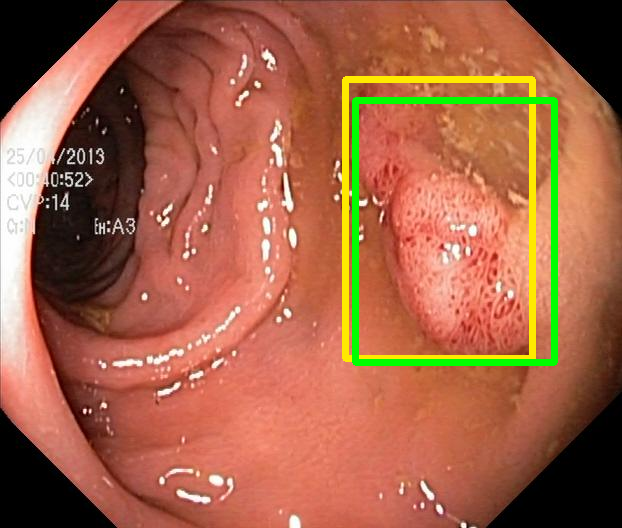} & \includegraphics[width=2.5cm]{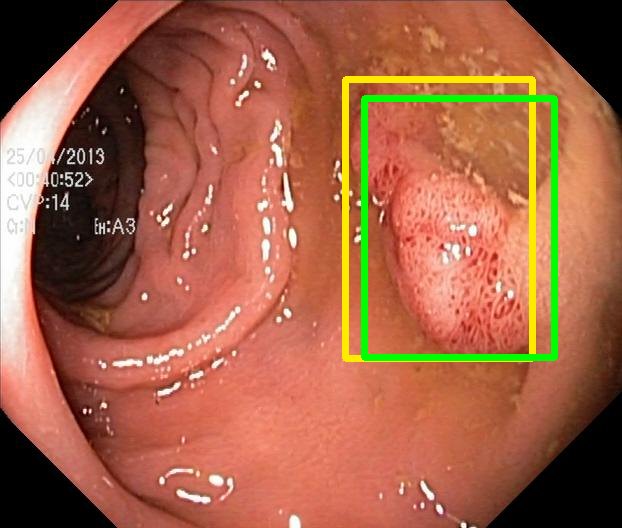} & \includegraphics[width=2.5cm]{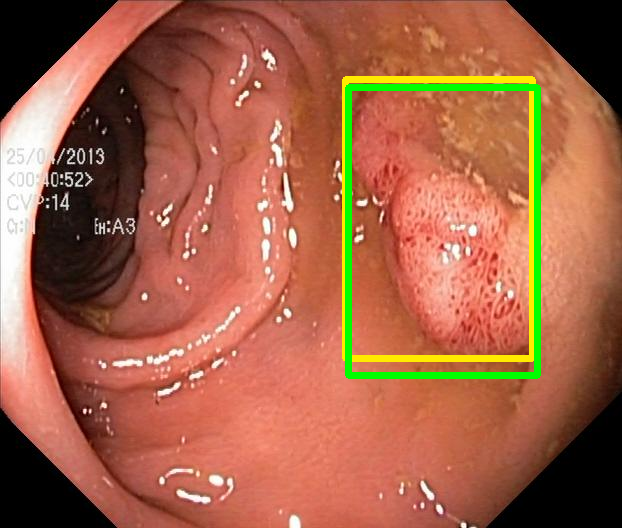} & \includegraphics[width=2.5cm]{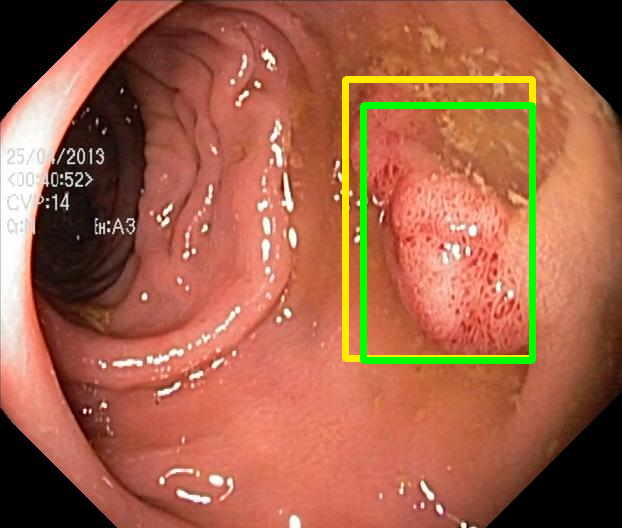} & \includegraphics[width=2.5cm]{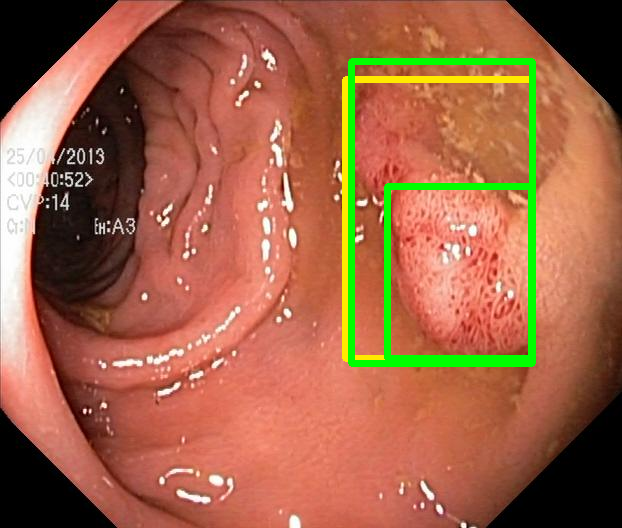} & \includegraphics[width=2.5cm]{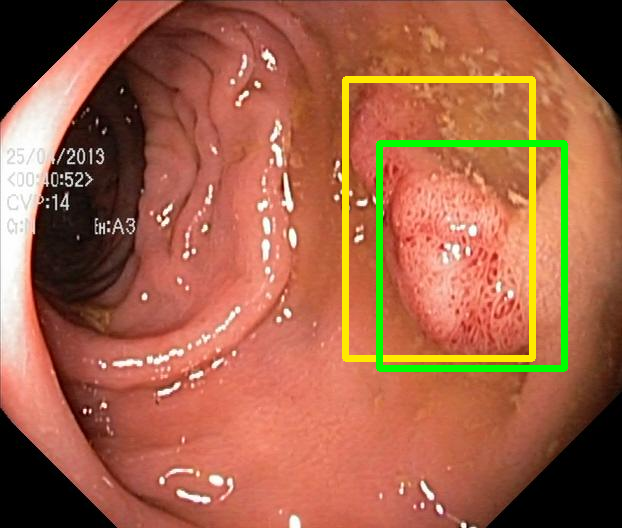} \\

  VT-HK-MC & VT-HK-MA  & VT-IN-MC  & VT-IN-MA  & VT-IN-SL  & VT-NA-NA  \\ 
 \includegraphics[width=2.5cm]{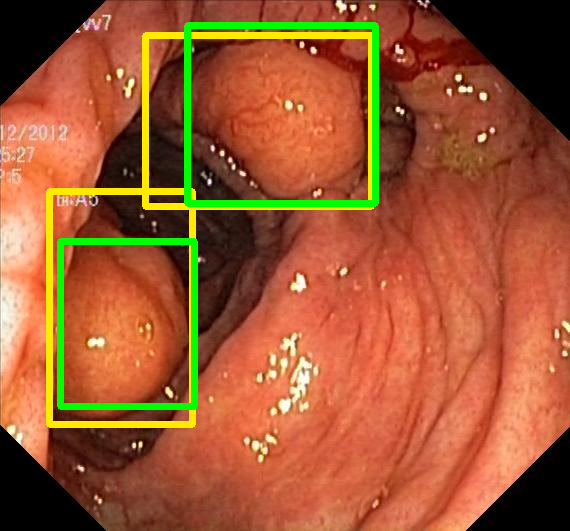}& \includegraphics[width=2.5cm]{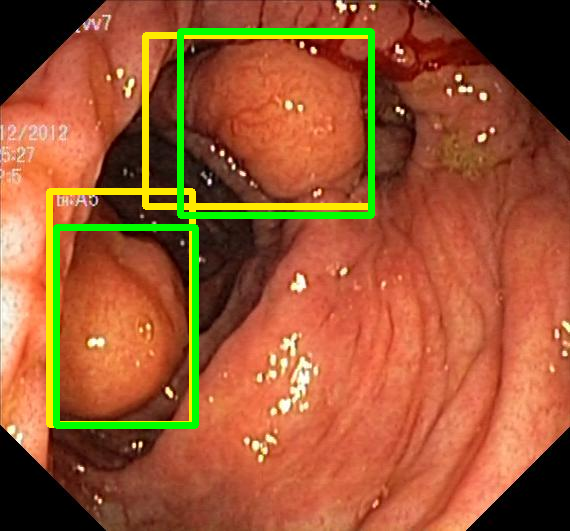} & \includegraphics[width=2.5cm]{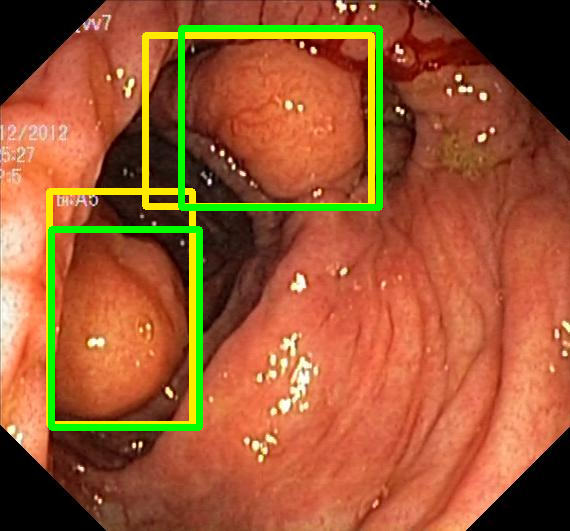} & \includegraphics[width=2.5cm]{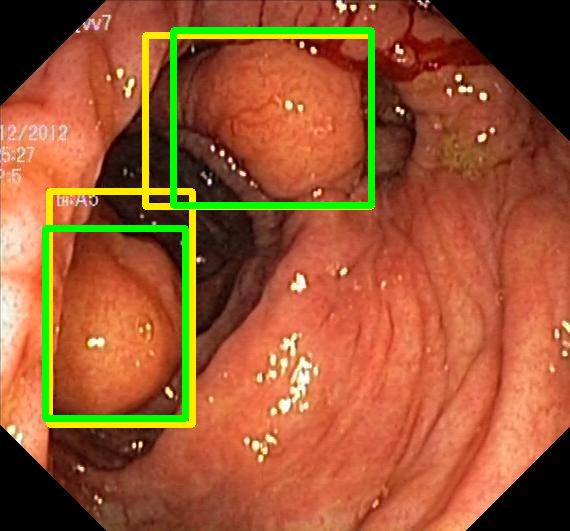} & \includegraphics[width=2.5cm]{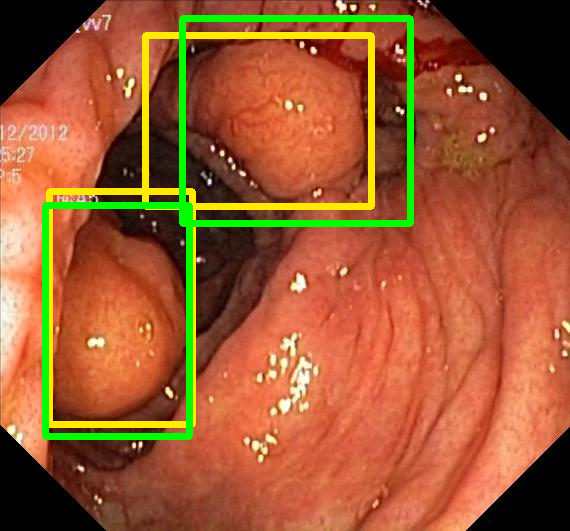} & \includegraphics[width=2.5cm]{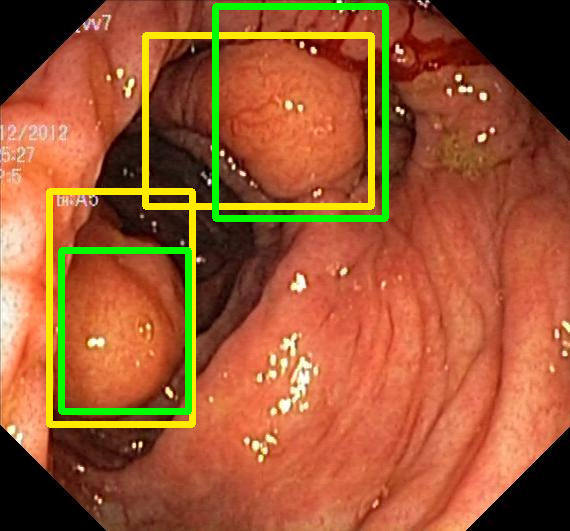} \\
   \includegraphics[width=2.5cm]{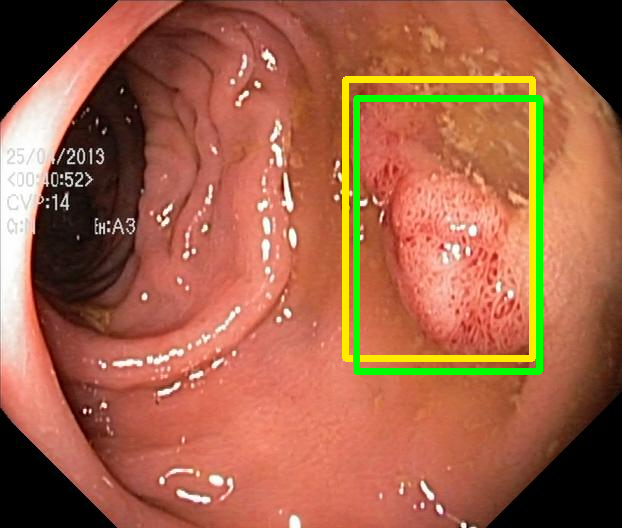} & \includegraphics[width=2.5cm]{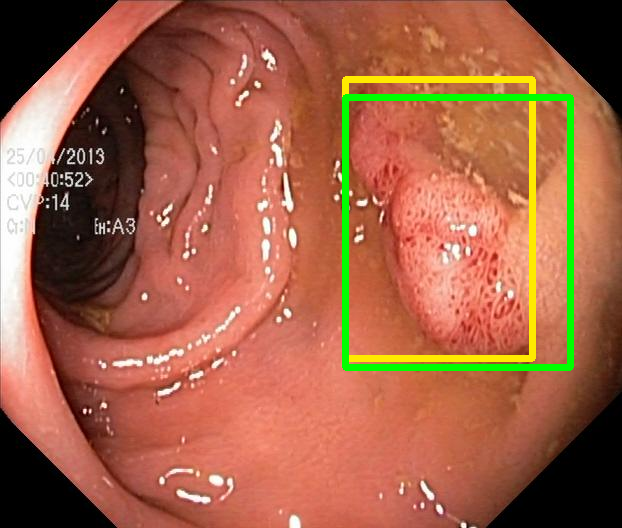} & \includegraphics[width=2.5cm]{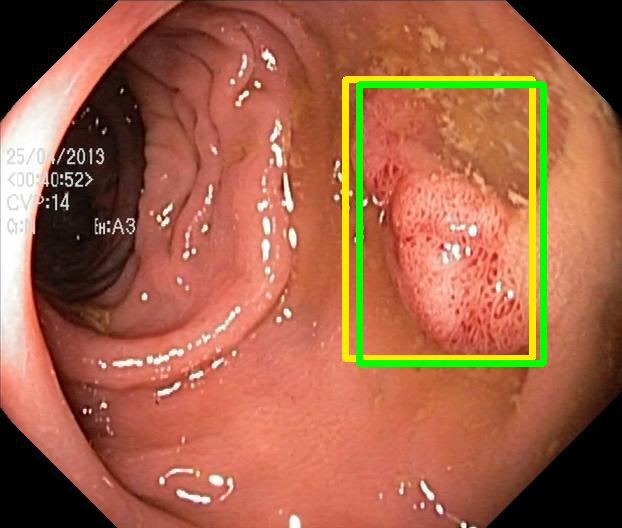} & \includegraphics[width=2.5cm]{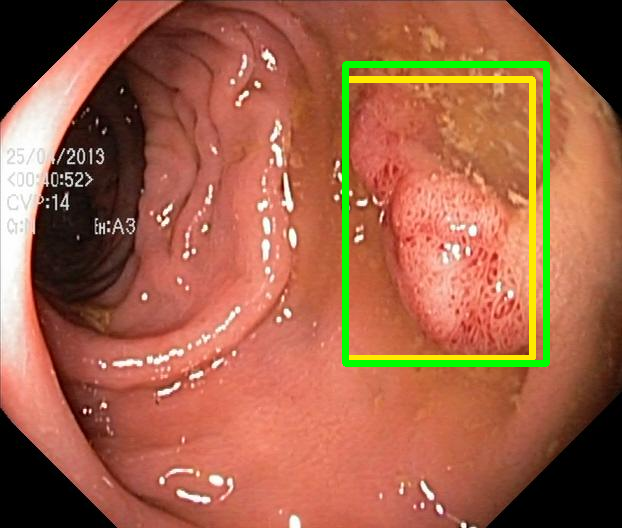} & \includegraphics[width=2.5cm]{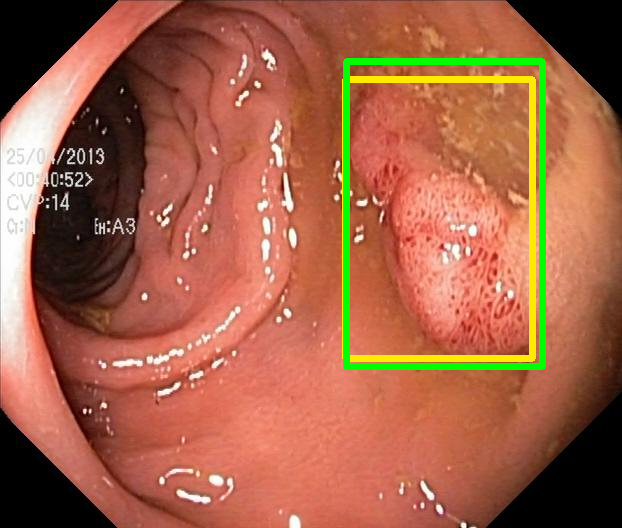}& \includegraphics[width=2.5cm]{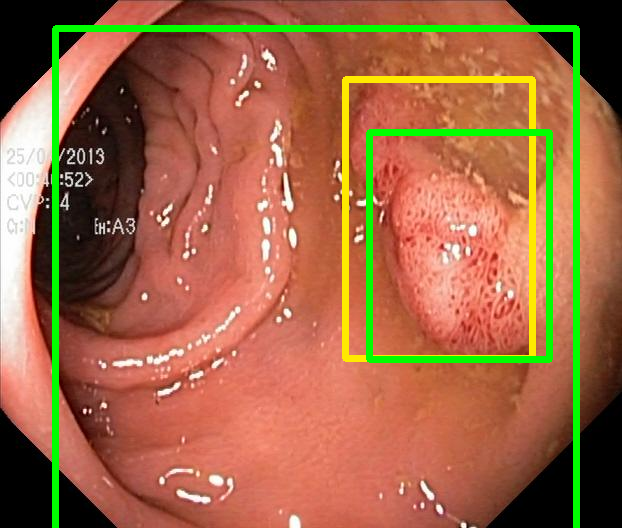}
\end{tabular}}
    \caption{Targets (yellow bounding boxes) and predictions (green bounding boxes) for two randomly selected instances of the Kvasir-SEG test set. For conciseness, we denote ResNet50s with \textit{RN}, ViT-Bs with \textit{VT}, Hyperkvasir-unlabelled with \textit{HK}, ImageNet-1k with \textit{IN}, MoCo v3 with \textit{MC}, Barlow Twins with \textit{BT}, MAE with \textit{MA}, supervised pretraining with \textit{SL}, and no pretraining with \textit{NA-NA}.}
    \label{fig:example_detection}
\end{figure*}

We evaluate the resulting object detection models using the test data, which is pre-processed in the same manner as the validation data, with AP@[.5:.95] (AP for conciseness), AP@.5 (AP$_{50}$), and AP@.75 (AP$_{75}$) computed for predicted bounded boxes with a confidence score $\ge$0.05. For all metrics, a higher value indicates better performance. The results are presented in Table \ref{tab:detection}, and some examples for predicted bounding boxes with a confidence score $\ge$0.5 are shown in Fig. \ref{fig:example_detection}.

\begin{table*}[ht]
\caption{\label{tab:detection}Performance in polyp detection. The best results for each architecture are highlighted as bold, and the best results overall are underlined.}
\centering
\begin{tabular}{cccccc}
\toprule
Backbone arch.            & Pretraining data                      & Pretraining algo. & AP & AP$_{50}$ & AP$_{75}$                                                
\\ \hline
\multirow{6}{*}{ResNet50} & \multirow{2}{*}{Hyperkvasir-unlabel.} & MoCo v3           & 0.604 &  0.895 &  0.702                            
\\
                          &                                       & Barlow Twins      & 0.647 &  0.895 & 0.717                                                          
                          \\ \cmidrule{2-6} 
                          & \multirow{3}{*}{ImageNet-1k}          & MoCo v3           & 0.640 &  0.905 & 0.731                                                    
                          \\ 
                          &                                       & Barlow Twins      & \textbf{\underline{0.653}}  &  \textbf{0.906} & \textbf{\underline{0.772}}   
                          \\ \cmidrule{3-6}
                          &                                       & Supervised        & 0.633 &  0.889 &  0.716                                                       
                          \\ \cmidrule{2-6} 
                          & None                                  & None              & 0.492  &  0.795 &  0.614                                                    
                          \\ \cmidrule{1-6}
\multirow{6}{*}{ViT-B}    & \multirow{2}{*}{Hyperkvasir-unlabel.} & MoCo v3           & 0.549  &  0.867 &  0.578                                                    
\\
                          &                                       & MAE               & 0.563  & 0.839  &  0.666
                          \\ \cmidrule{2-6} 
                          & \multirow{3}{*}{ImageNet-1k}          & MoCo v3           & 0.572 &  0.873 & 0.652                                                        
                          \\
                          &                                       & MAE               & \textbf{0.643} &  \textbf{\underline{0.921}} &  \textbf{0.748}                                                  
                          \\ \cmidrule{3-6}
                          &                                       & Supervised        &  0.577 &  0.832 &  0.685    
                          \\ \cmidrule{2-6} 
                          & None                                  & None              & 0.281 &  0.609 &  0.197                                   
                          \\ \bottomrule
\end{tabular}
\end{table*}

\section{Semantic segmentation}
Semantic segmentation is the problem of determining which, out of a predefined set of classes, each pixel in an image should be assigned to. In the context of GIE, the predefined set of classes will typically include a background class that accounts for anything that is not of interest, as well as any classes that are of interest, for example, polyps, tools, artefacts, or disease. In this section, we detail and present our evaluation of the fine-tuned performance of backbones in polyp segmentation specifically, which is notably a binary segmentation problem.

\subsection{Data}
We used two datasets in our semantic segmentation experiments, namely Kvasir-SEG\cite{kvasirseg} and CVC-ClinicDB\cite{cvc}. Kvasir-SEG has already been discussed in the context of our object detection experiments, and we use the same training/validation/test split here. CVC-ClinicDB includes 612 GIE images, each of which shows at least one polyp and is paired with a binary segmentation map indicating which pixels correspond to a polyp and which don't. We applied a random 80\%/10\%/10\% training/validation/test split, where the validation data is used to determine whether to save the weights after each epoch of training on the training data, and the test data is reserved for evaluating the model after fine-tuning.

\subsection{Decoders}
For our semantic segmentation experiments, we used the listed ResNet50 backbones with a DeepLabV3+\cite{deeplab} decoder, using the \texttt{segmentation-models-pytorch} implementation. We then used the ViT-B backbones with the segmentation variant of the dense prediction transformer (DPT)\cite{dpt} decoder, using the implementation provided in the official codebase.

\subsection{Fine-tuning procedure}
We separately train each model to perform polyp segmentation with each dataset through the same procedure. We use the common fine-tuning procedure hyperparameters given in Table \ref{tab:common} and pre-process the training images using the pipeline detailed in Table \ref{tab:preprocessing}. Transformations are also applied to the segmentation maps in accordance with any spatial transformations applied to the image. The loss is then computed using the Dice loss function\cite{diceloss}, and we use mDice as the validation metric:

\begin{equation}
    \text{mDice} = \frac{1}{N_e}\sum_{i=1}^{N_e}\frac{2\mathrm{TP}_i + \epsilon}{2\mathrm{TP}_i+\mathrm{FP}_i+\mathrm{FN}_i+\epsilon}\label{mDice}
\end{equation}

\noindent where $N_e$ is the number of instances in the validation/test set, $\mathrm{TP}_i$ is the number of true positives for the $i^{th}$ image, $\mathrm{FP}_i$ is the number of false positives, $\mathrm{FN}_i$ is the number of false negatives, and $\epsilon=1e-8$. The transformations applied to the validation images include the same resizing and normalisation applied to the training images, with the validation maps also resized to $224\times 224$. Finally, the model is trained on this basis for 200 epochs, with the parameters saved after each epoch that leads to an improvement in mDice on the validation set, with any batch normalisation synchronised across GPUs.

\subsection{Evaluation}
We evaluate the resulting semantic segmentation models using the corresponding test data, where the images are pre-processed in the same manner as the validation images, but the segmentation maps are left at their original size. The predictions are therefore resized to this original size using bilinear interpolation prior to binarisation. We then use four metrics, namely the mDice (as defined in \eqref{mDice}), mIoU, mPrecision, and mRecall:

\begin{equation}
    \text{mIoU}= \frac{1}{N_e}\sum_{i=1}^{N_e}\frac{\mathrm{TP}_i+\epsilon}{\mathrm{TP}_i+\mathrm{FP}_i+\mathrm{FN}_i+\epsilon}
\end{equation}

\begin{equation}
    \text{mPrecision}= \frac{1}{N_e}\sum_{i=1}^{N_e}\frac{\mathrm{TP}_i+\epsilon}{\mathrm{TP}_i+\mathrm{FP}_i+\epsilon}
\end{equation}

\begin{equation}
    \text{mRecall}= \frac{1}{N_e}\sum_{i=1}^{N_e}\frac{\mathrm{TP}_i+\epsilon}{\mathrm{TP}_i+\mathrm{FN}_i+\epsilon}
\end{equation}

\begin{figure*}[ht]
\makebox[\textwidth][c]{\begin{tabular}{cccccccc}
  Input & Target & RN-HK-MC & RN-HK-BT  & RN-IN-MC  & RN-IN-BT  & RN-IN-SL  & RN-NA-NA  \\ 
 \includegraphics[width=1.8cm]{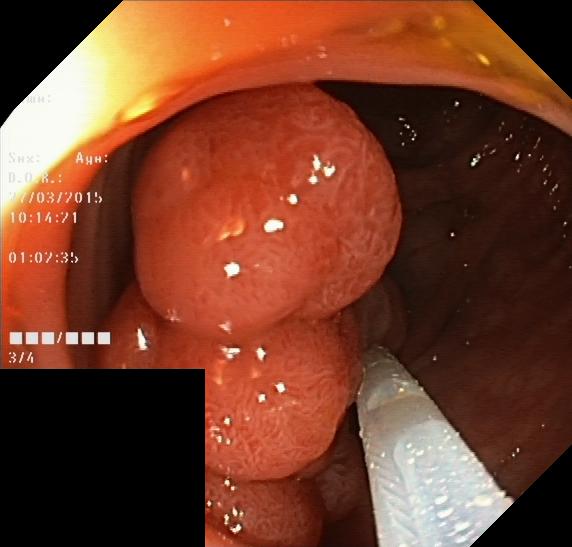}& \includegraphics[width=1.8cm]{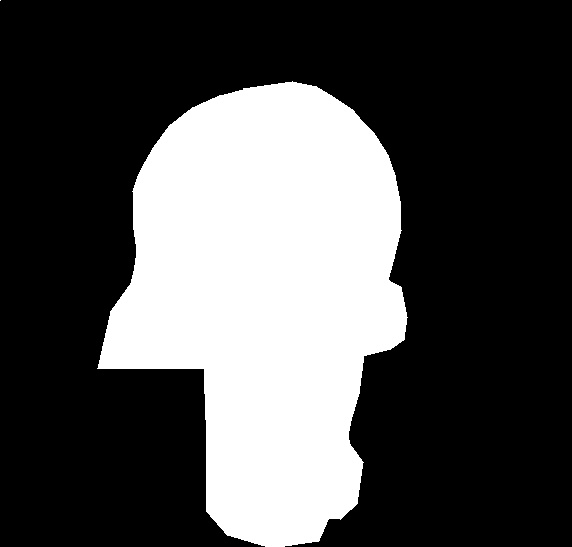} & 
 \includegraphics[width=1.8cm]{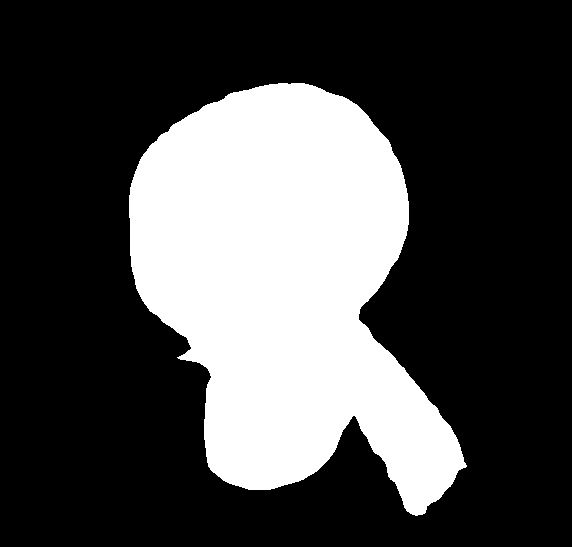}& \includegraphics[width=1.8cm]{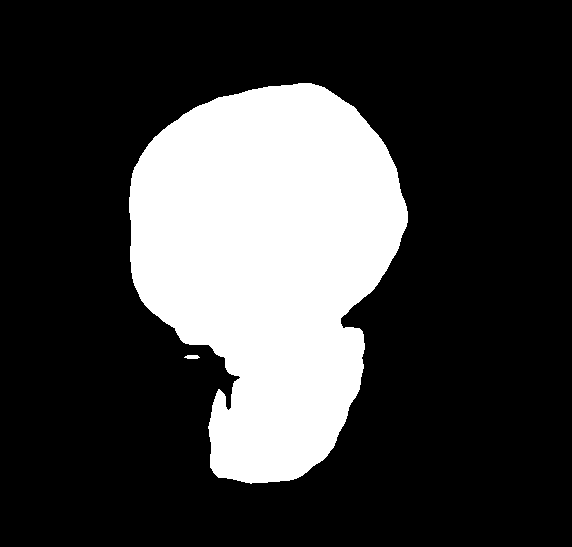} & \includegraphics[width=1.8cm]{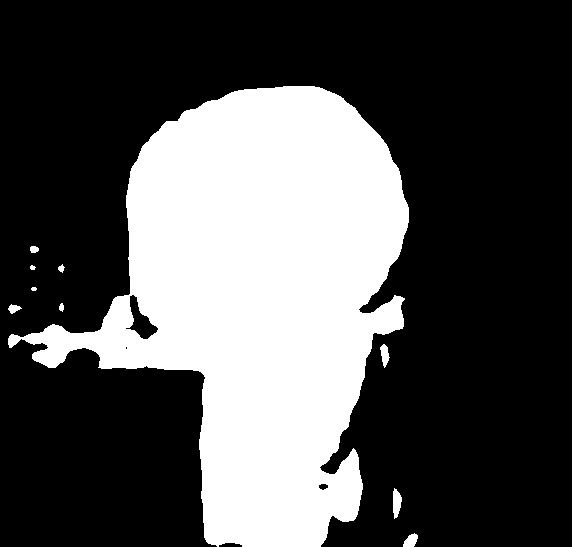} & \includegraphics[width=1.8cm]{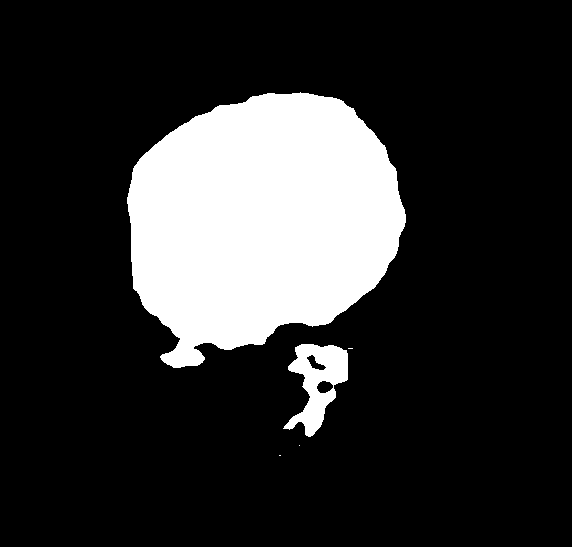} & \includegraphics[width=1.8cm]{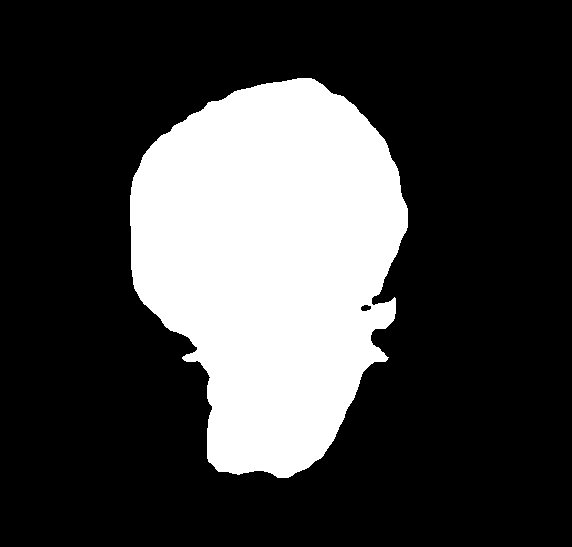} & \includegraphics[width=1.8cm]{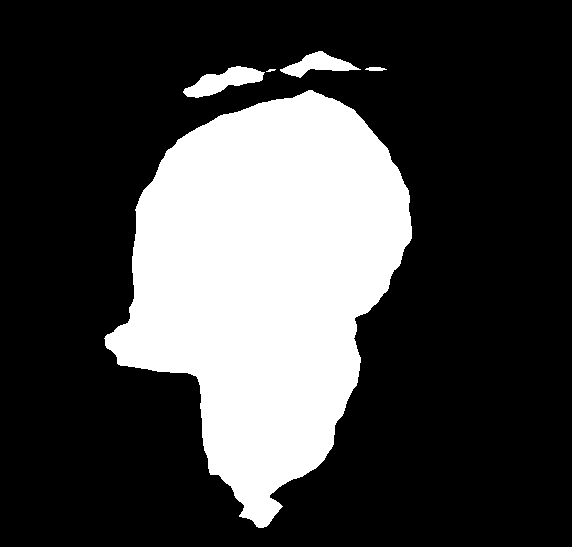} \\
  \includegraphics[width=1.8cm]{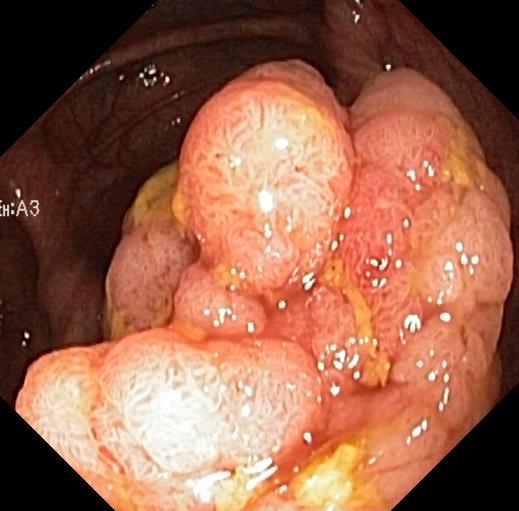}& \includegraphics[width=1.8cm]{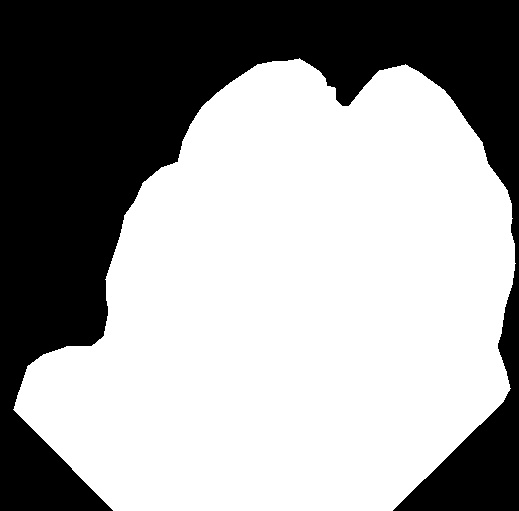} & \includegraphics[width=1.8cm]{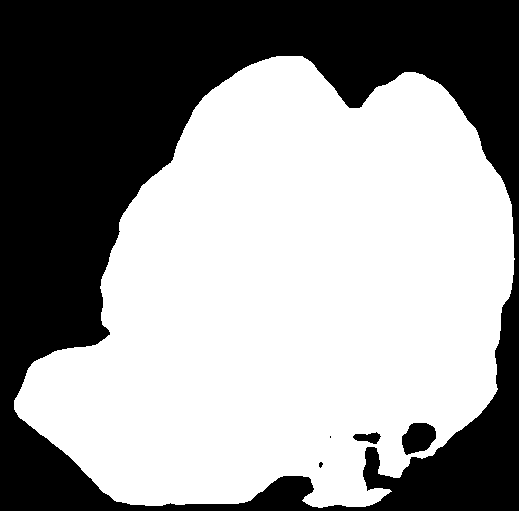} & \includegraphics[width=1.8cm]{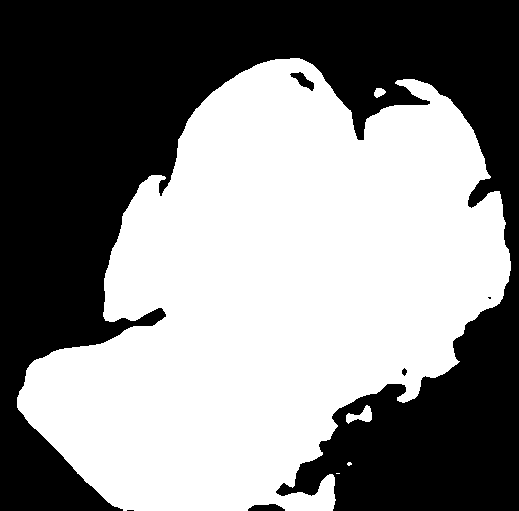} & \includegraphics[width=1.8cm]{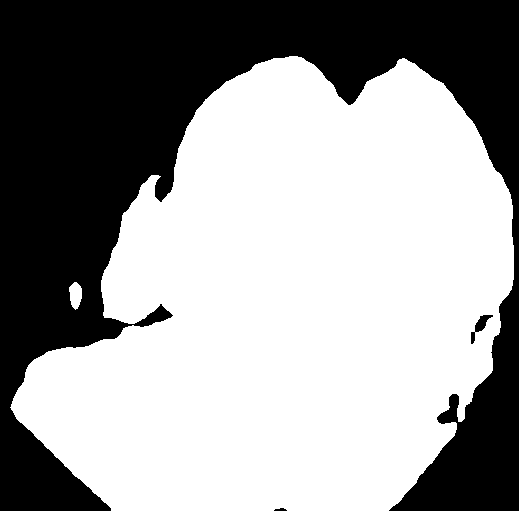} & \includegraphics[width=1.8cm]{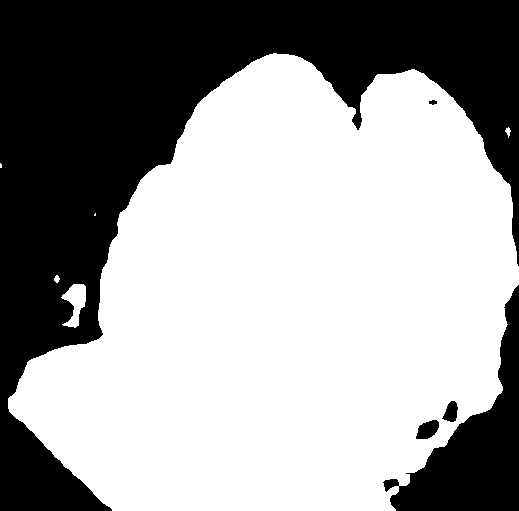} & \includegraphics[width=1.8cm]{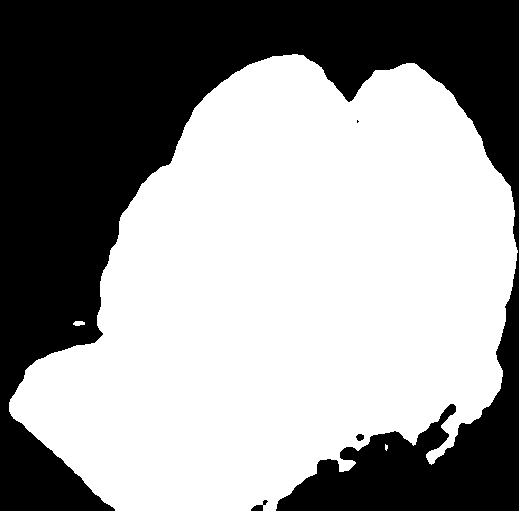} & \includegraphics[width=1.8cm]{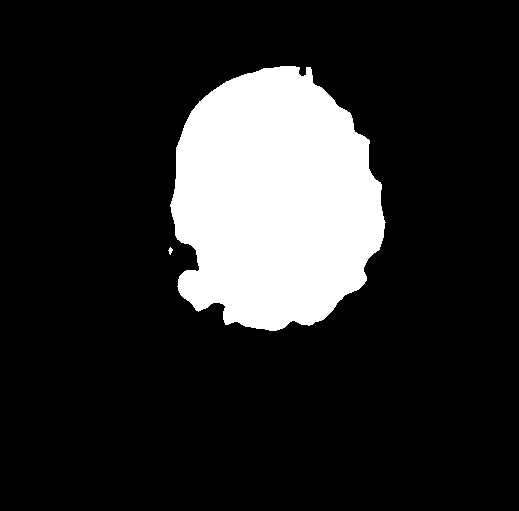} \\

  Input & Target & VT-HK-MC & VT-HK-MA  & VT-IN-MC  & VT-IN-MA  & VT-IN-SL  & VT-NA-NA  \\ 
 \includegraphics[width=1.8cm]{cju2ysg748ru80878sp6j0gm0RGB.jpg}& \includegraphics[width=1.8cm]{cju2ysg748ru80878sp6j0gm0.jpg} & 
 \includegraphics[width=1.8cm]{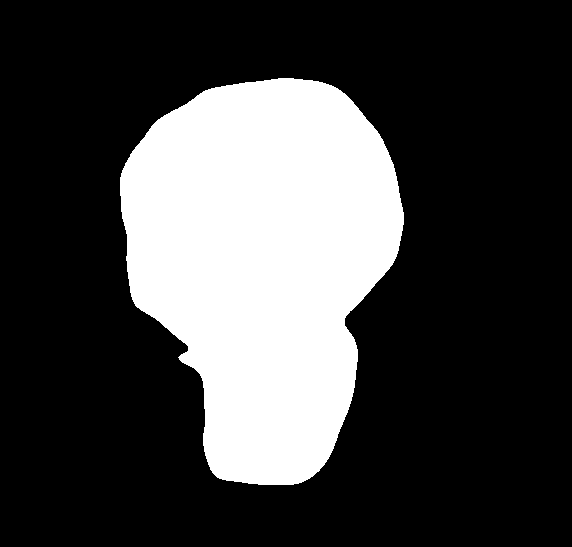}& \includegraphics[width=1.8cm]{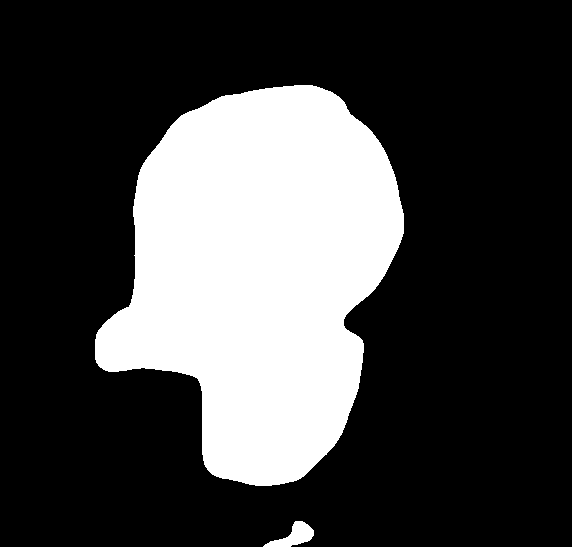} & \includegraphics[width=1.8cm]{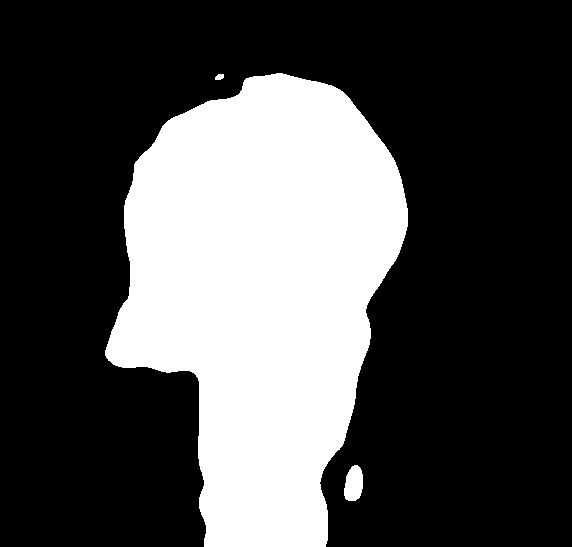} & \includegraphics[width=1.8cm]{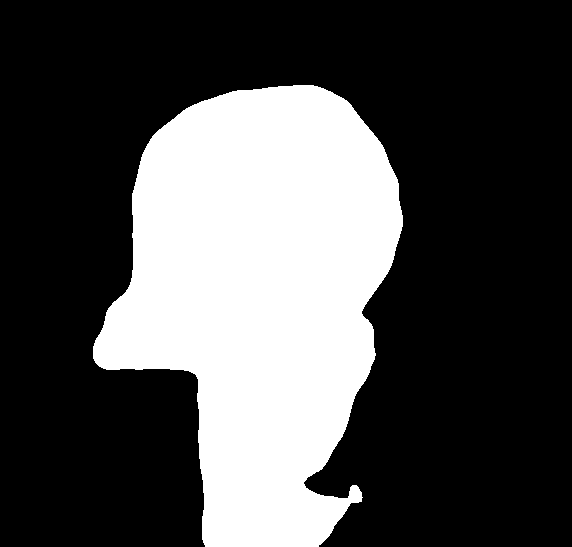} & \includegraphics[width=1.8cm]{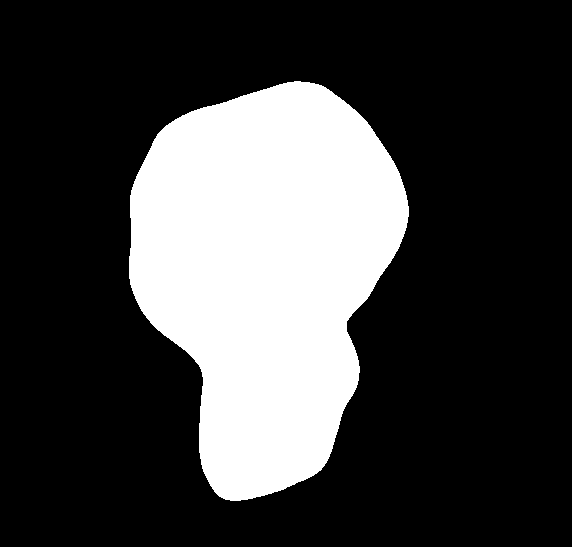} & \includegraphics[width=1.8cm]{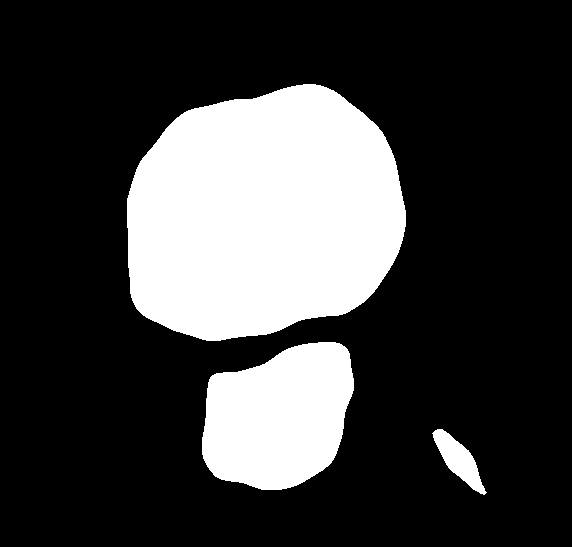} \\
  \includegraphics[width=1.8cm]{cju17otoe119u0799nqcbl8n1RGB.jpg}& \includegraphics[width=1.8cm]{cju17otoe119u0799nqcbl8n1.jpg} & \includegraphics[width=1.8cm]{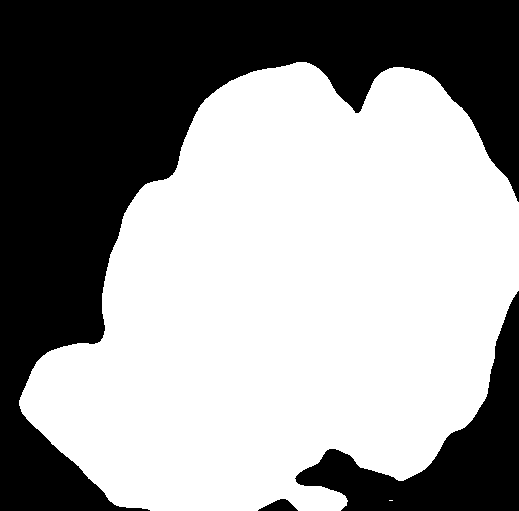} & \includegraphics[width=1.8cm]{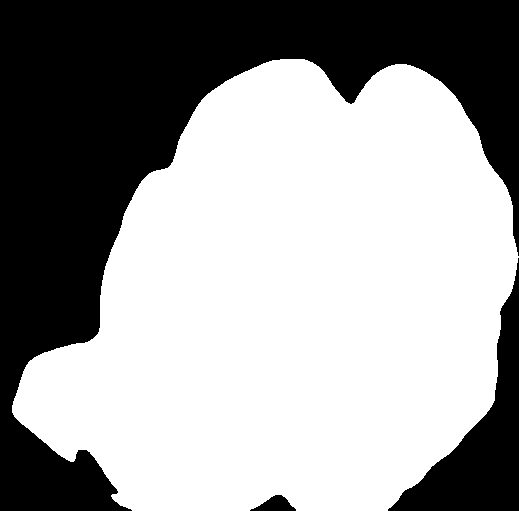} & \includegraphics[width=1.8cm]{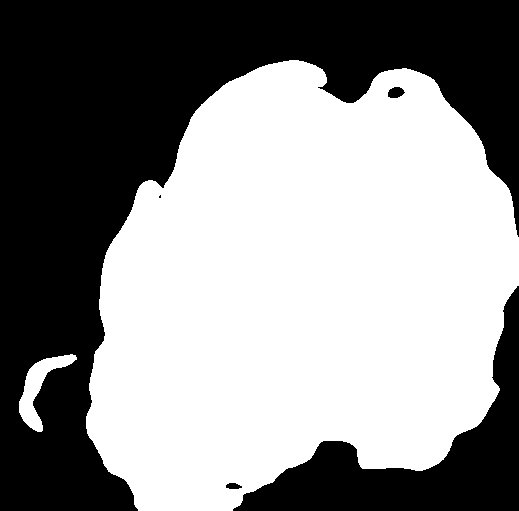} & \includegraphics[width=1.8cm]{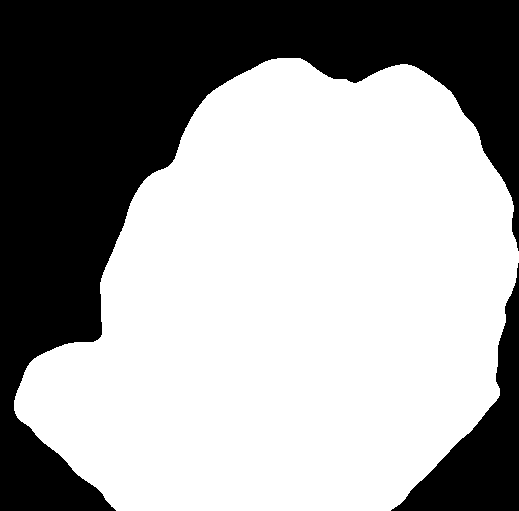} & \includegraphics[width=1.8cm]{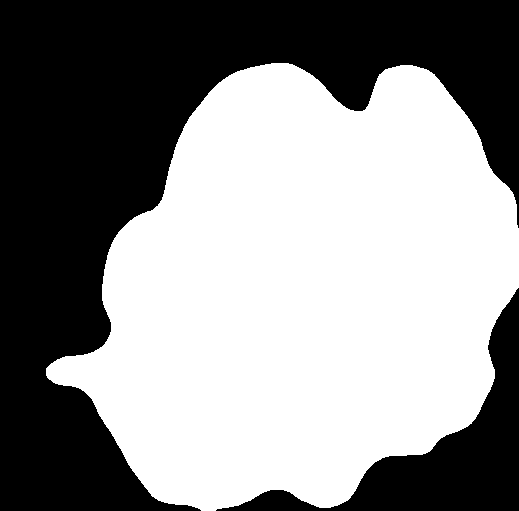}& \includegraphics[width=1.8cm]{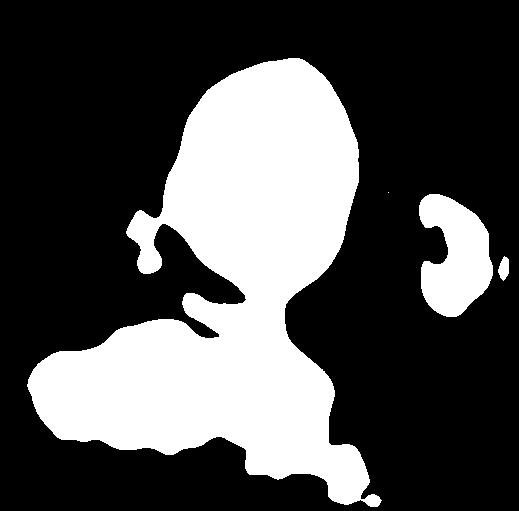}
\end{tabular}}
    \caption{Targets and predictions for two randomly selected instances of the Kvasir-SEG test set. For conciseness, we denote ResNet50s with \textit{RN}, ViT-Bs with \textit{VT}, Hyperkvasir-unlabelled with \textit{HK}, ImageNet-1k with \textit{IN}, MoCo v3 with \textit{MC}, Barlow Twins with \textit{BT}, MAE with \textit{MA}, supervised pretraining with \textit{SL}, and no pretraining with \textit{NA-NA}.}
    \label{fig:example_kvasir}
\end{figure*}

For all metrics, a higher value indicates better performance. The results for Kvasir-SEG are presented in Table \ref{tab:kvasir} and the results for CVC-ClinicDB are presented in Table \ref{tab:cvc}. Examples for Kvasir-SEG are shown in Fig. \ref{fig:example_kvasir}.

\begin{table*}[ht]
\caption{\label{tab:kvasir}Performance in polyp segmentation with Kvasir-SEG. The best results for each architecture are highlighted as bold, and the best results overall are underlined.}
\centering
\begin{tabular}{ccccccc}
\toprule
Backbone arch.            & Pretraining data                      & Pretraining algo. & mDice & mIoU & mPrecision & mRecall                                                   \\ \hline
\multirow{6}{*}{ResNet50} & \multirow{2}{*}{Hyperkvasir-unlabel.} & MoCo v3           & 0.841 &  0.753 &  0.857 &  0.878                              \\
                          &                                       & Barlow Twins      & 0.852 &  0.772 & 0.859  &  0.900                                                        \\ \cmidrule{2-7} 
                          & \multirow{3}{*}{ImageNet-1k}          & MoCo v3           & \textbf{0.883}  &  \textbf{0.812} &  0.866 & \textbf{\underline{0.936}}                                                        \\ 
                          &                                       & Barlow Twins      & 0.873  &  0.795 &  0.879 & 0.899                                                        \\ \cmidrule{3-7}
                          &                                       & Supervised        & 0.871 &  0.800 &  \textbf{0.882} & 0.893                                                        \\ \cmidrule{2-7} 
                          & None                                  & None              & 0.632  & 0.506  &  0.639 &  0.780                                                        \\ \cmidrule{1-7}
\multirow{6}{*}{ViT-B}    & \multirow{2}{*}{Hyperkvasir-unlabel.} & MoCo v3           & 0.861 &  0.788 &  0.867 &  0.898                                                        \\
                          &                                       & MAE               & 0.885 &  0.816 &  0.899 &  0.906 \\ \cmidrule{2-7} 
                          & \multirow{3}{*}{ImageNet-1k}          & MoCo v3           & 0.889 & 0.824  & 0.900  &  \textbf{0.907}                                                        \\
                          &                                       & MAE               & \textbf{\underline{0.896}} &  \textbf{\underline{0.834}} & \textbf{\underline{0.921}}  &  0.902                                                        \\ \cmidrule{3-7}
                          &                                       & Supervised        & 0.871  &  0.795 &  0.894 & 0.883                                                        \\ \cmidrule{2-7} 
                          & None                                  & None              & 0.755 &  0.650 &  0.785 &  0.815                                                        \\ \bottomrule
\end{tabular}
\end{table*}

\begin{table*}[ht]
\caption{\label{tab:cvc}Performance in polyp segmentation with CVC-ClinicDB. The best results for each architecture are highlighted as bold, and the best results overall are underlined.}
\centering
\begin{tabular}{ccccccc}
\toprule
Backbone arch.            & Pretraining data                      & Pretraining algo. & mDice & mIoU & mPrecision & mRecall                                                   \\ \hline
\multirow{6}{*}{ResNet50} & \multirow{2}{*}{Hyperkvasir-unlabel.} & MoCo v3           & 0.909 &  0.839 & 0.906  &  \textbf{0.921}                              \\
                          &                                       & Barlow Twins      & 0.880  &  0.799 &  0.916 &  0.863                                                        \\ \cmidrule{2-7} 
                          & \multirow{3}{*}{ImageNet-1k}          & MoCo v3           & \textbf{0.920}  &  \textbf{0.856} &  \textbf{0.938} &  0.909                                                        \\ 
                          &                                       & Barlow Twins      & 0.901 & 0.826  &  0.893 &  0.920                                                        \\ \cmidrule{3-7}
                          &                                       & Supervised        & 0.879  &  0.805 &  0.933 & 0.861                                                        \\ \cmidrule{2-7} 
                          & None                                  & None              & 0.595  & 0.462  & 0.584  &  0.678                                                        \\ \cmidrule{1-7}
\multirow{6}{*}{ViT-B}    & \multirow{2}{*}{Hyperkvasir-unlabel.} & MoCo v3           & 0.909  & 0.848  & 0.916  &  0.926                                                        \\
                          &                                       & MAE               & 0.901  &  0.838 &  0.920 &  0.903 \\ \cmidrule{2-7} 
                          & \multirow{3}{*}{ImageNet-1k}          & MoCo v3           & 0.911  & 0.848  & \textbf{\underline{0.940}}  & 0.901                                                        \\
                          &                                       & MAE               & \textbf{\underline{0.927}}  &  \textbf{\underline{0.867}} &  0.926 & \textbf{\underline{0.933}}                                                        \\ \cmidrule{3-7}
                          &                                       & Supervised        & 0.907  &  0.841 &  0.912 &  0.910                                                        \\ \cmidrule{2-7} 
                          & None                                  & None              & 0.813  &  0.708 &  0.839 & 0.826                                                        \\ \bottomrule
\end{tabular}
\end{table*}

\section{Monocular depth estimation}
Monocular depth estimation is the problem of predicting the length of the ray of light, that a particular pixel in an image corresponds to, between the camera and the object that the ray of light has come from, for every pixel in the image. Since the absolute scale of the scene can only be determined from the parallax observed with a second view, the problem is however inherently ill-posed and only relative scale can be determined. In this section, we detail and present our evaluation of the fine-tuned performance of backbones in monocular depth estimation in colonoscopy.

\subsection{Data}
The data used for our depth estimation experiments is taken from the C3VD dataset\cite{c3vd}, the only dataset that we know of which includes images captured with a clinical GIE camera (colonoscope, specifically) with paired ground truth depth maps. The dataset was collected by recording segments (sigmoid, descending, transcending, ascending, and cecum) of a high-fidelity 3D silicone phantom colon model with varying textures, emulating different patient-specific tissue features and vasculature patterns at varying optical depths, and varying illumination modes with a clinical colonoscope. Views of an equivalent 3D virtual colon model were then registered with key frames of the resulting videos, allowing for the rendering of a ground truth depth map for each frame, as well as a surface normal, optical flow, and occlusion map. Each video is also paired with ground truth camera pose, surface model, and coverage map. 22 videos were recorded, with variation in the segment, camera pose, textures, and illumination, amounting to 10015 frames in total. We selected 18 videos (8610 frames) for training, 2 videos (977 frames) for validation, and 2 videos (528 frames) for testing, where the validation and test sets each include one randomly sampled video of the cecum and one randomly sampled video of the transcending segment, since the majority of videos were of one of these segments (8 of cecum and 9 of transcending).

\subsection{Decoders}
For our monocular depth estimation experiments, we used the listed ViT-B backbones with the depth estimation variant of the dense prediction transformer (DPT)\cite{dpt} decoder, using the implementation provided in the official codebase. Since there is no clear precedent for a decoder architecture for ResNet50-based depth estimation\footnote{Popular dense prediction architectures that adopt certain details of ResNets in their design and which may be suitable for depth estimation, such as ResUNet\cite{resunet} or ResUNet++\cite{resunet++}, do not actually use a ResNet encoder.}, we designed our own. This decoder, designed to mirror the architecture of ResNet50, has three fusion levels. The first starts with the final feature maps output by a ResNet50 and halves the number of channels with a $1\times1$ convolutional layer followed by batch normalisation, before upsampling the resulting feature maps to twice the resolution with bilinear interpolation and concatenating it with the feature maps output by the previous level of the ResNet50. The concatenated features are then processed by three blocks that have the same design as the blocks used in each level of ResNet50. The second and third levels of the decoder follow the same logic as the first, except that they start with the output of the previous level of the decoder. A prediction head, which has the same design as the prediction head used in the depth estimation variant of the DPT decoder, is then used to predict a depth map from the output of the third level.

\subsection{Fine-tuning procedure}
In the fine-tuning of both model architectures, we use the common fine-tuning procedure hyperparameters given in Table \ref{tab:common} and pre-process the training images using the pipeline detailed in Table \ref{tab:preprocessing}. Transformations are also applied to the depth maps in accordance with any spatial transformations applied to the image, with absolute depth values scaled to $[0,1]$. The loss is then computed using the scale- and shift-invariant (SSI) mean squared error (MSE)\cite{midas} with a multi-scale shift-invariant gradient matching term\cite{megadepth}, which is computed only on the pixels that are covered by the lens (corners are not covered --- see examples in Fig. \ref{fig:example_depth}), and we use the mSSI-MSE for pixels covered by the lens as the validation metric:

\begin{equation}
    \text{mSSI-MSE} = \frac{1}{N_eN_v}\sum_{i=1}^{N_e}\sum_{j=1}^{N_v}\left(s_i\hat{y}_{i,j}+t_i-y_{i,j}\right)^2
\end{equation}

\noindent where $N_v$ is the number of pixels covered by the lens in an image, $\hat{y}_{i,j}$ is the output value for the $j^{th}$ pixel covered by the lens in the $i^{th}$ image, $y_{i,j}$ is the corresponding target value, and $s_i$ and $t_i$ are the scale and shift computed using the closed form solution to the standard least squares problem:

\begin{equation}
\mathbf{h}_i^*=\mathrm{arg}\min_{\mathbf{h}_i}\sum_{j=1}^{N_v}\left(\hat{\mathbf{y}}_{i,j}^\top\mathbf{h}_i -y_{i,j}\right)^2
\end{equation}

\noindent where $\mathbf{h}_i=\left(s_i,t_i\right)^\top$ and $\hat{\mathbf{y}}_{i,j}=\left(\hat{y}_{i,j},1\right)^\top$. The transformations applied to the validation images include the same padding, resizing, and normalisation applied to the training images, with the validation maps also padded and resized to $224\times 224$ and depth values scaled to $[0,1]$. Finally, the model is trained on this basis for 50 epochs, with the parameters saved after each epoch that leads to an improvement in SSI MSE on the validation set, with any batch normalisation synchronised across GPUs.

\subsection{Evaluation}
\begin{figure*}[ht]
\makebox[\textwidth][c]{\begin{tabular}{cccccccc}
  Input & Target & RN-HK-MC & RN-HK-BT  & RN-IN-MC  & RN-IN-BT  & RN-IN-SL  & RN-NA-NA  \\ 
 \includegraphics[width=1.8cm]{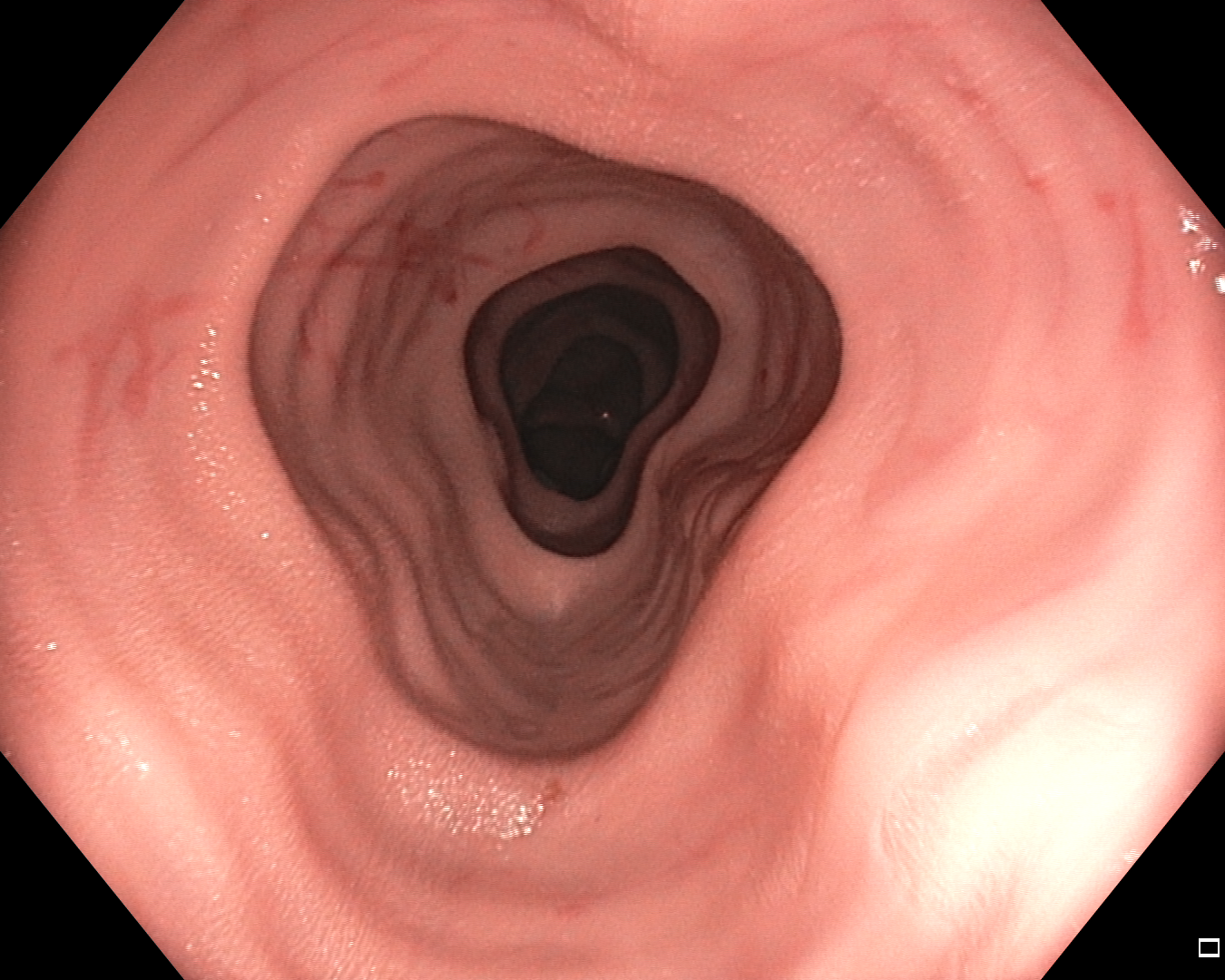}& \includegraphics[width=1.8cm]{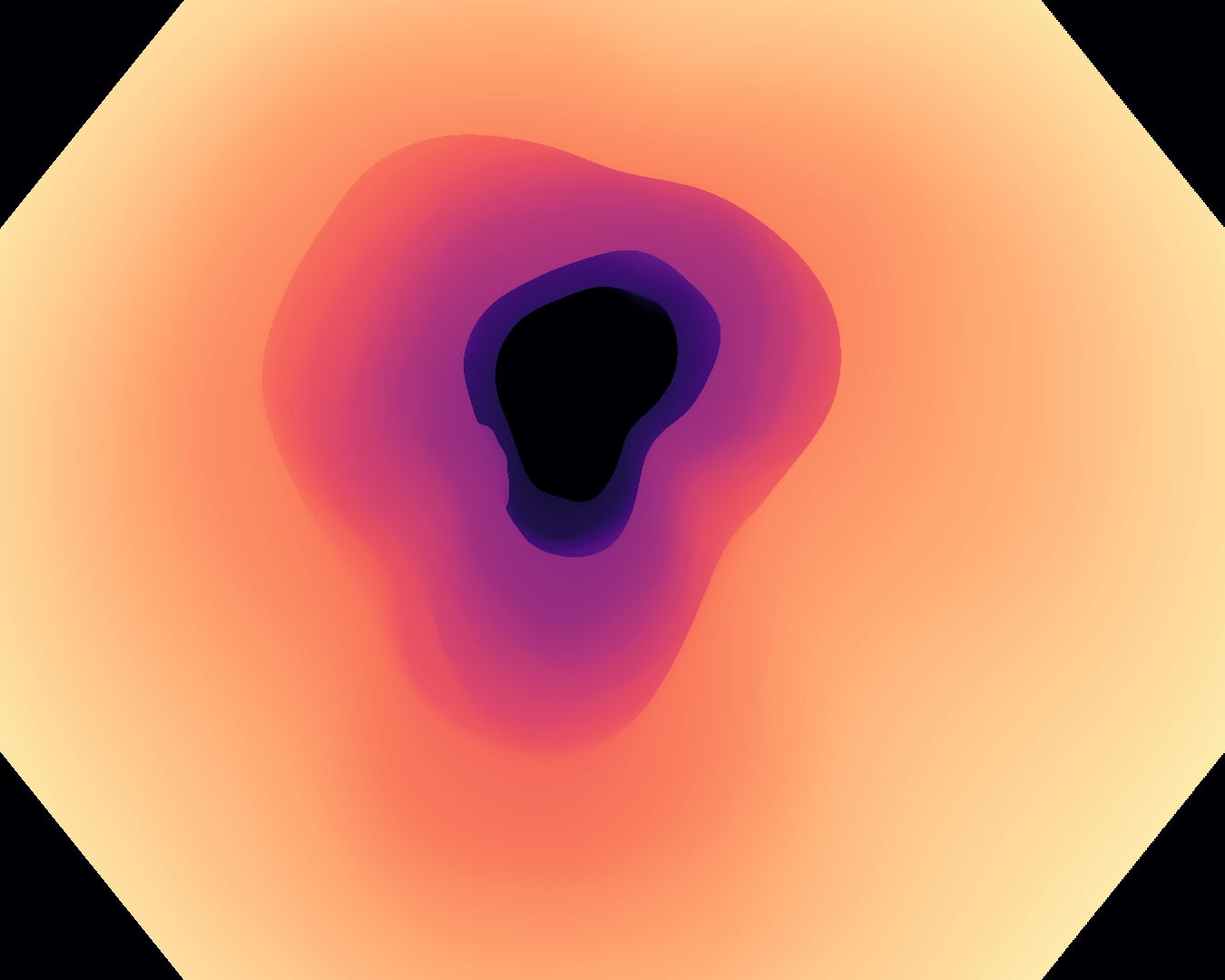} &
 \includegraphics[width=1.8cm]{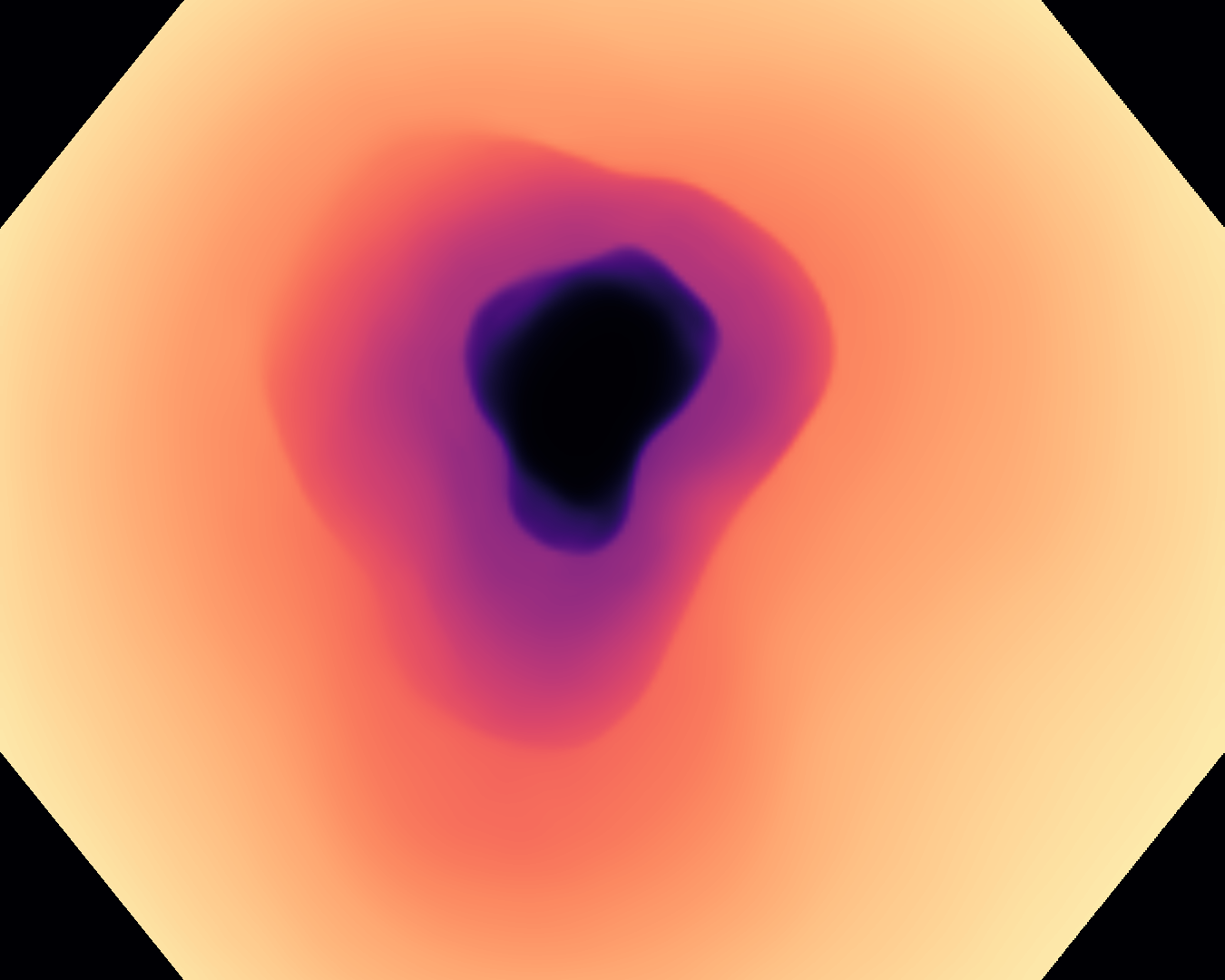}& \includegraphics[width=1.8cm]{test11_resnet50-Hyperkvasir_barlowtwins_init-frozen_False.png} & \includegraphics[width=1.8cm]{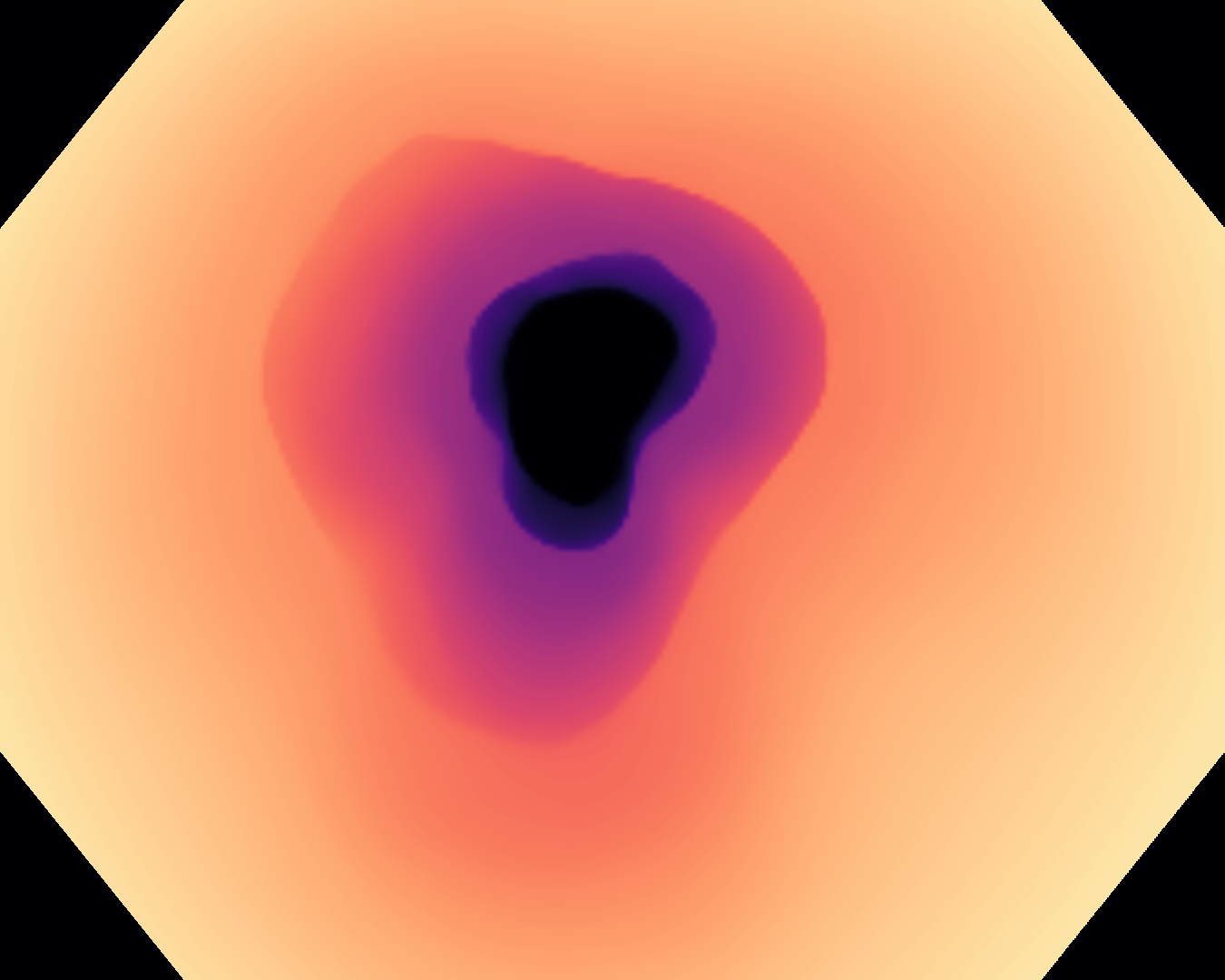} & \includegraphics[width=1.8cm]{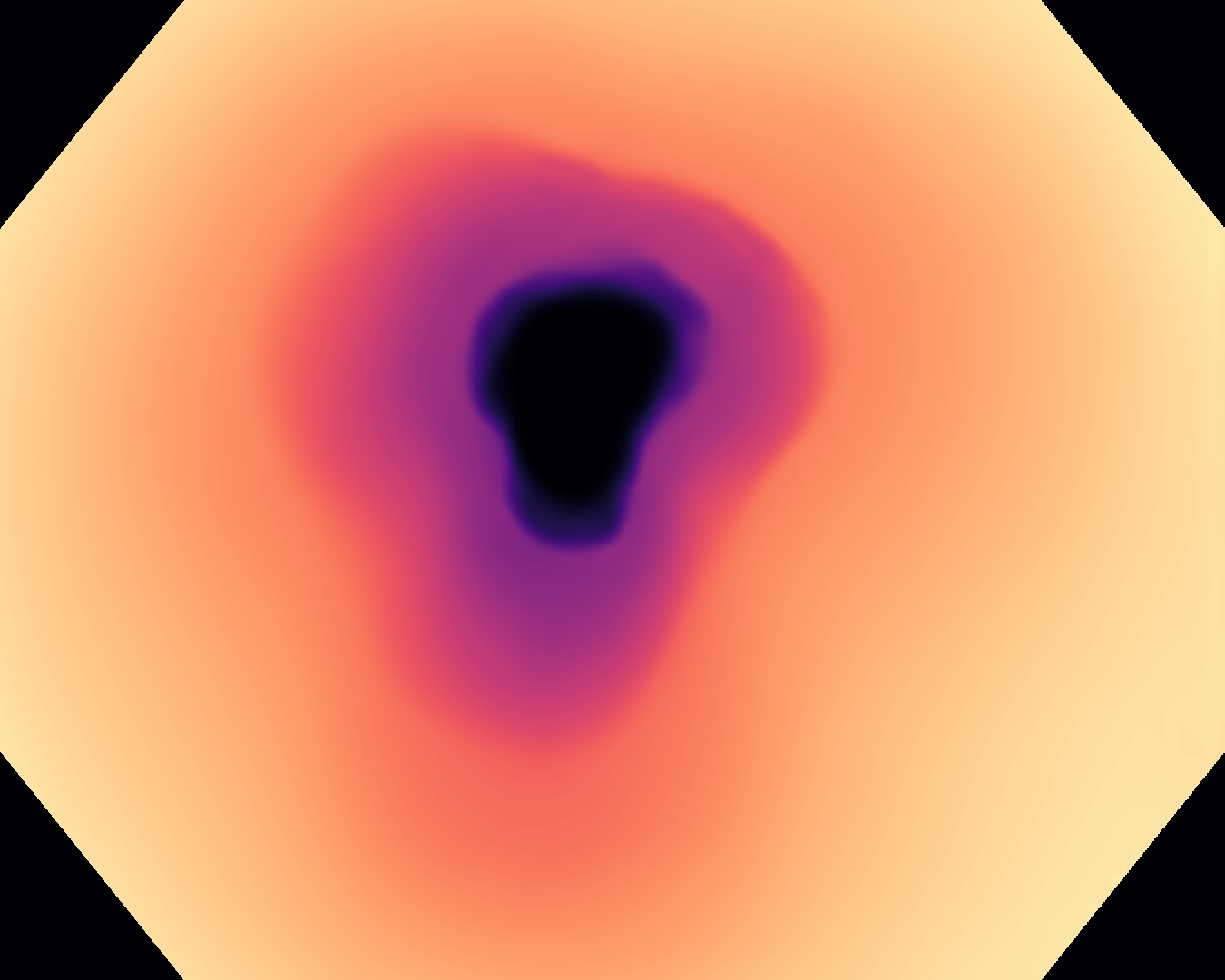} & \includegraphics[width=1.8cm]{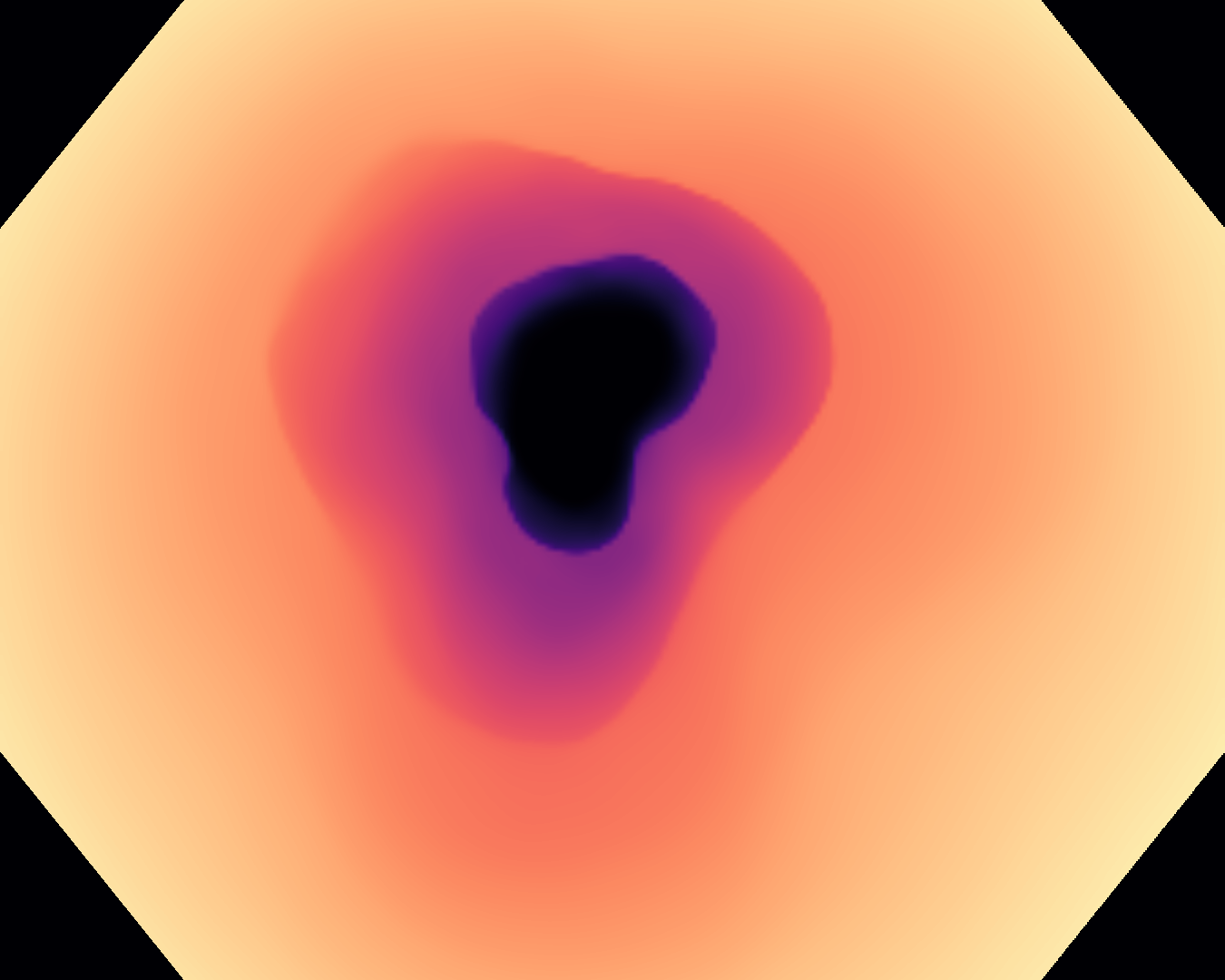} & \includegraphics[width=1.8cm]{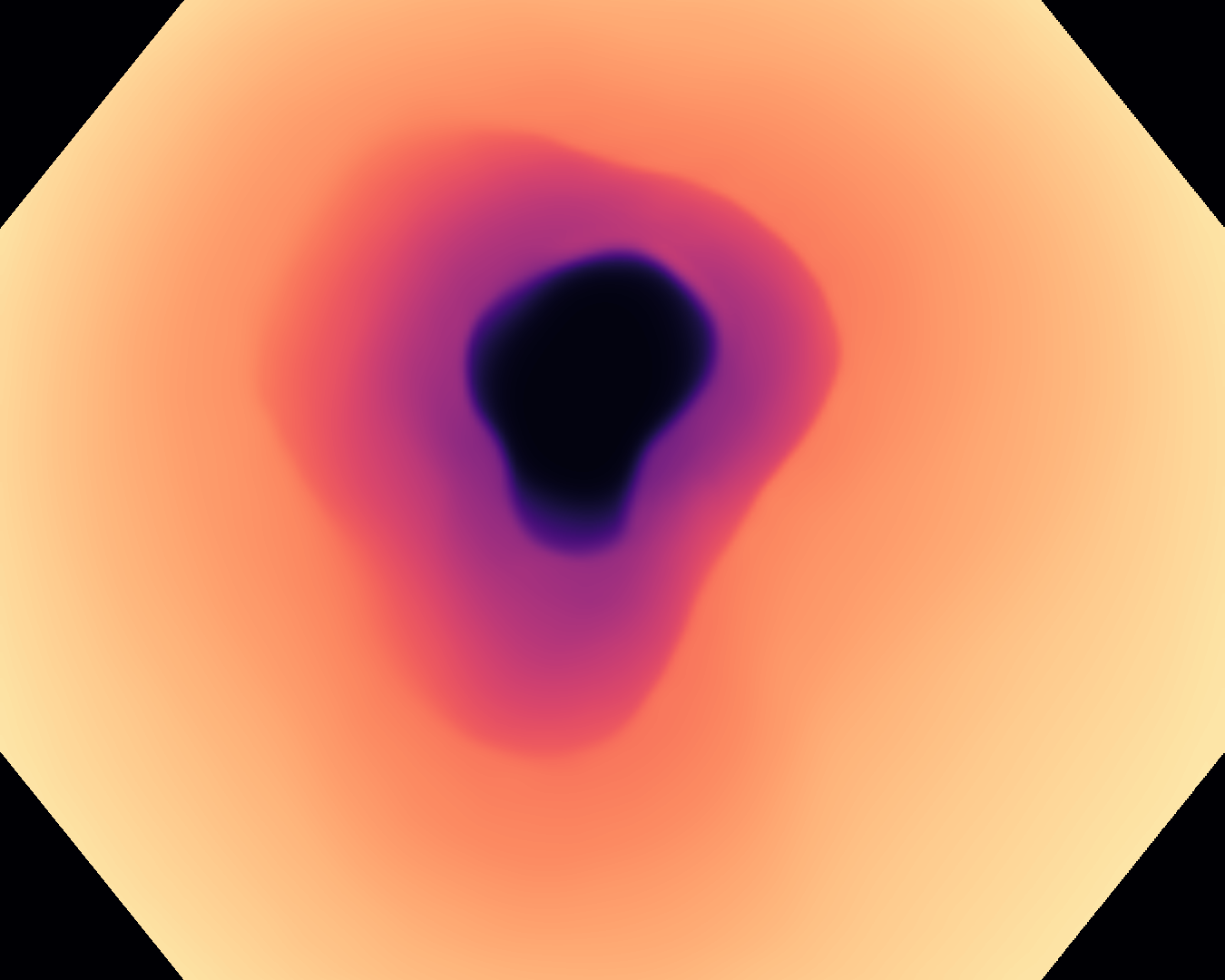} \\
   \includegraphics[width=1.8cm]{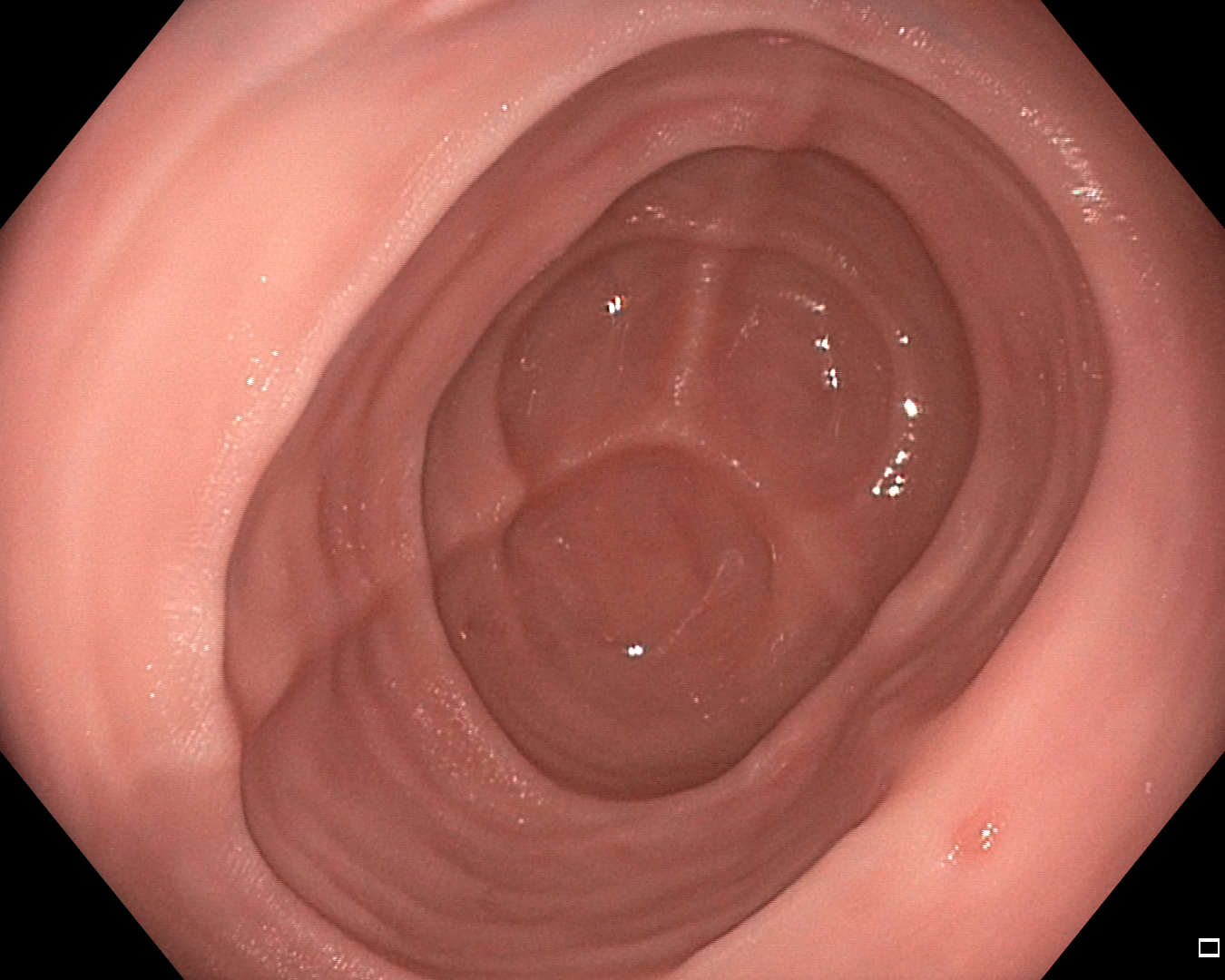}& \includegraphics[width=1.8cm]{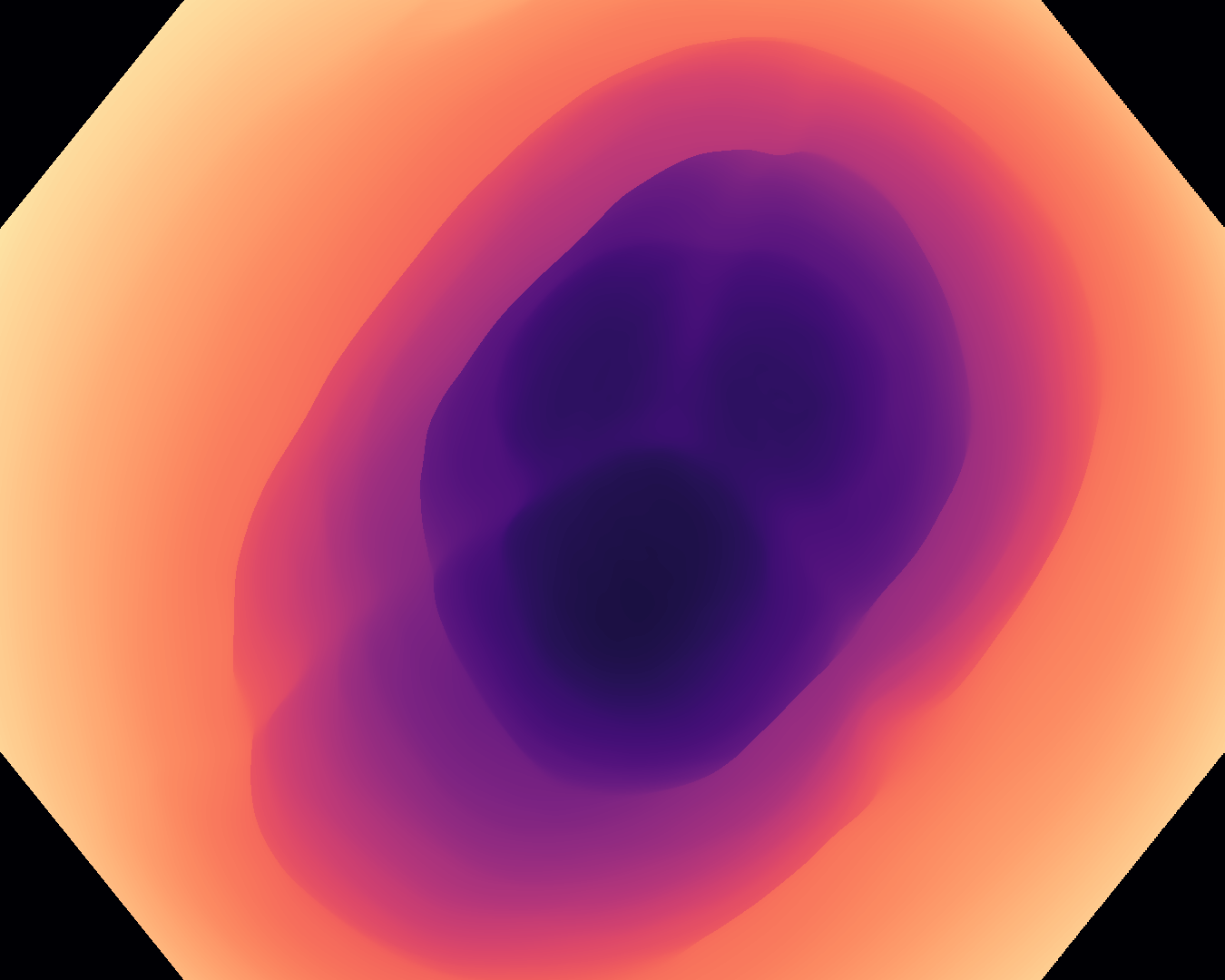}  & \includegraphics[width=1.8cm]{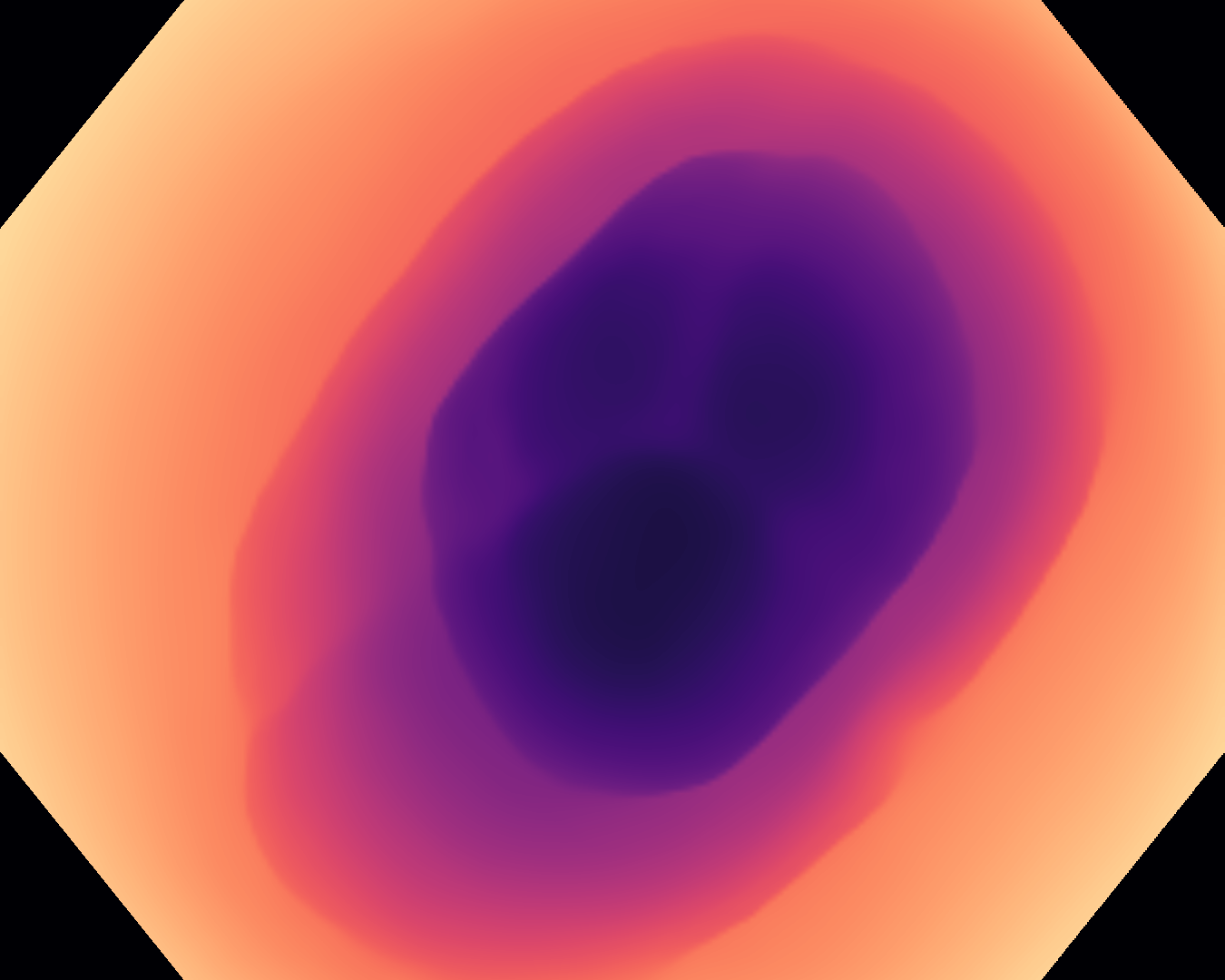} & \includegraphics[width=1.8cm]{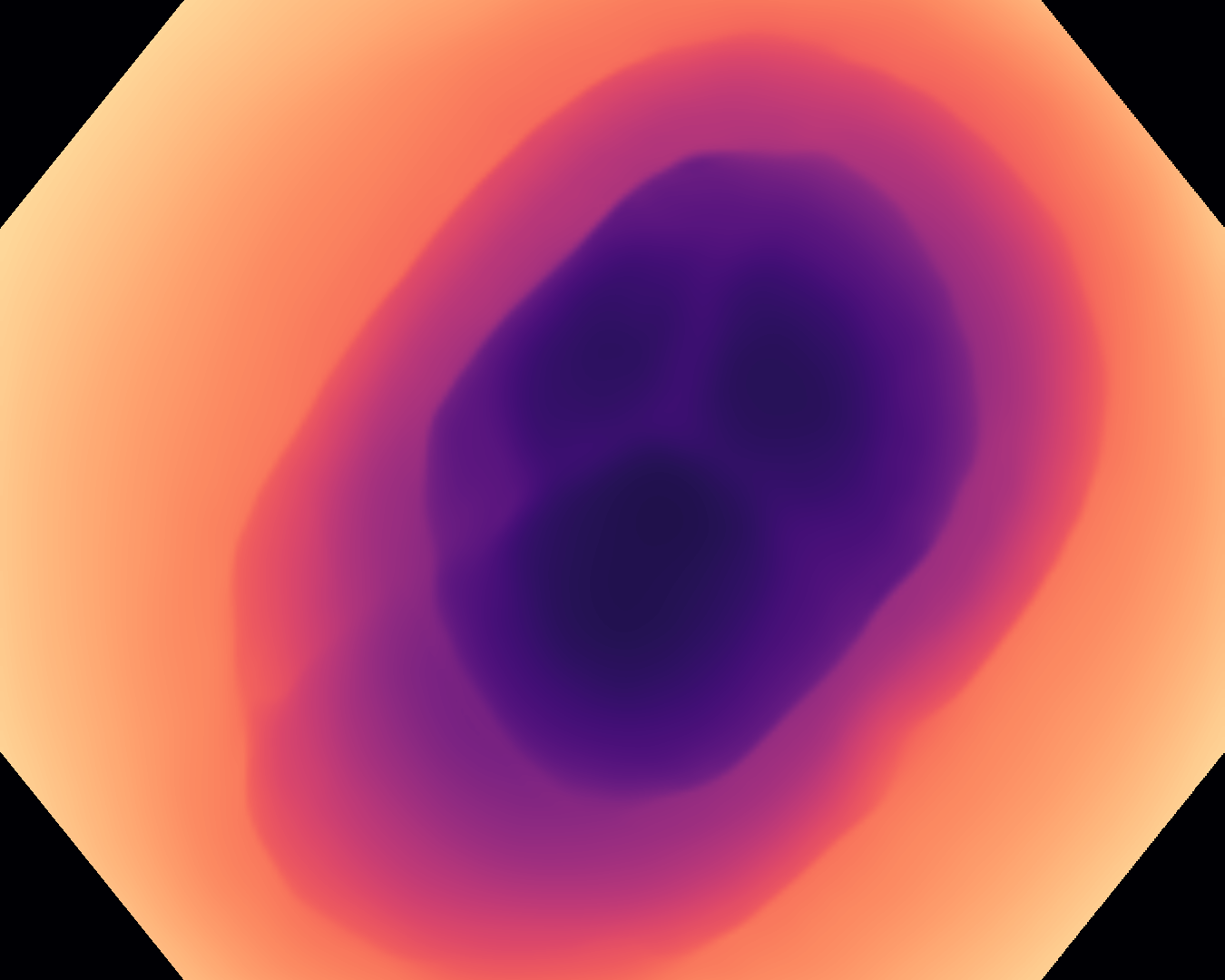} & \includegraphics[width=1.8cm]{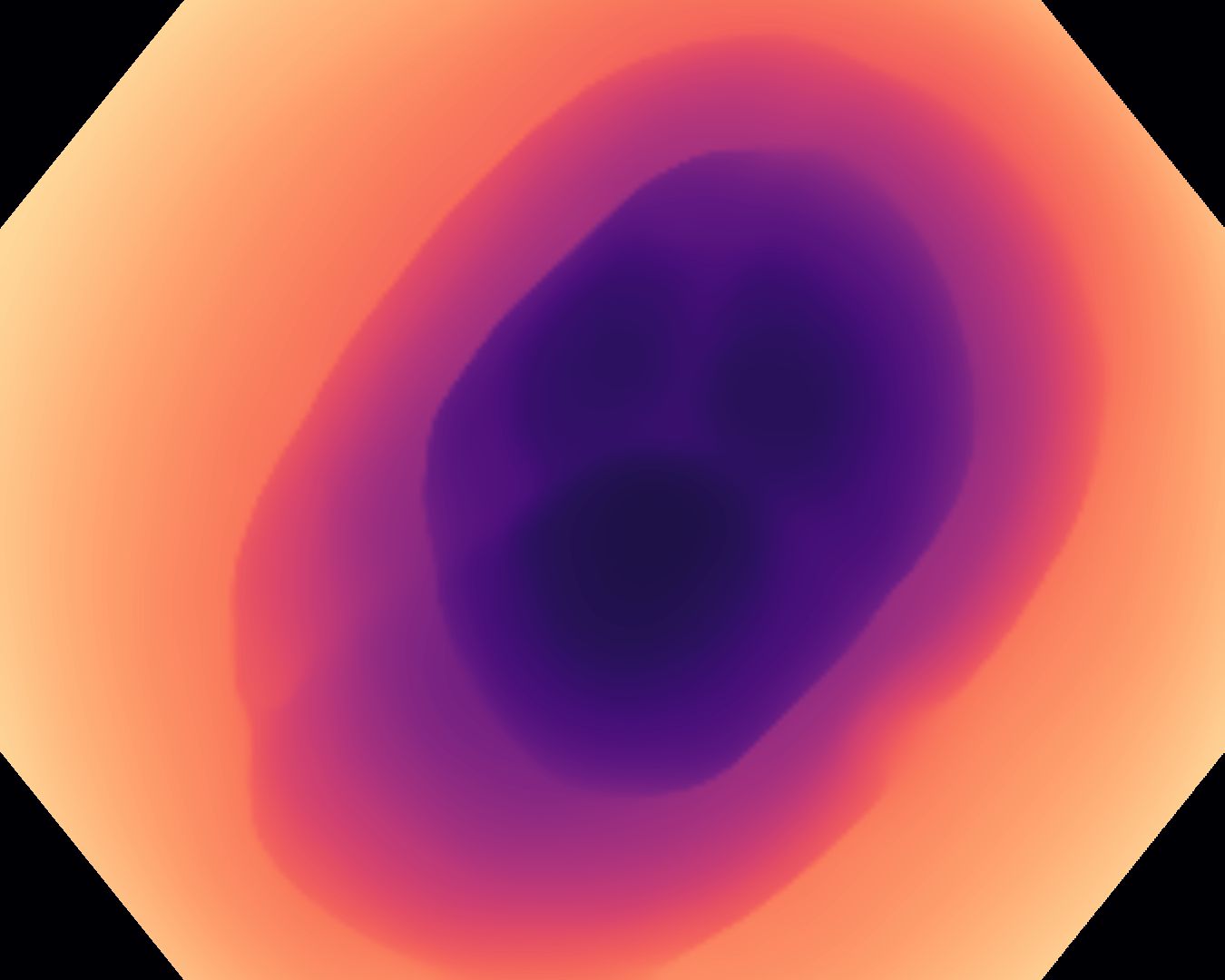} & \includegraphics[width=1.8cm]{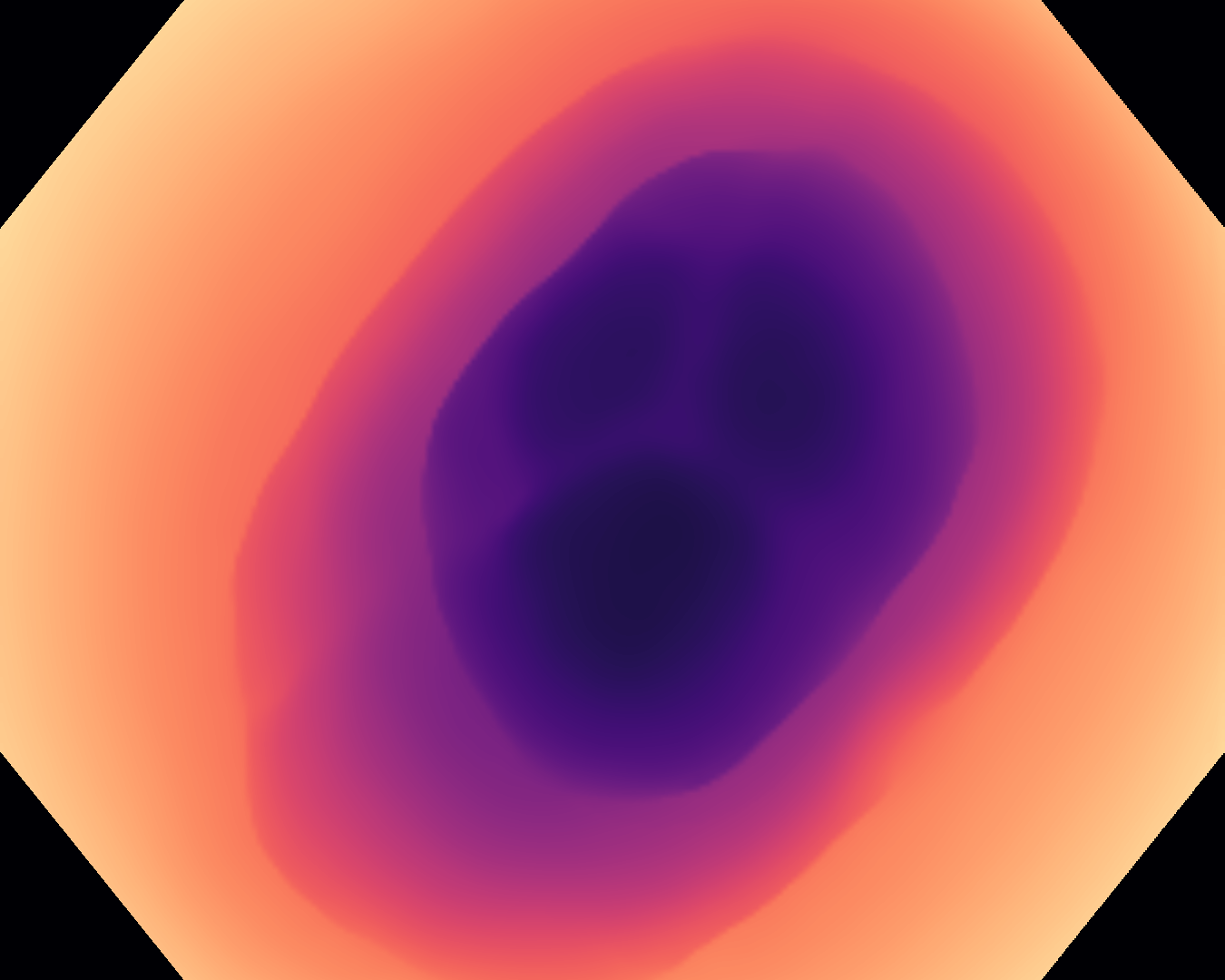} & \includegraphics[width=1.8cm]{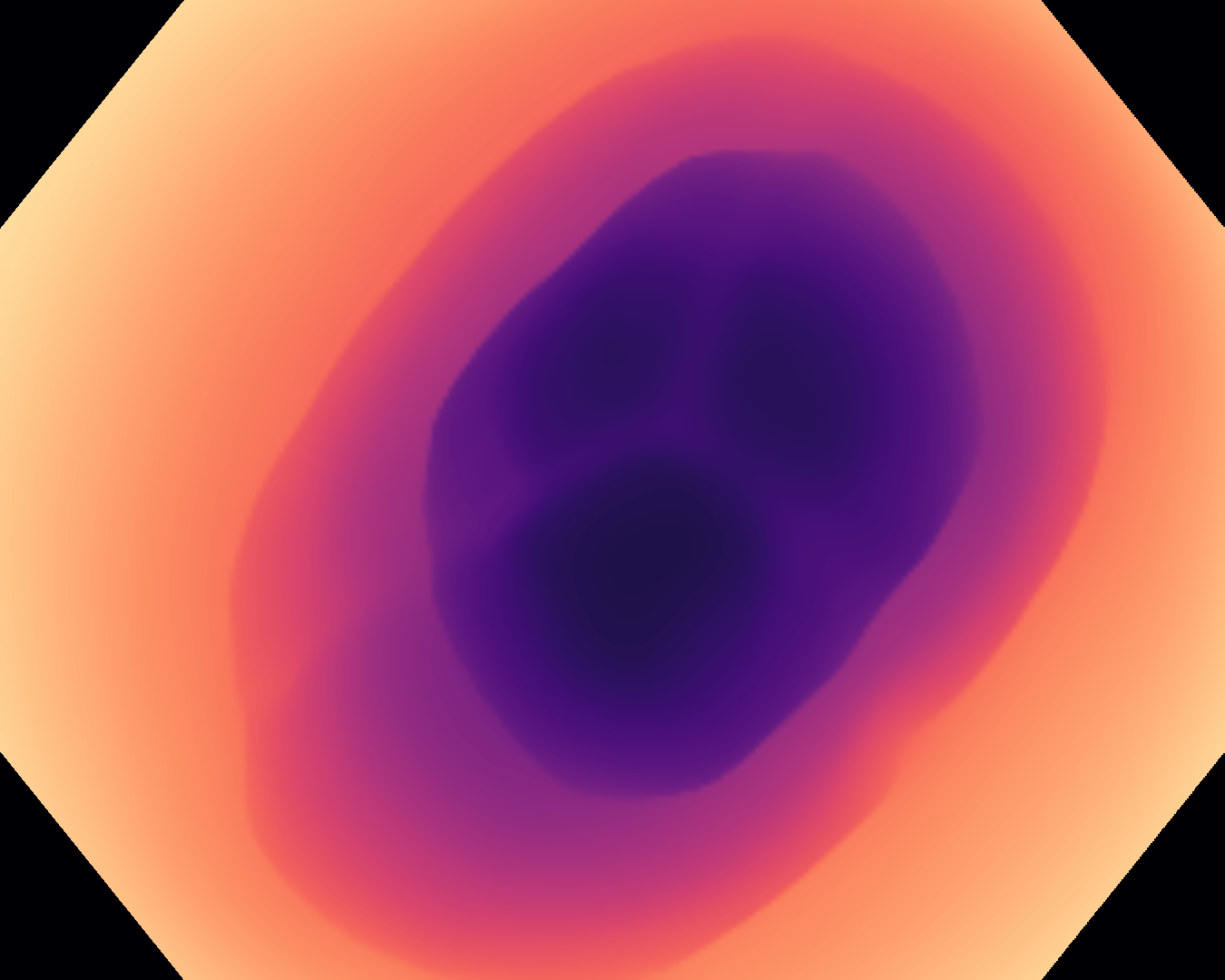} & \includegraphics[width=1.8cm]{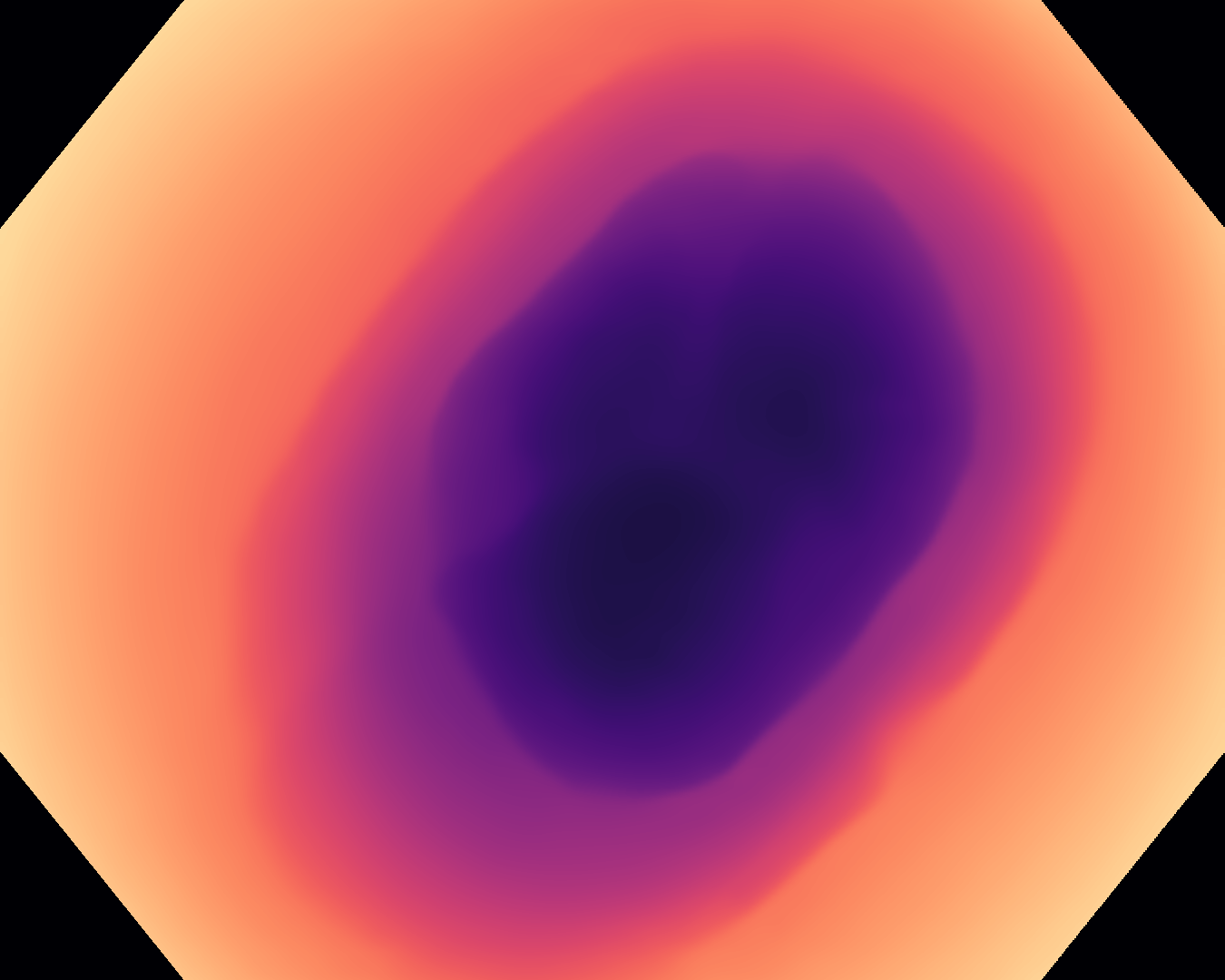} \\

  Input & Target & VT-HK-MC & VT-HK-MA  & VT-IN-MC  & VT-IN-MA  & VT-IN-SL  & VT-NA-NA  \\ 
 \includegraphics[width=1.8cm]{11_color.png}& \includegraphics[width=1.8cm]{GT11.png} & 
 \includegraphics[width=1.8cm]{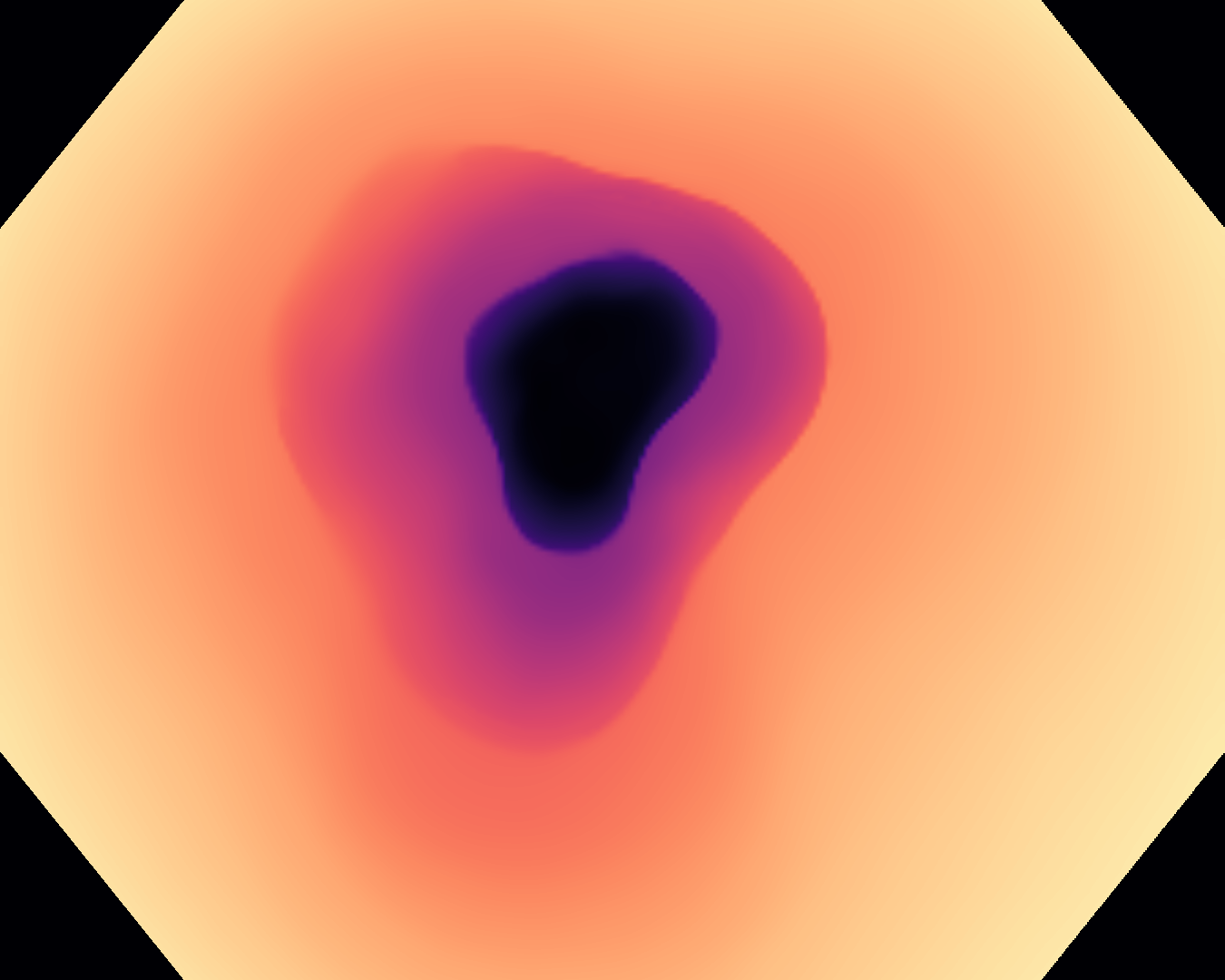}& \includegraphics[width=1.8cm]{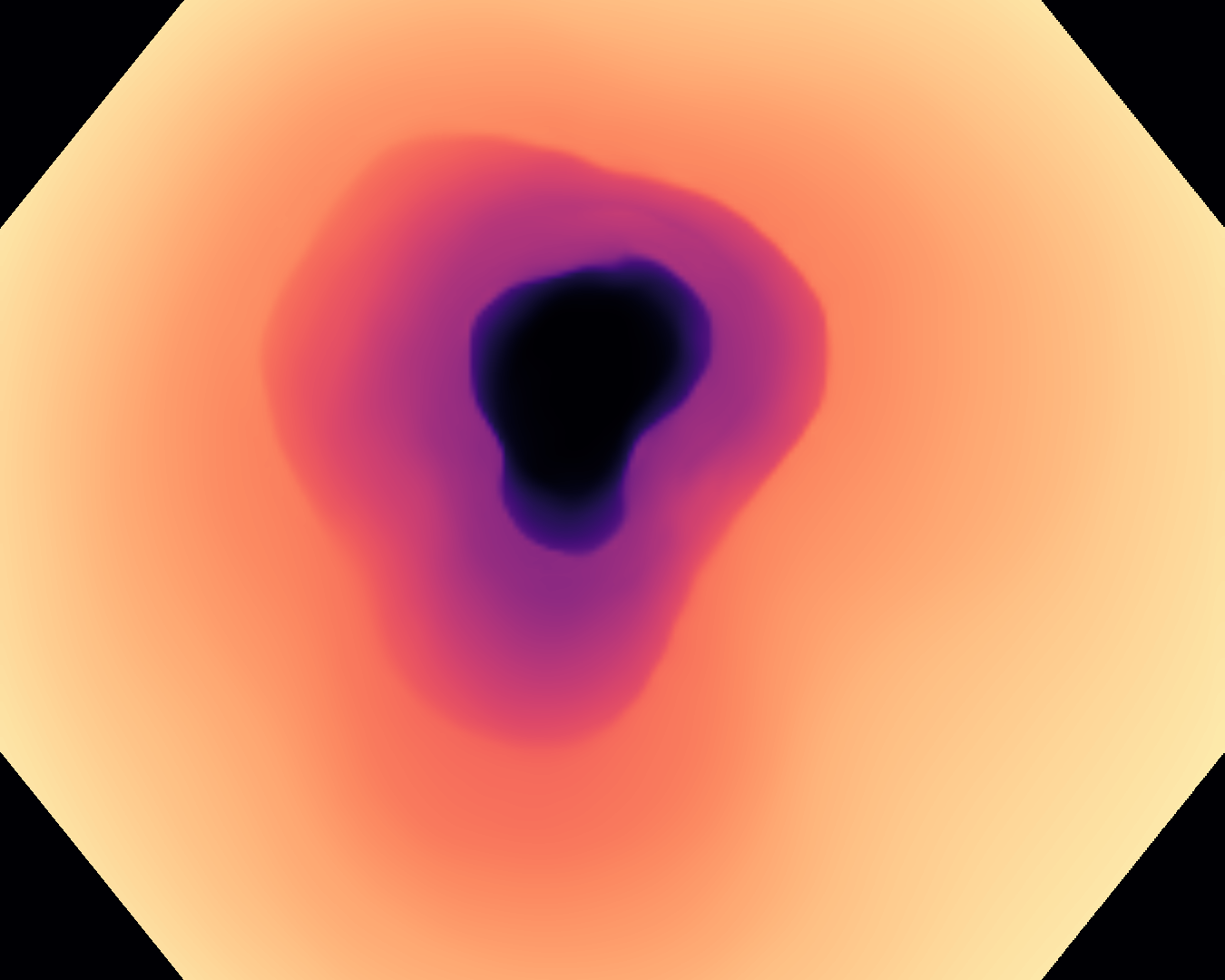} & \includegraphics[width=1.8cm]{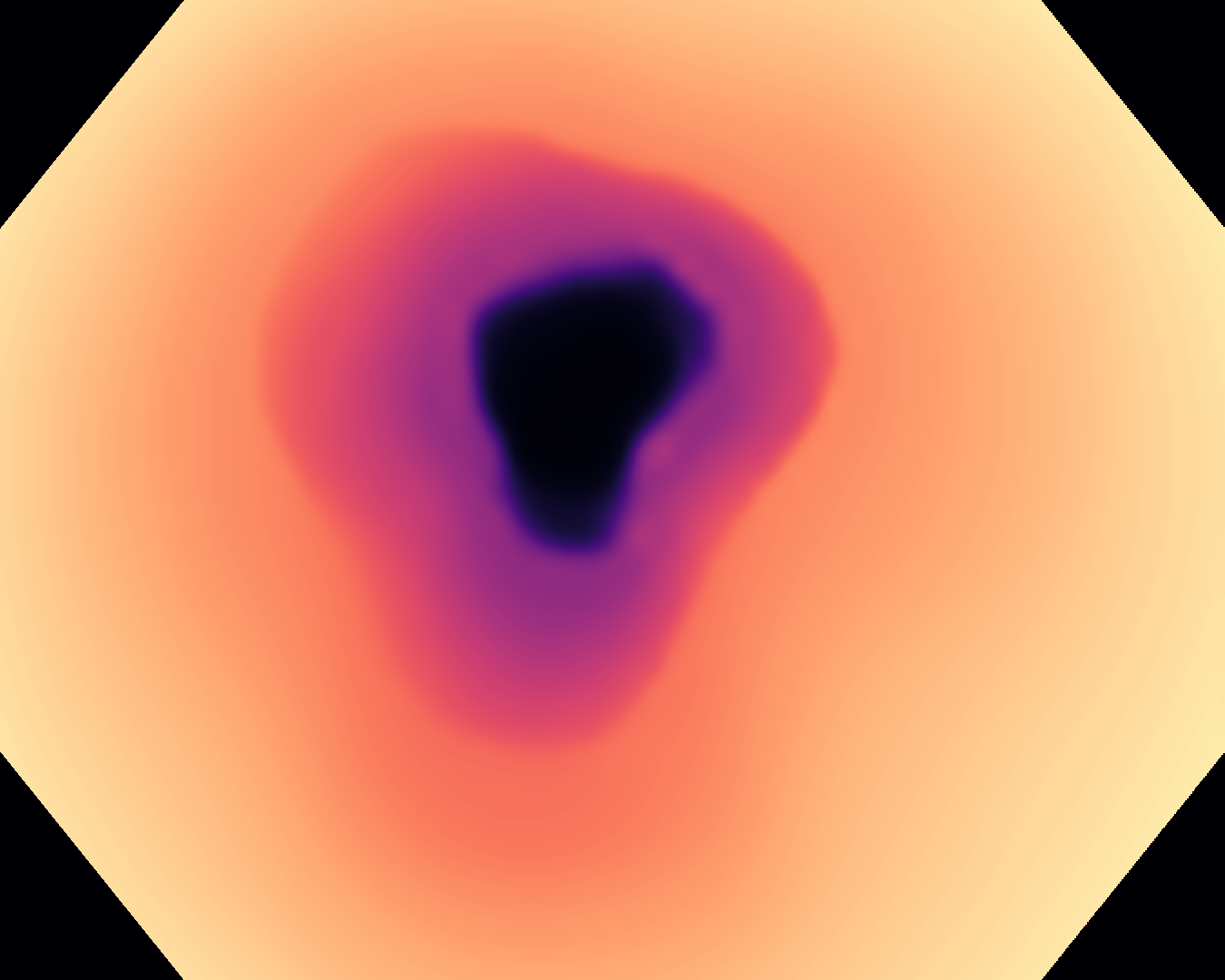} & \includegraphics[width=1.8cm]{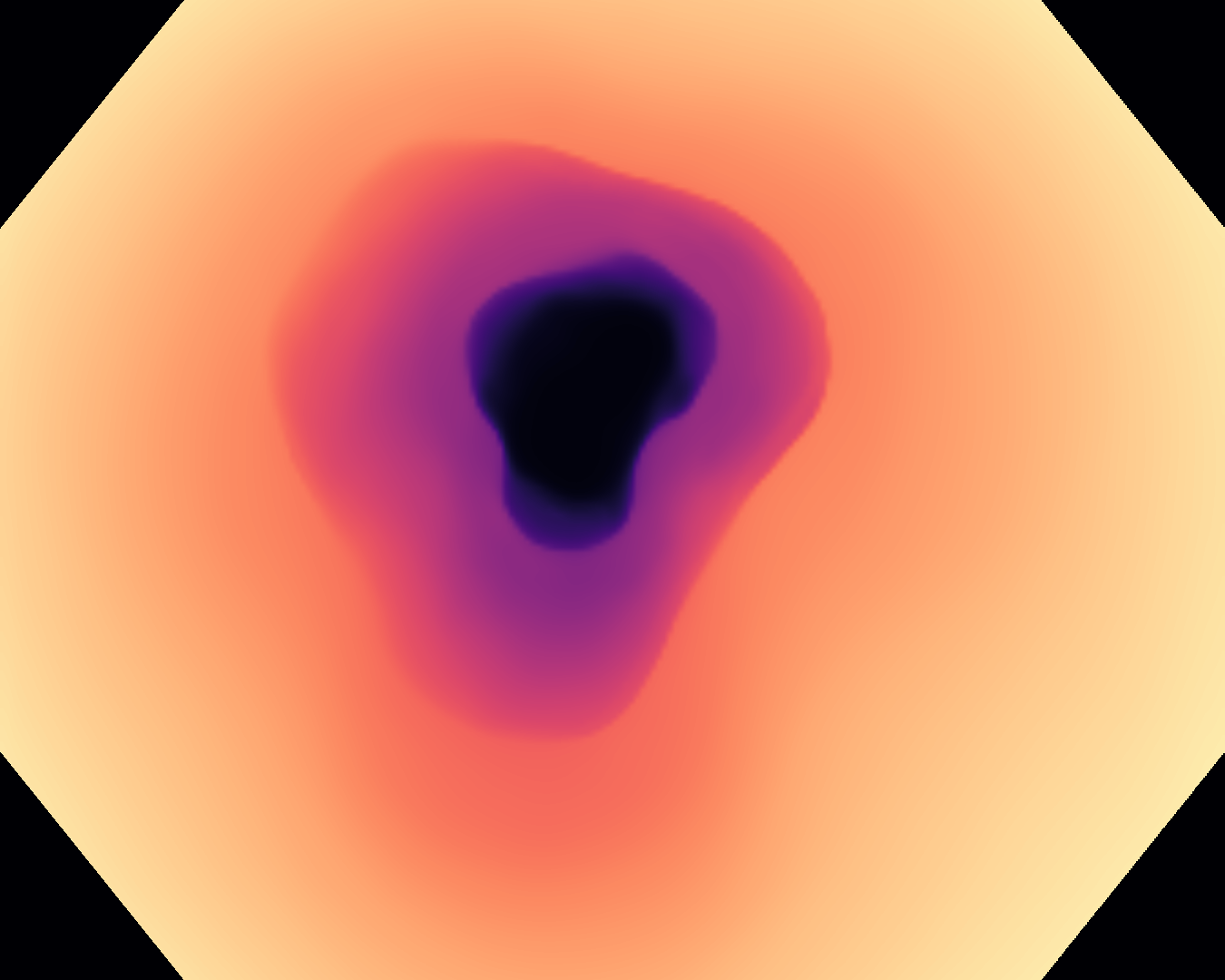} & \includegraphics[width=1.8cm]{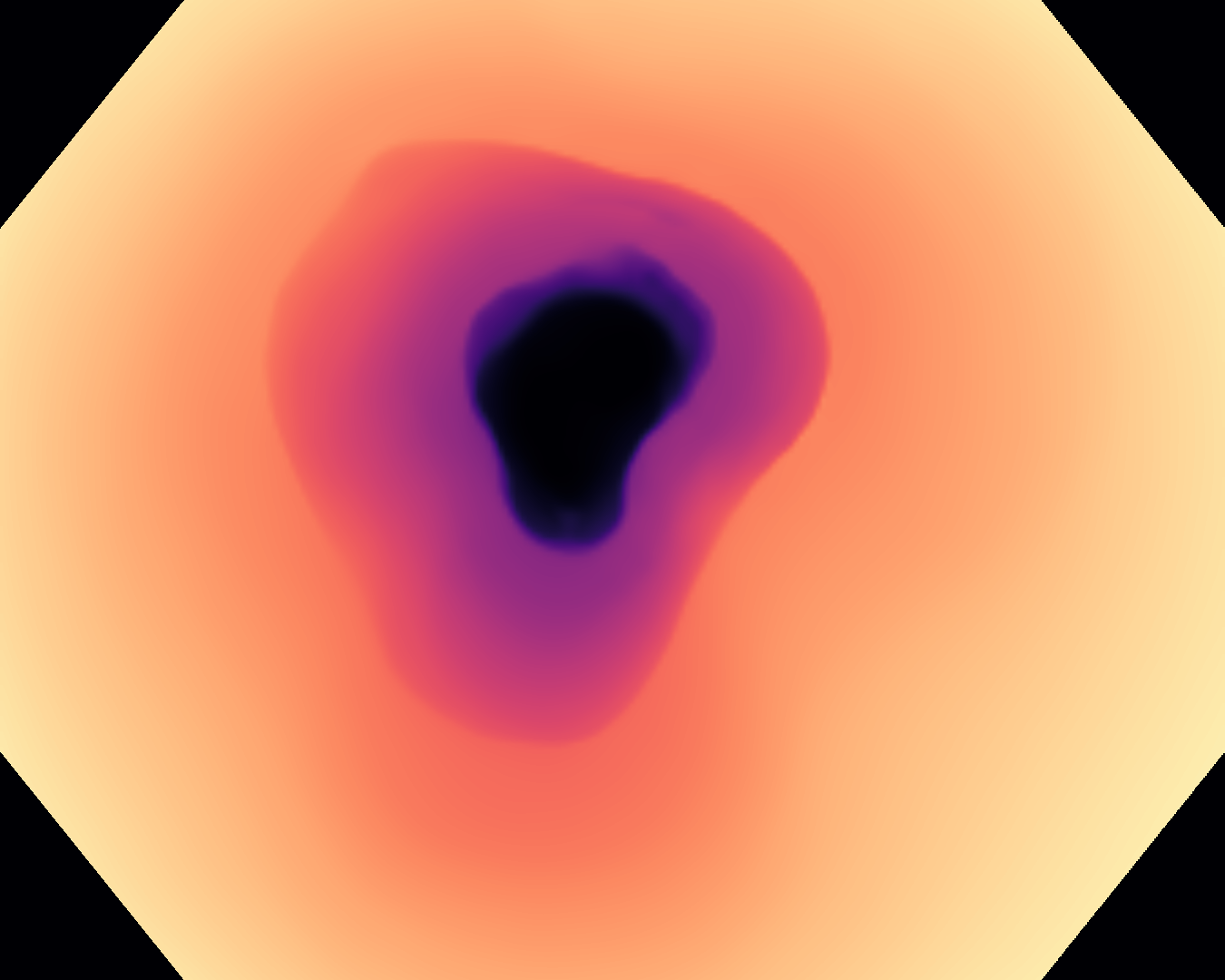} & \includegraphics[width=1.8cm]{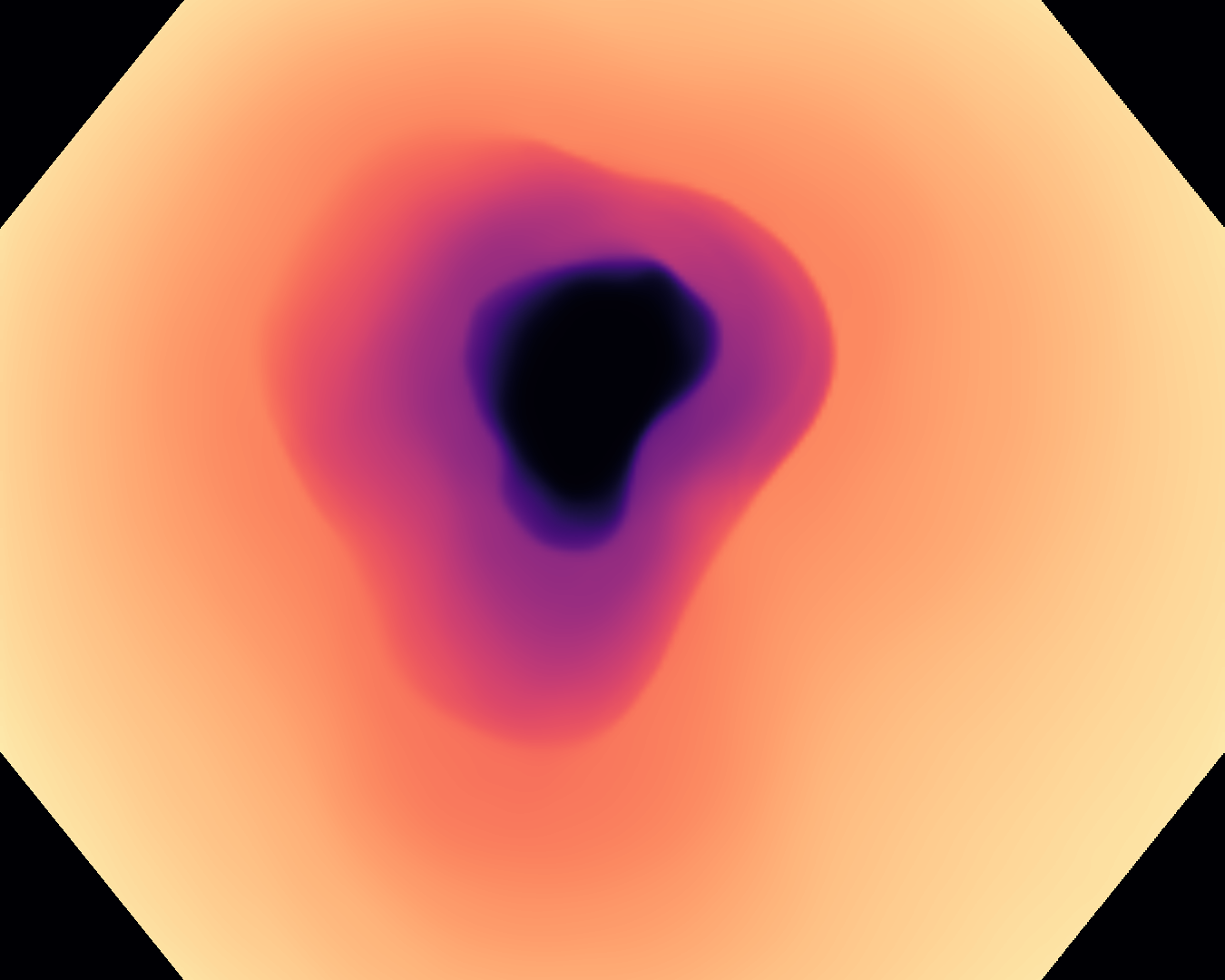} \\
   \includegraphics[width=1.8cm]{365_color.png}& \includegraphics[width=1.8cm]{GT468.png}  &\includegraphics[width=1.8cm]{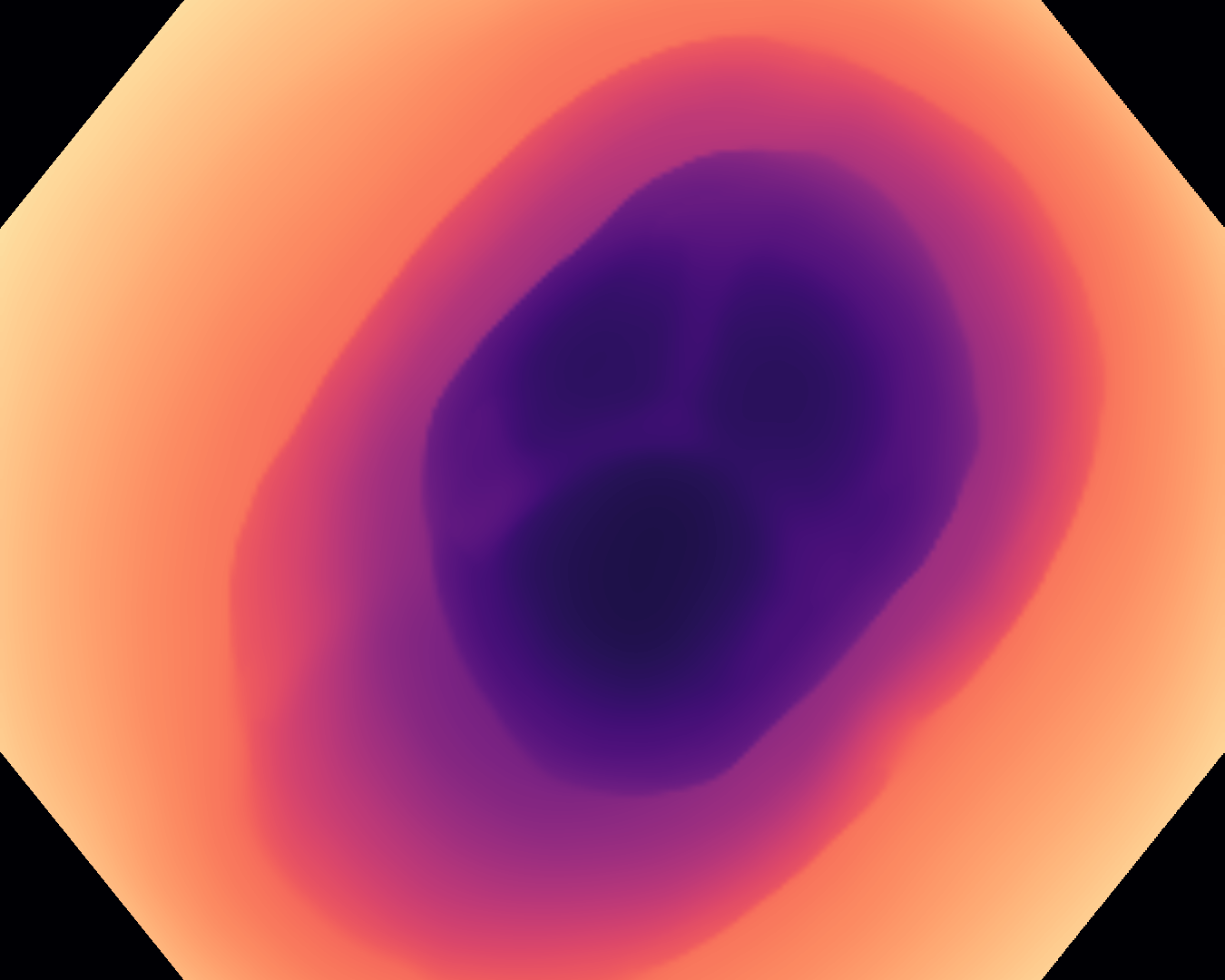} & \includegraphics[width=1.8cm]{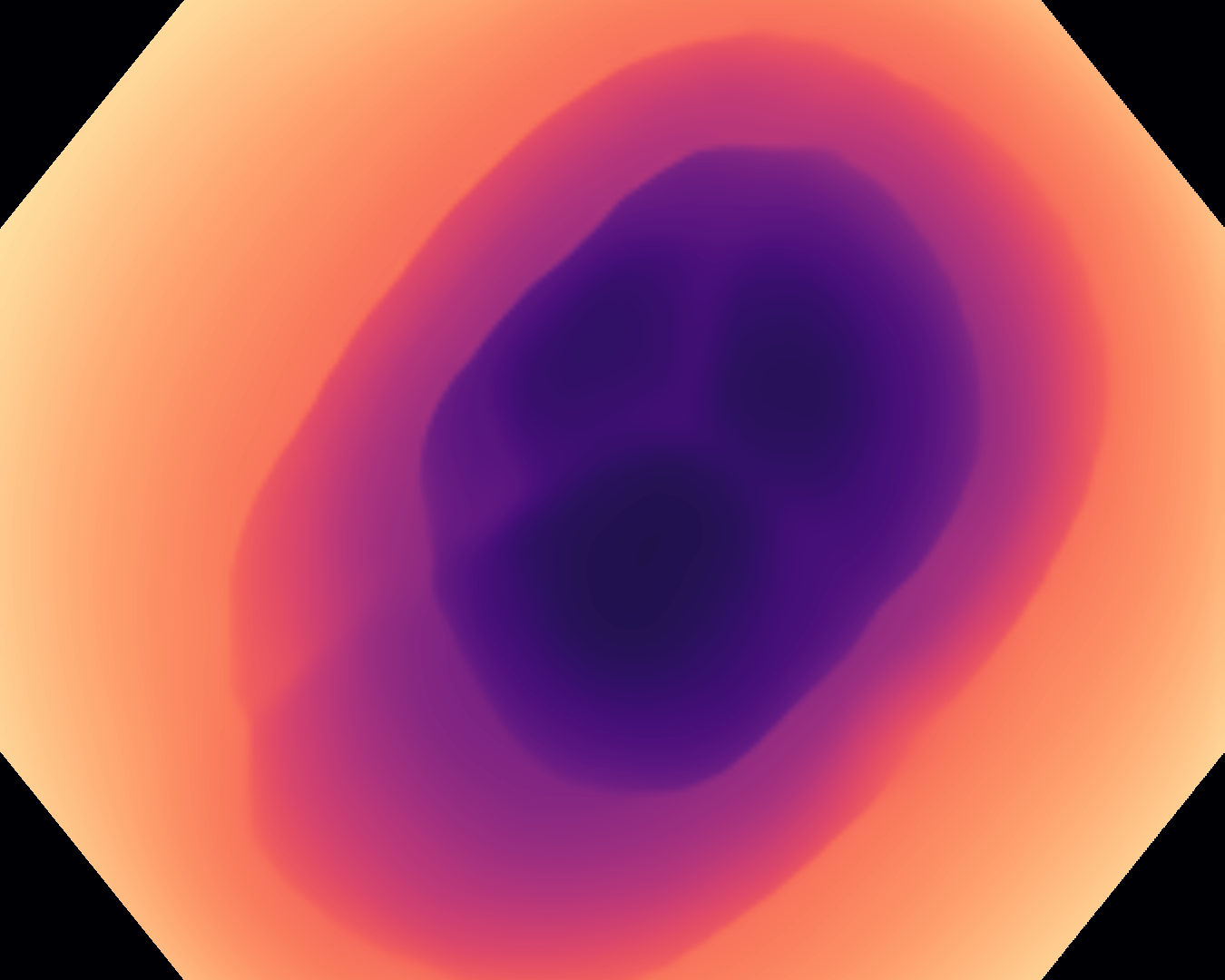} & \includegraphics[width=1.8cm]{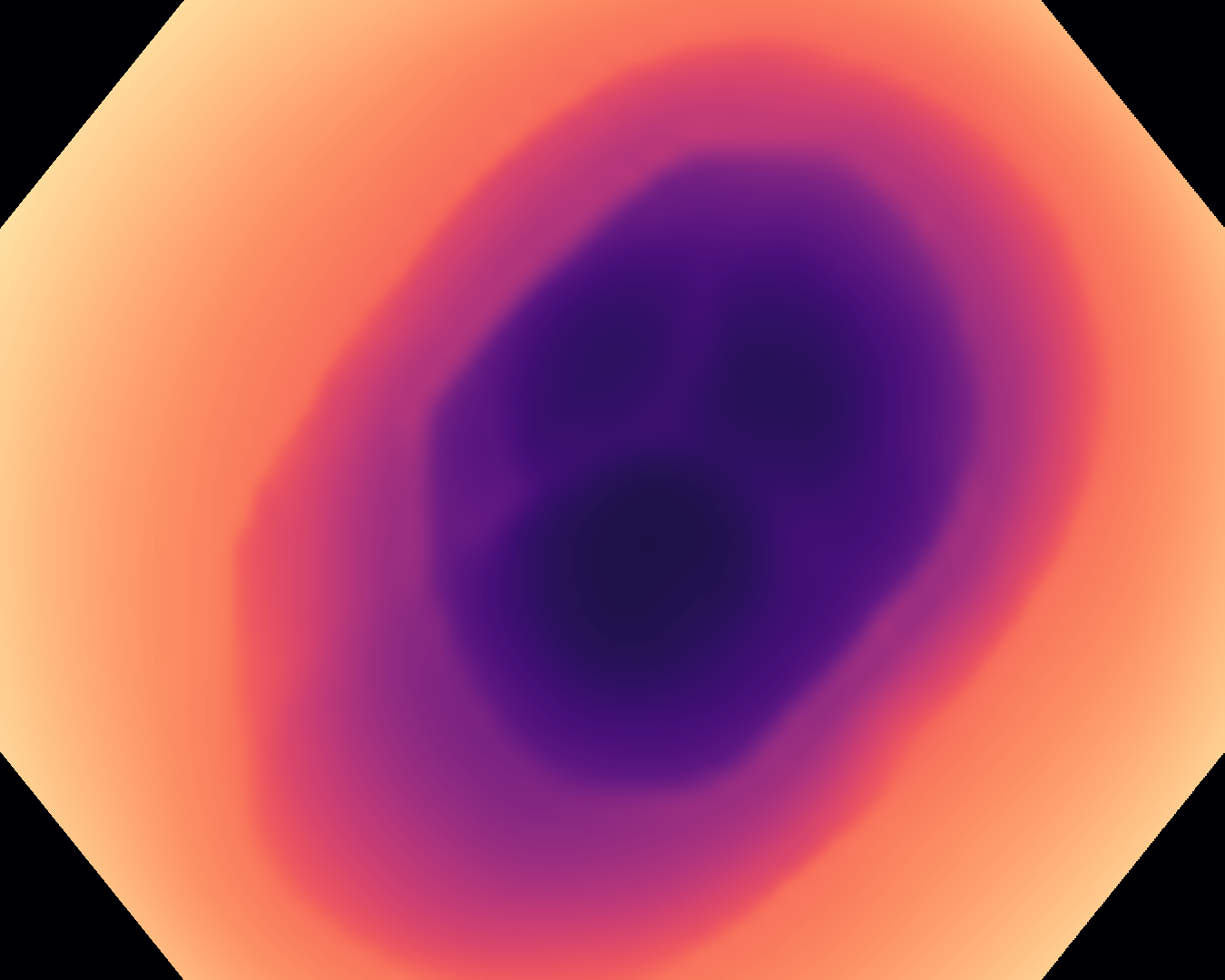} & \includegraphics[width=1.8cm]{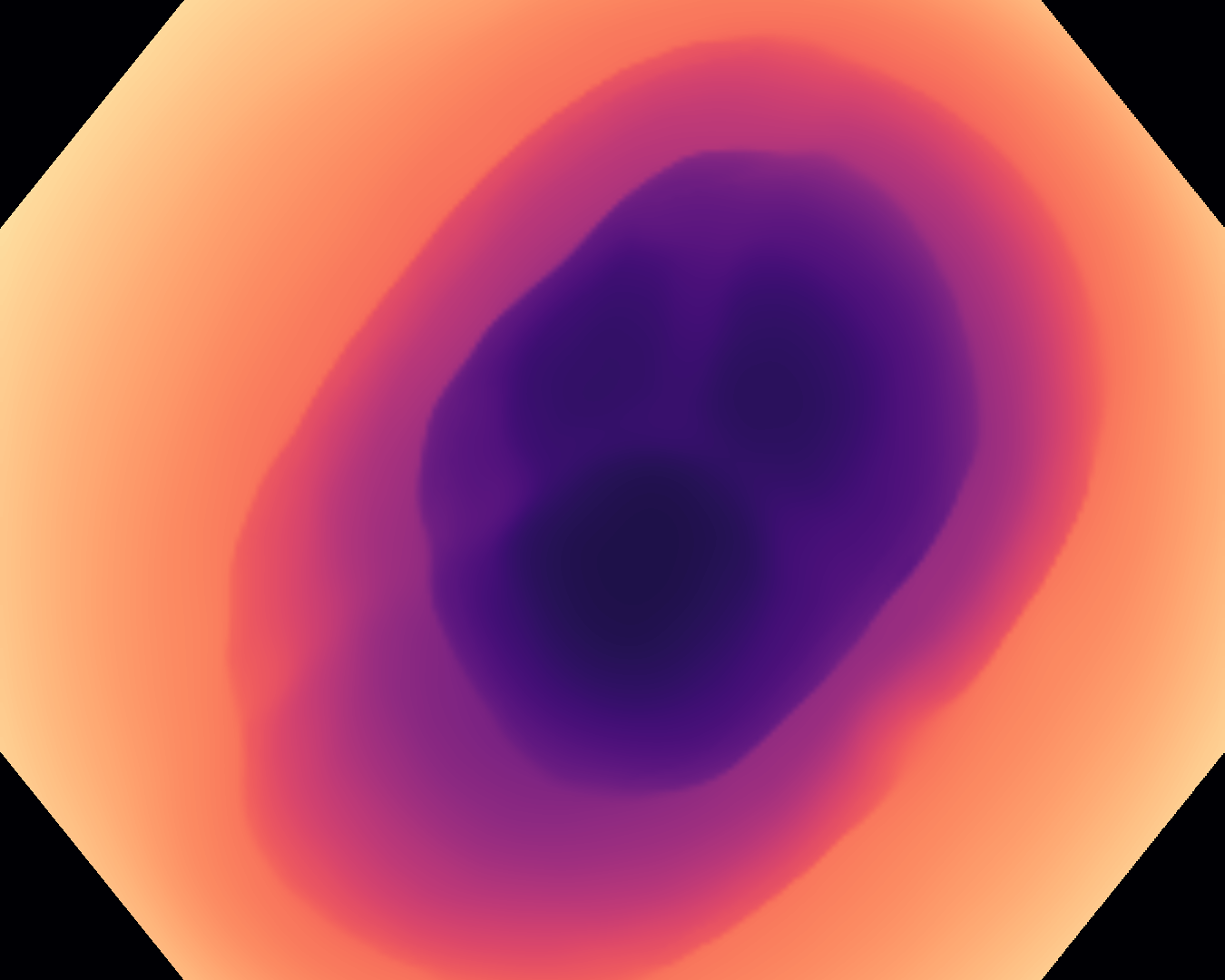} & \includegraphics[width=1.8cm]{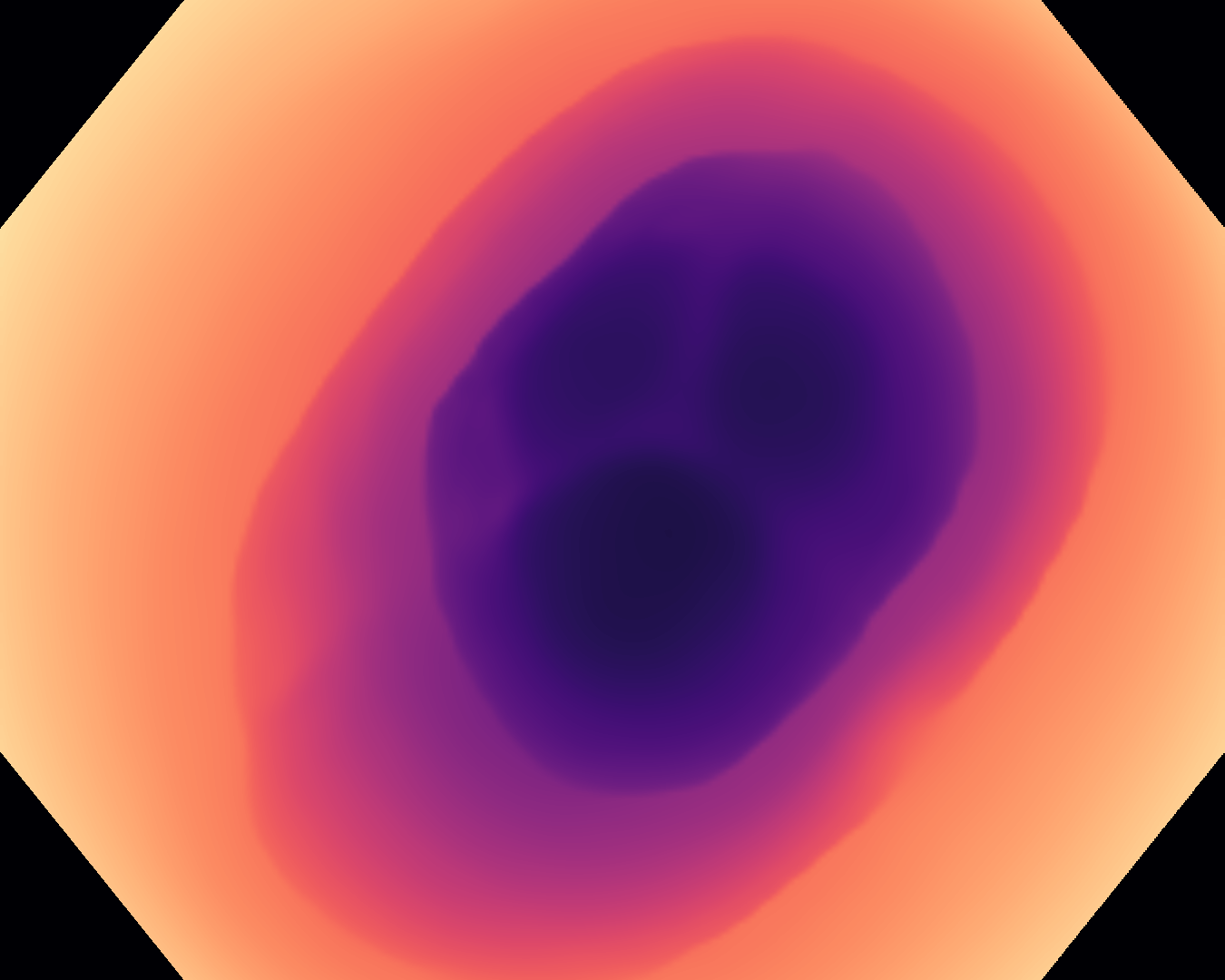}& \includegraphics[width=1.8cm]{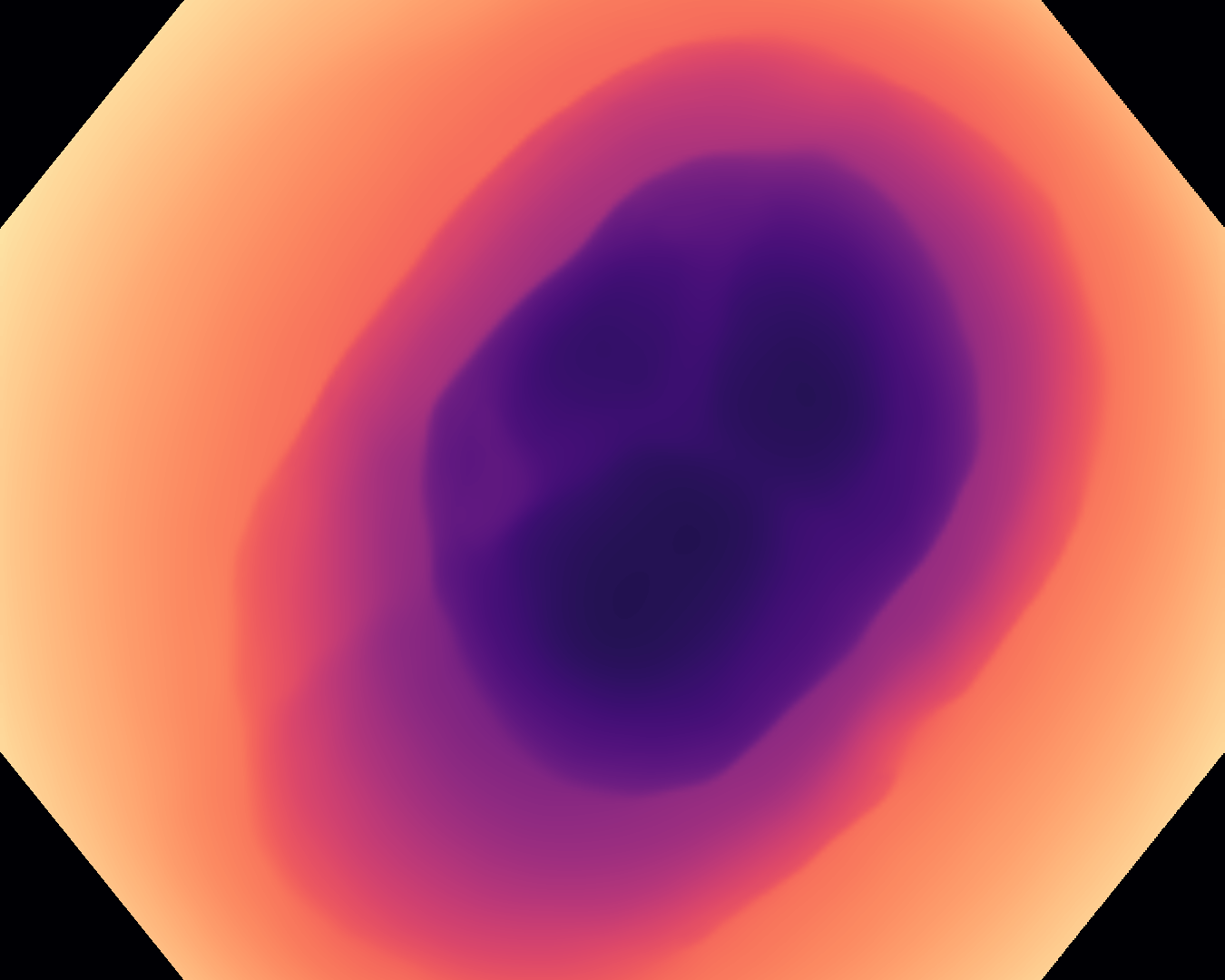}
\end{tabular}}
    \caption{Targets and post-processed predictions for a randomly selected instance from each of the test videos for C3VD. For conciseness, we denote ResNet50s with \textit{RN}, ViT-Bs with \textit{VT}, Hyperkvasir-unlabelled with \textit{HK}, ImageNet-1k with \textit{IN}, MoCo v3 with \textit{MC}, Barlow Twins with \textit{BT}, MAE with \textit{MA}, supervised pretraining with \textit{SL}, and no pretraining with \textit{NA-NA}.}
    \label{fig:example_depth}
\end{figure*}

\begin{figure*}[ht]
\makebox[\textwidth][c]{\begin{tabular}{ccccccc}
  Input & RN-HK-MC & RN-HK-BT  & RN-IN-MC  & RN-IN-BT  & RN-IN-SL  & RN-NA-NA  \\ 
 \includegraphics[width=2cm]{11_color.png} &
 \includegraphics[width=2cm]{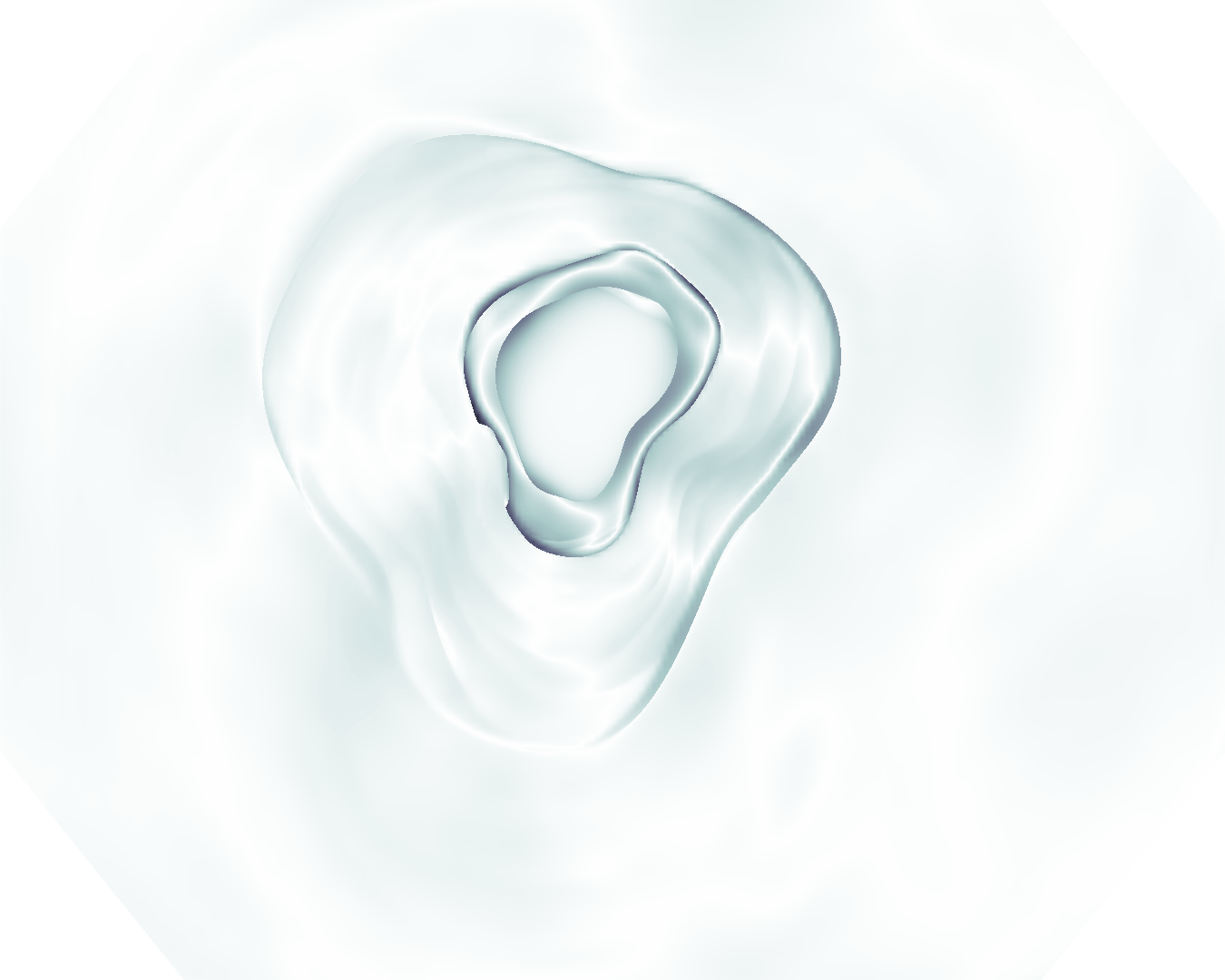}& \includegraphics[width=2cm]{diff11_resnet50-Hyperkvasir_barlowtwins_init-frozen_False.png} & \includegraphics[width=2cm]{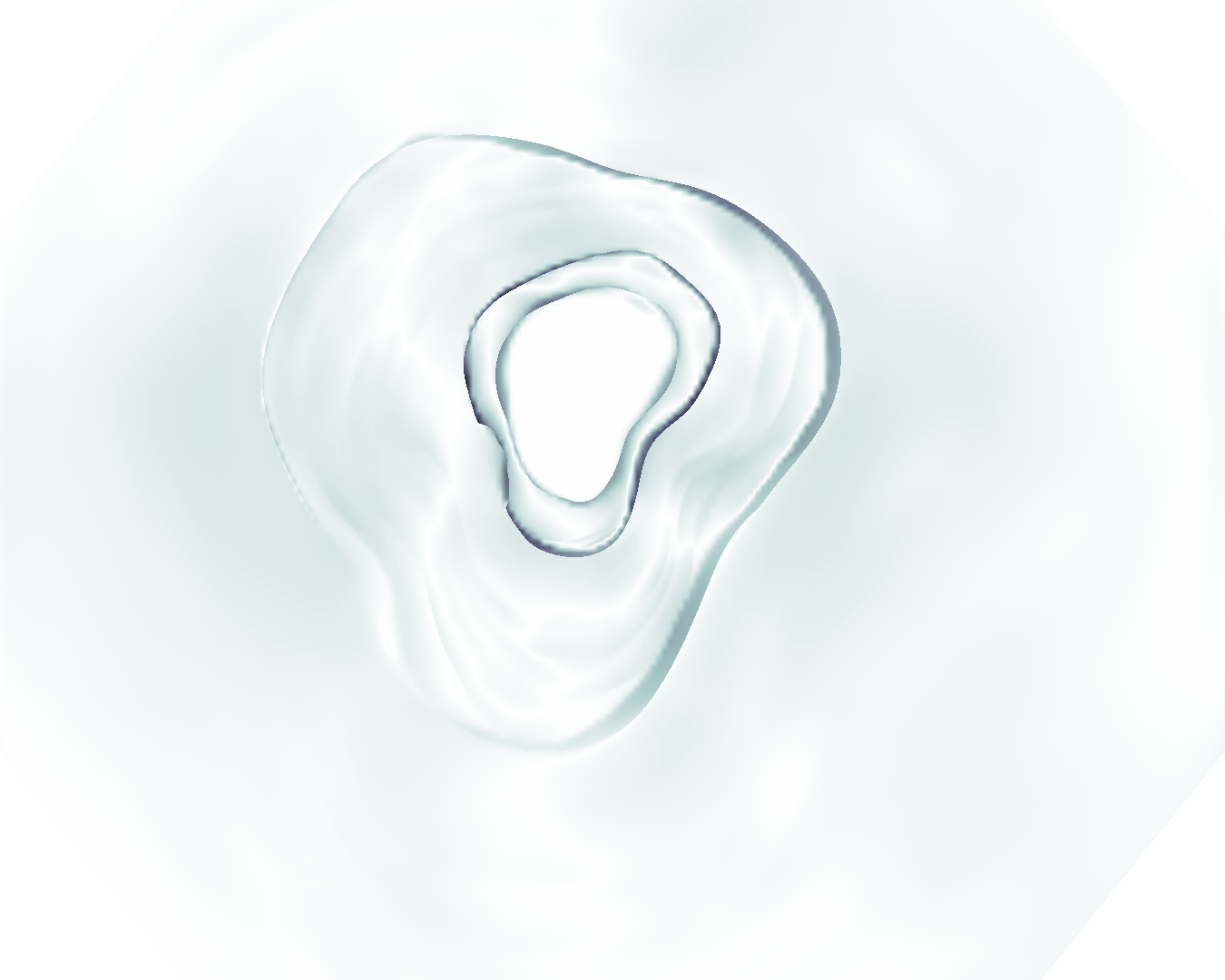} & \includegraphics[width=2cm]{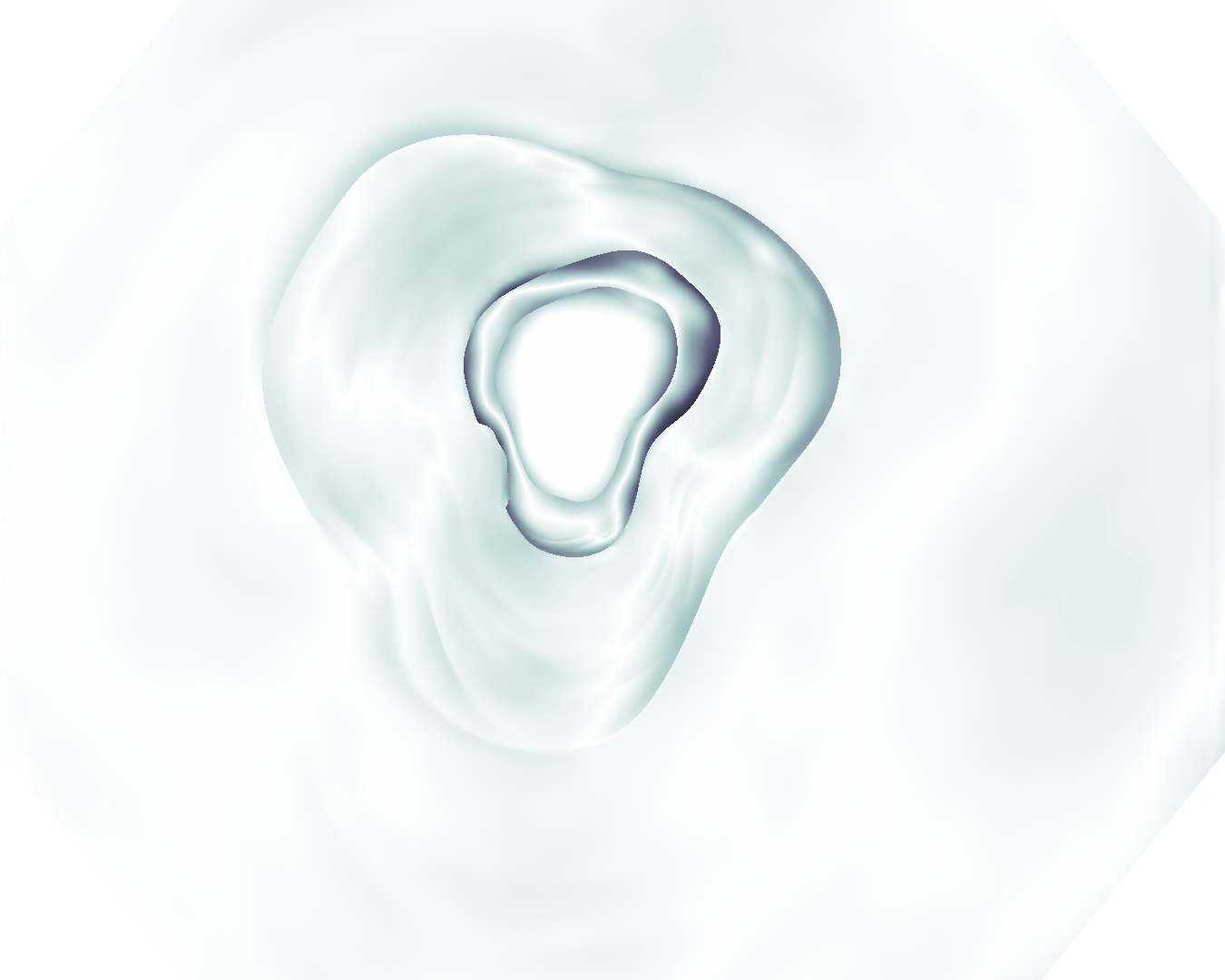} & \includegraphics[width=2cm]{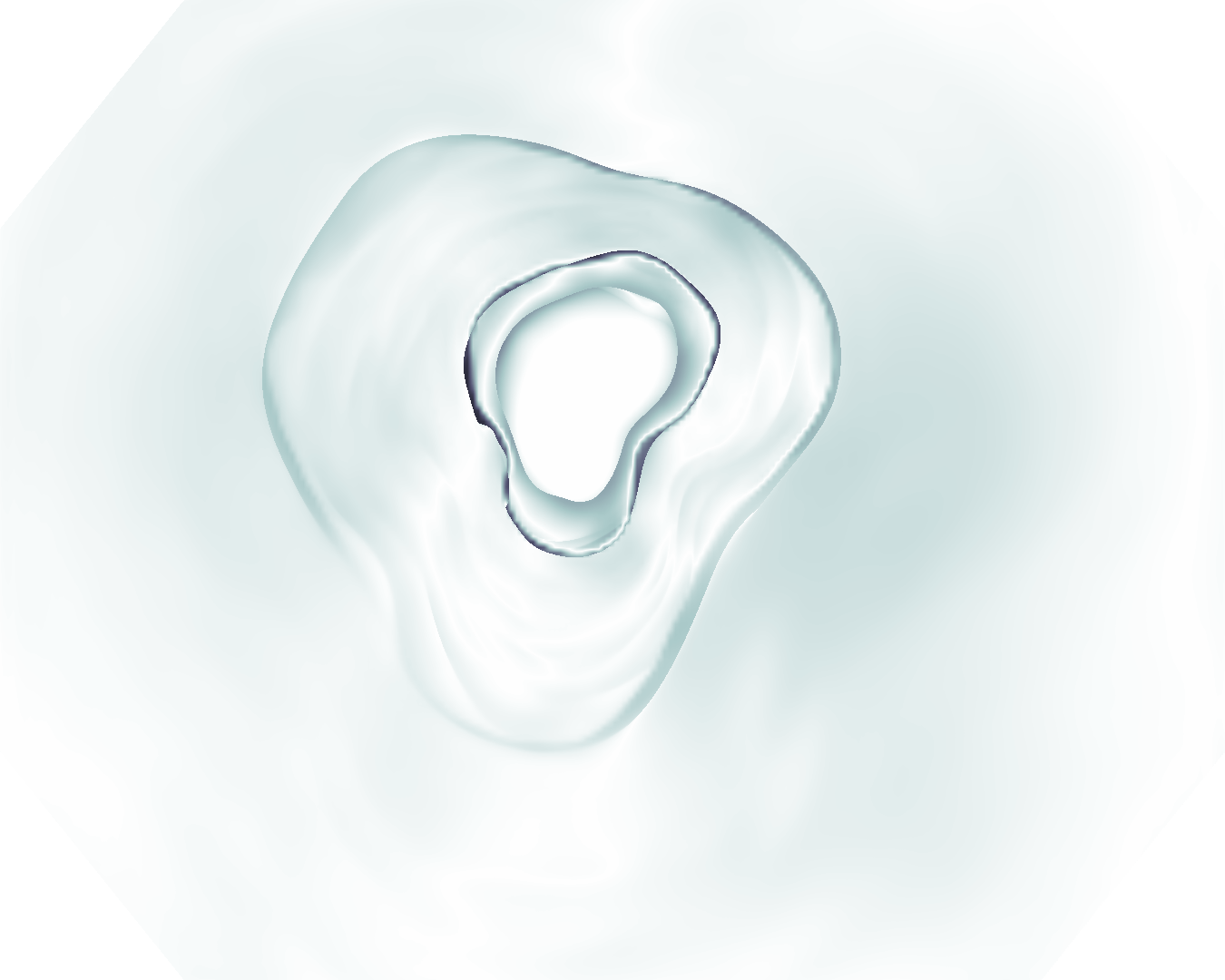} & \includegraphics[width=2cm]{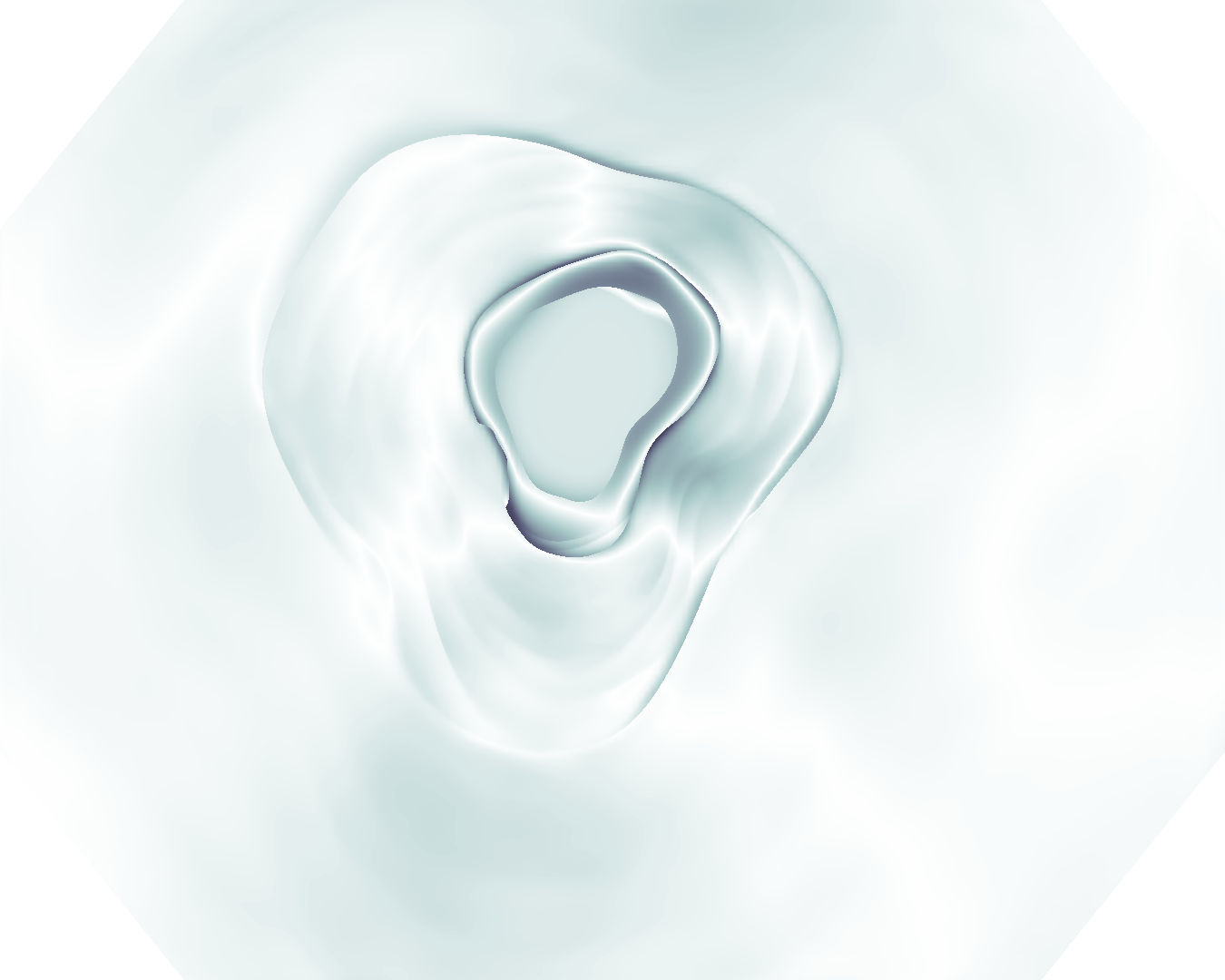} \\
   \includegraphics[width=2cm]{365_color.png}  & \includegraphics[width=2cm]{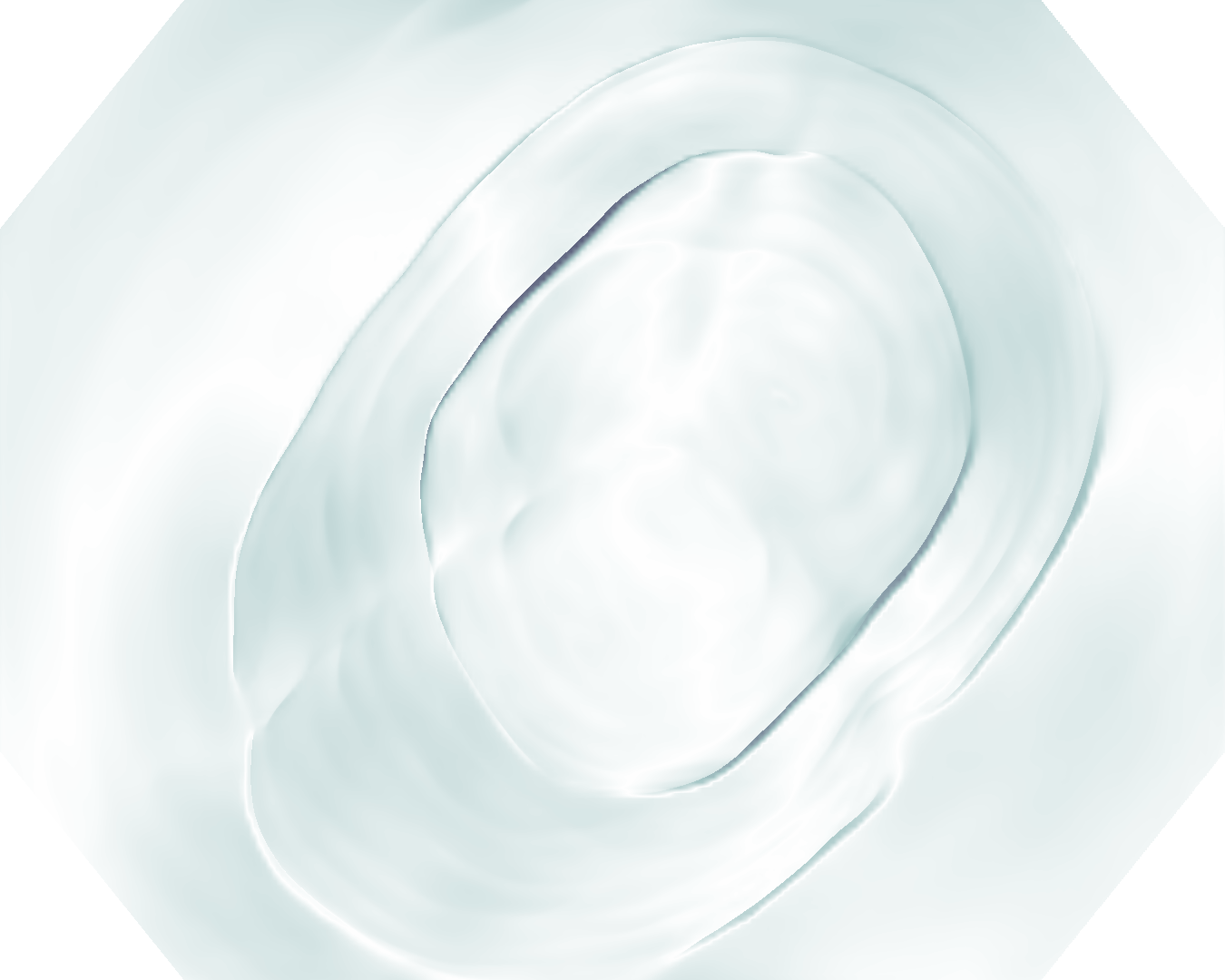} & \includegraphics[width=2cm]{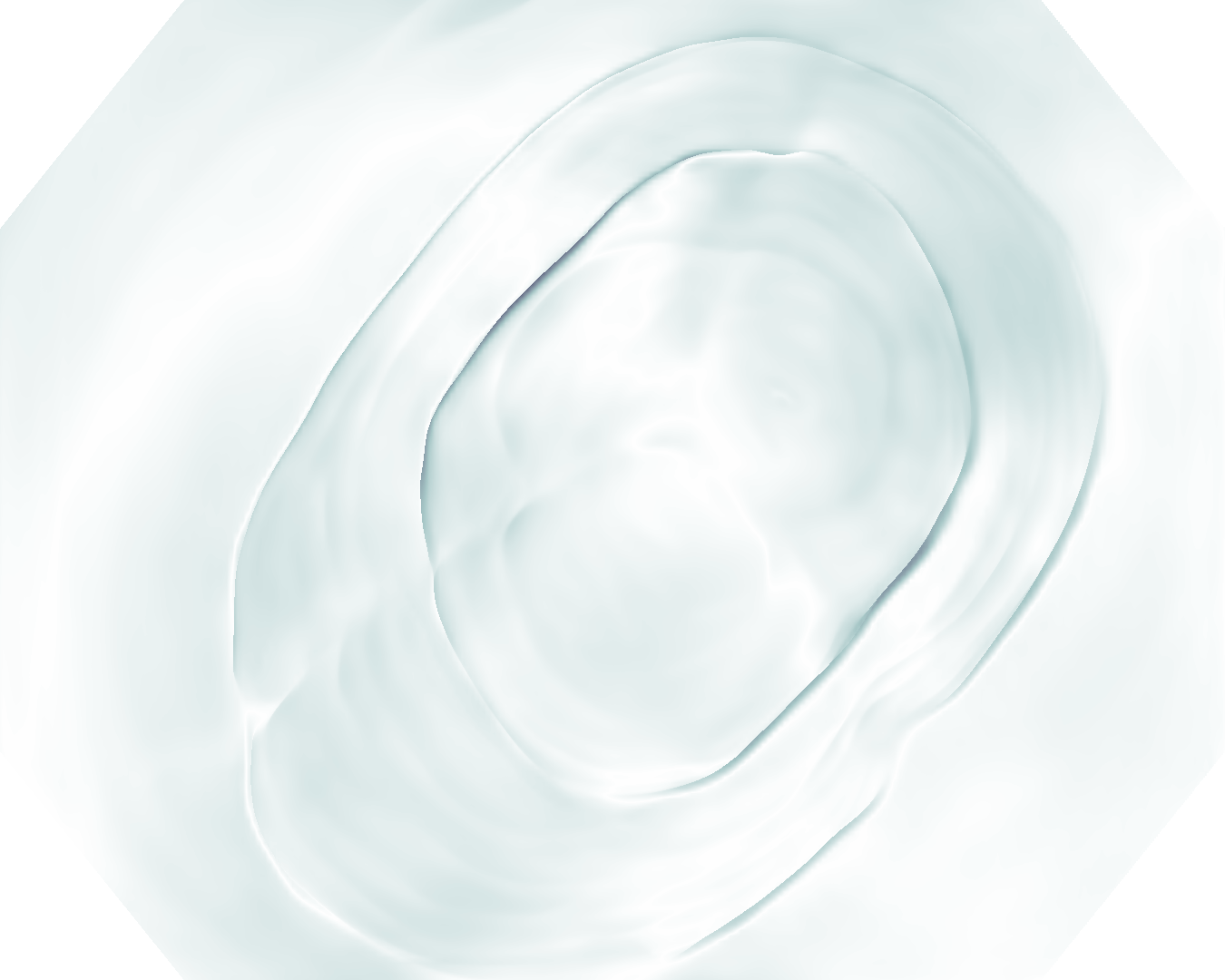} & \includegraphics[width=2cm]{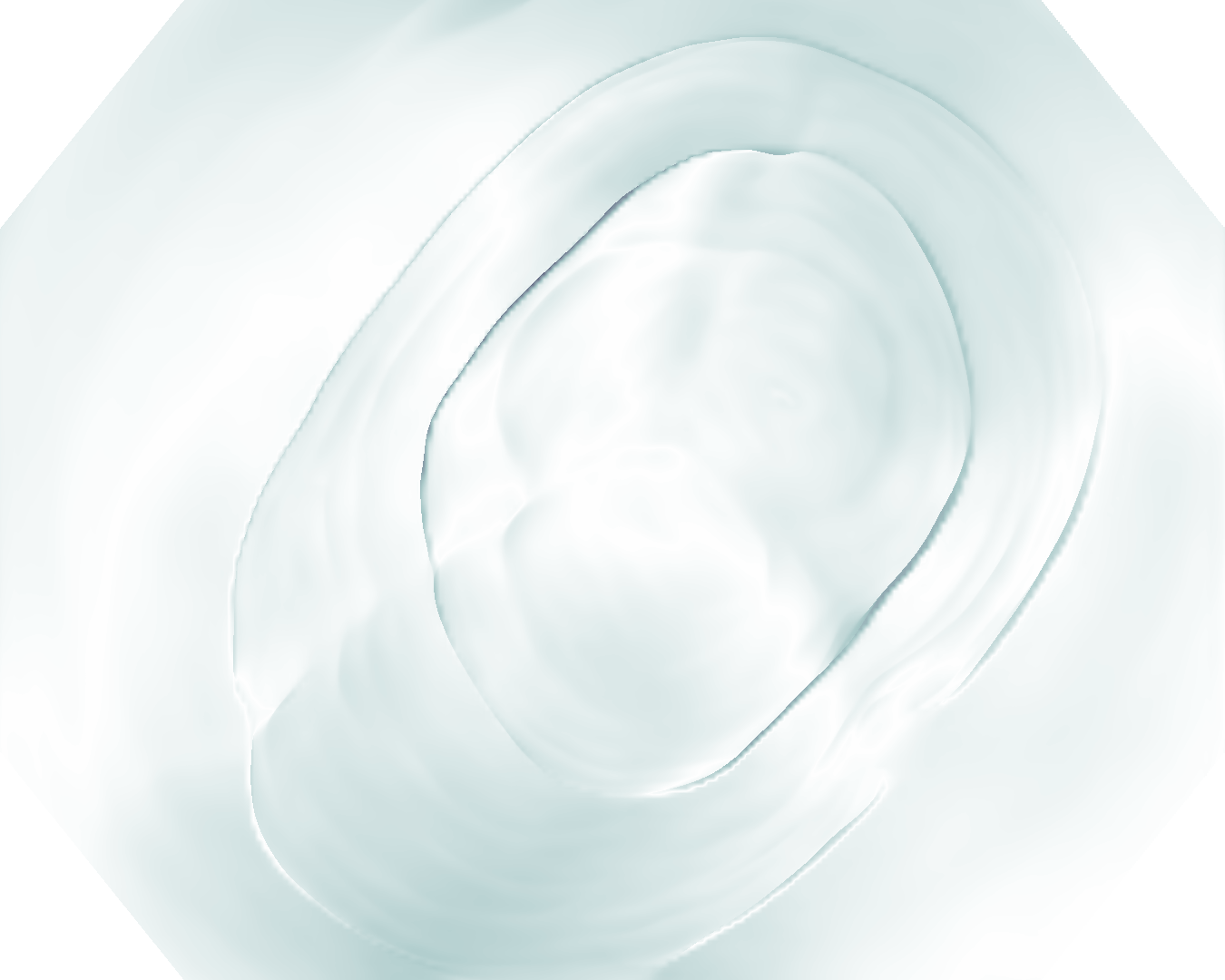} & \includegraphics[width=2cm]{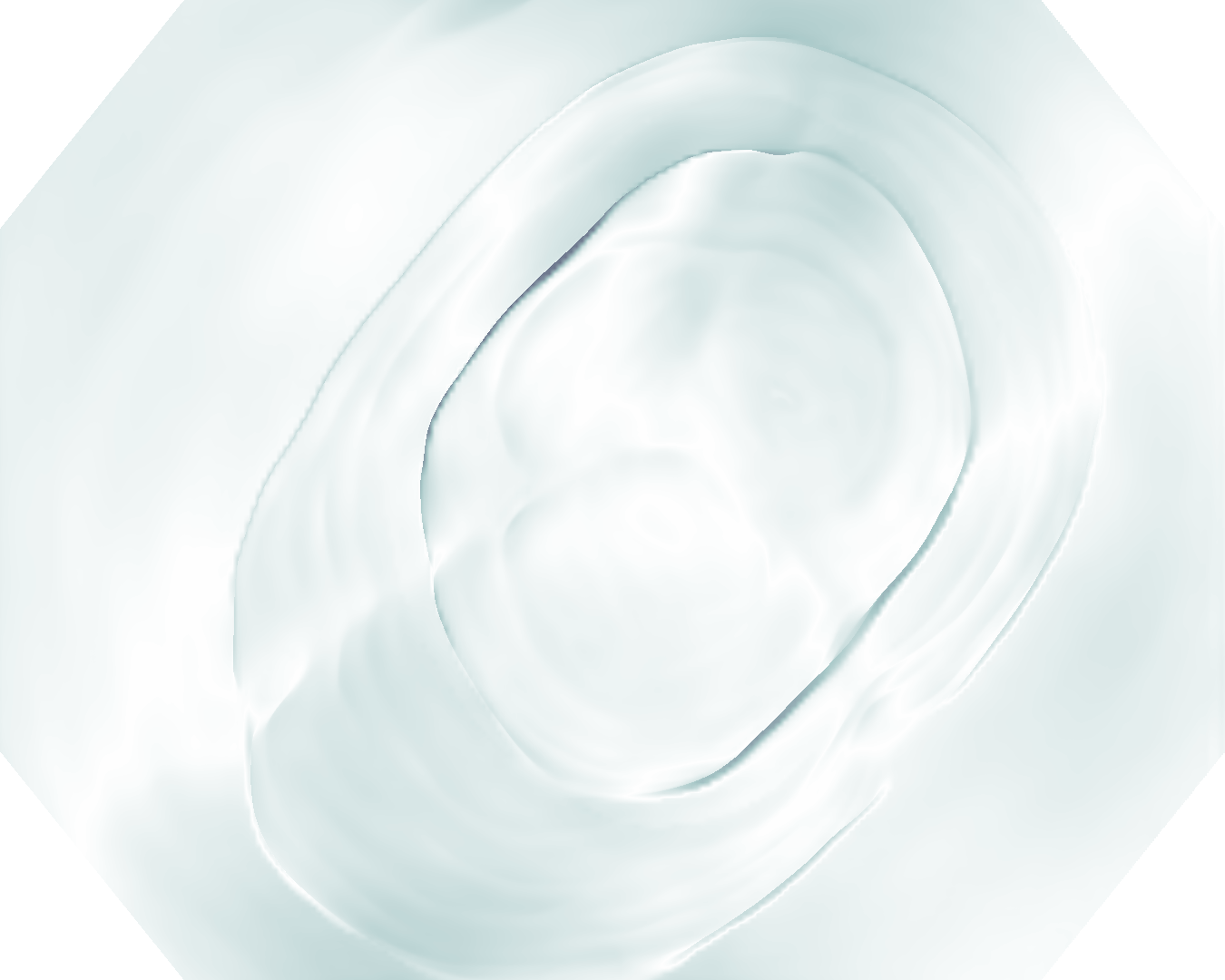} & \includegraphics[width=2cm]{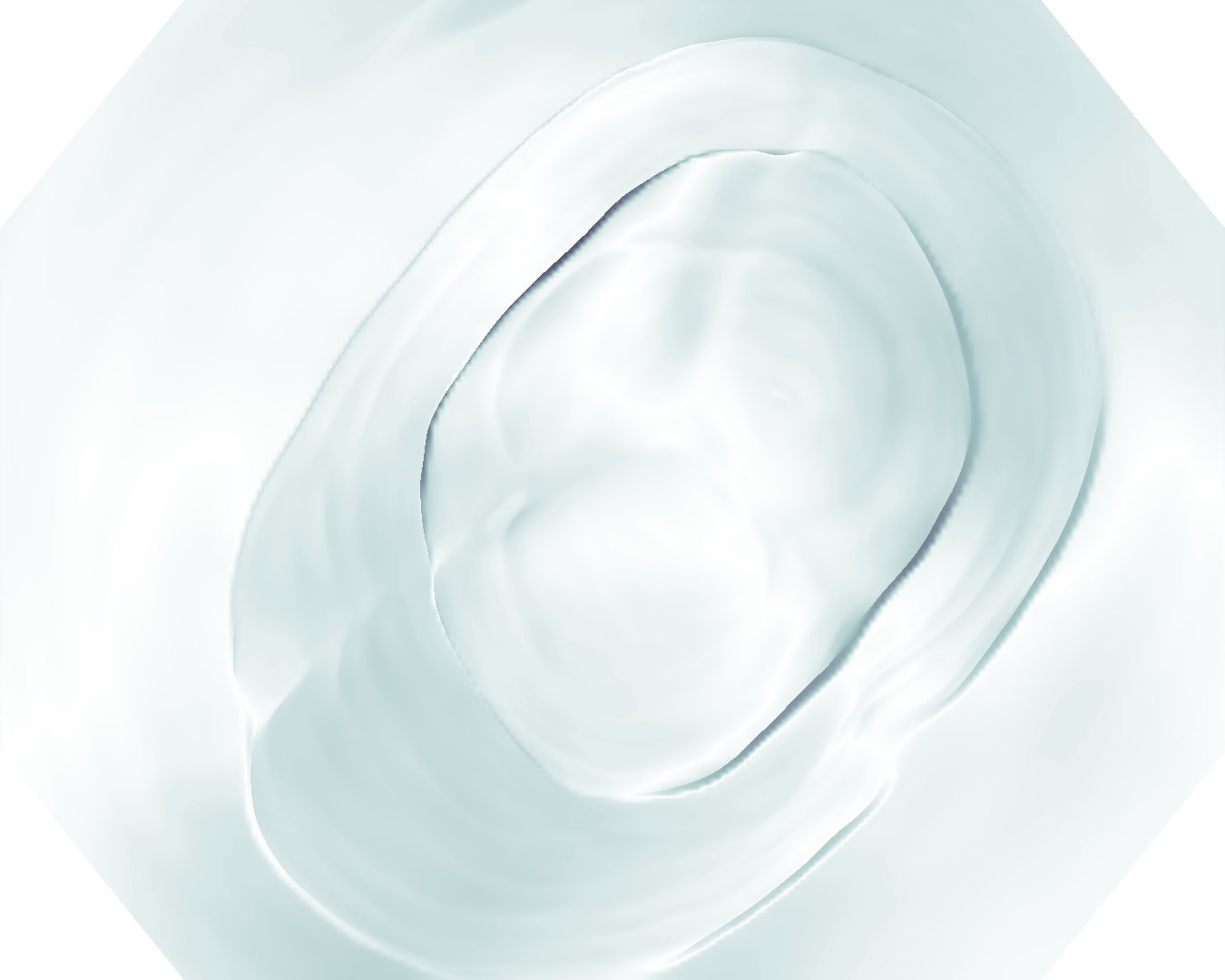} & \includegraphics[width=2cm]{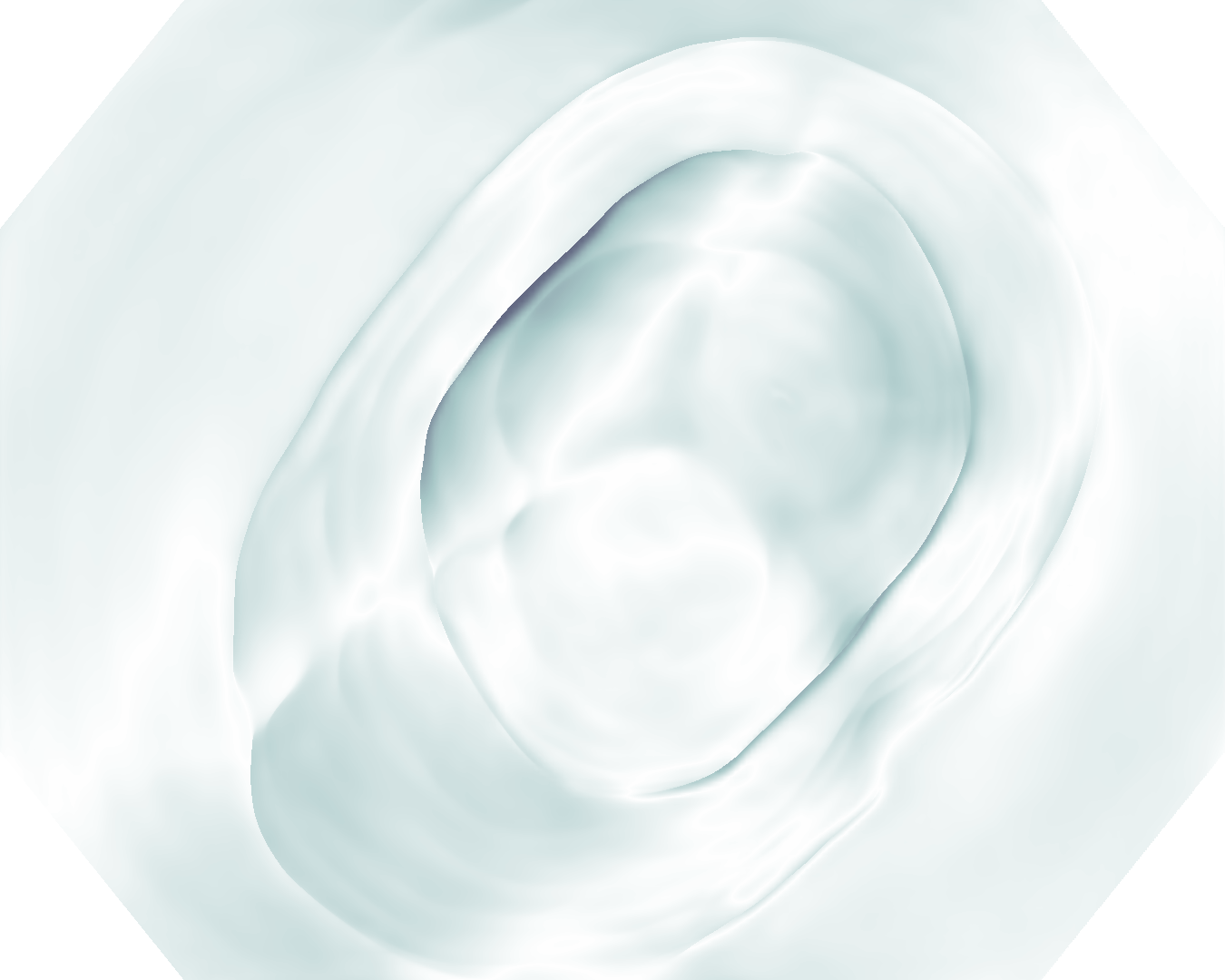} \\

  Input & VT-HK-MC & VT-HK-MA  & VT-IN-MC  & VT-IN-MA  & VT-IN-SL  & VT-NA-NA  \\ 
 \includegraphics[width=2cm]{11_color.png} & 
 \includegraphics[width=2cm]{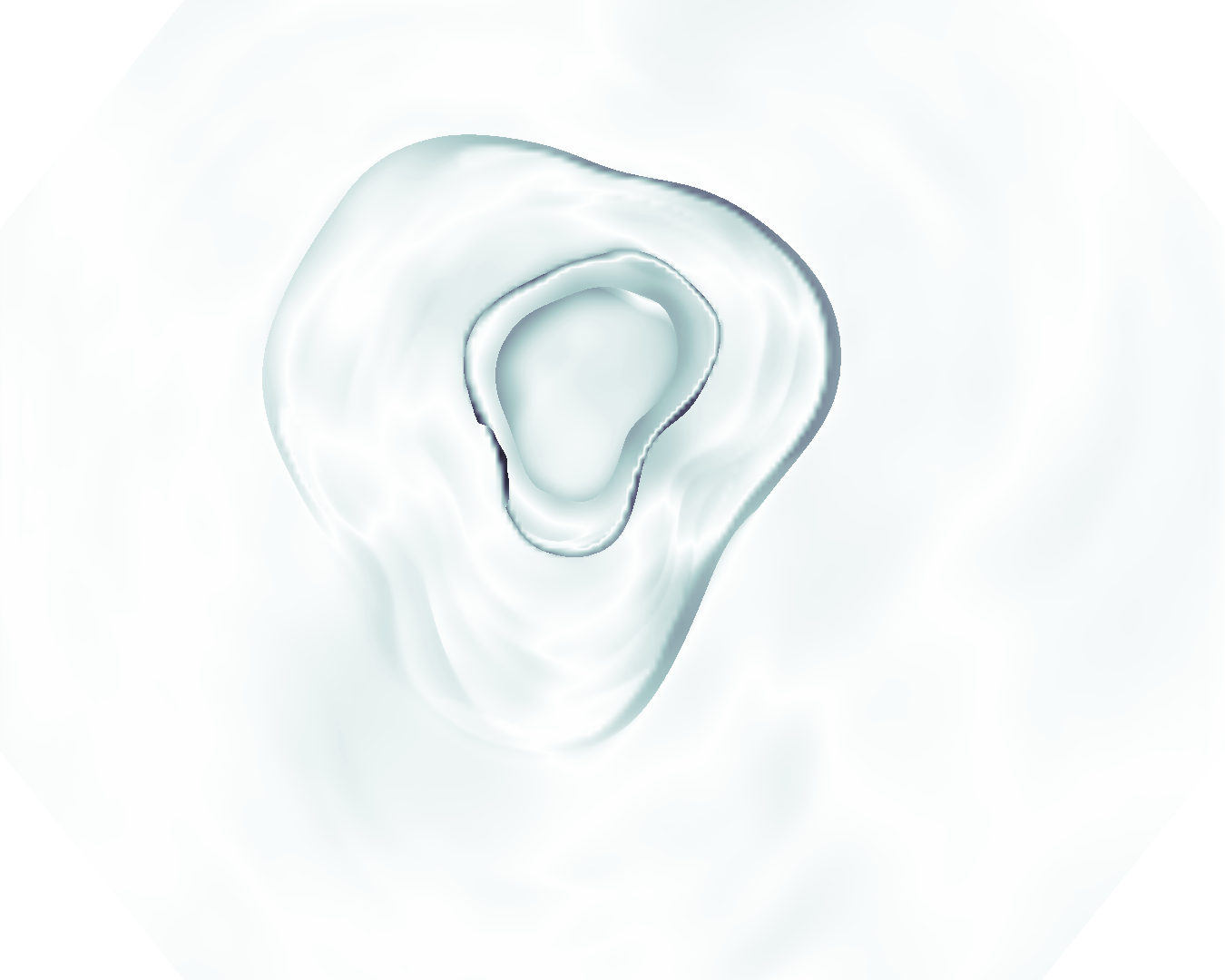}& \includegraphics[width=2cm]{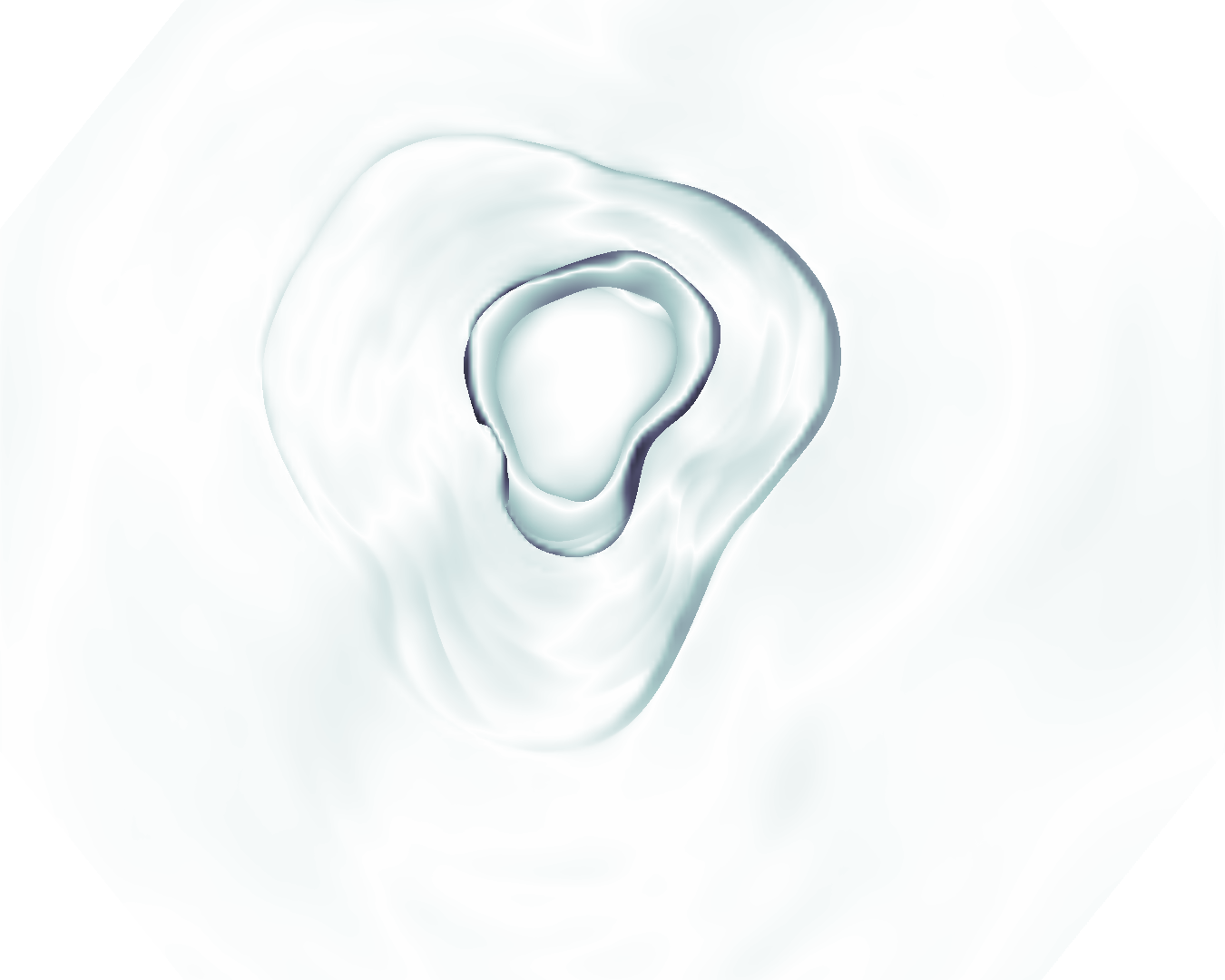} & \includegraphics[width=2cm]{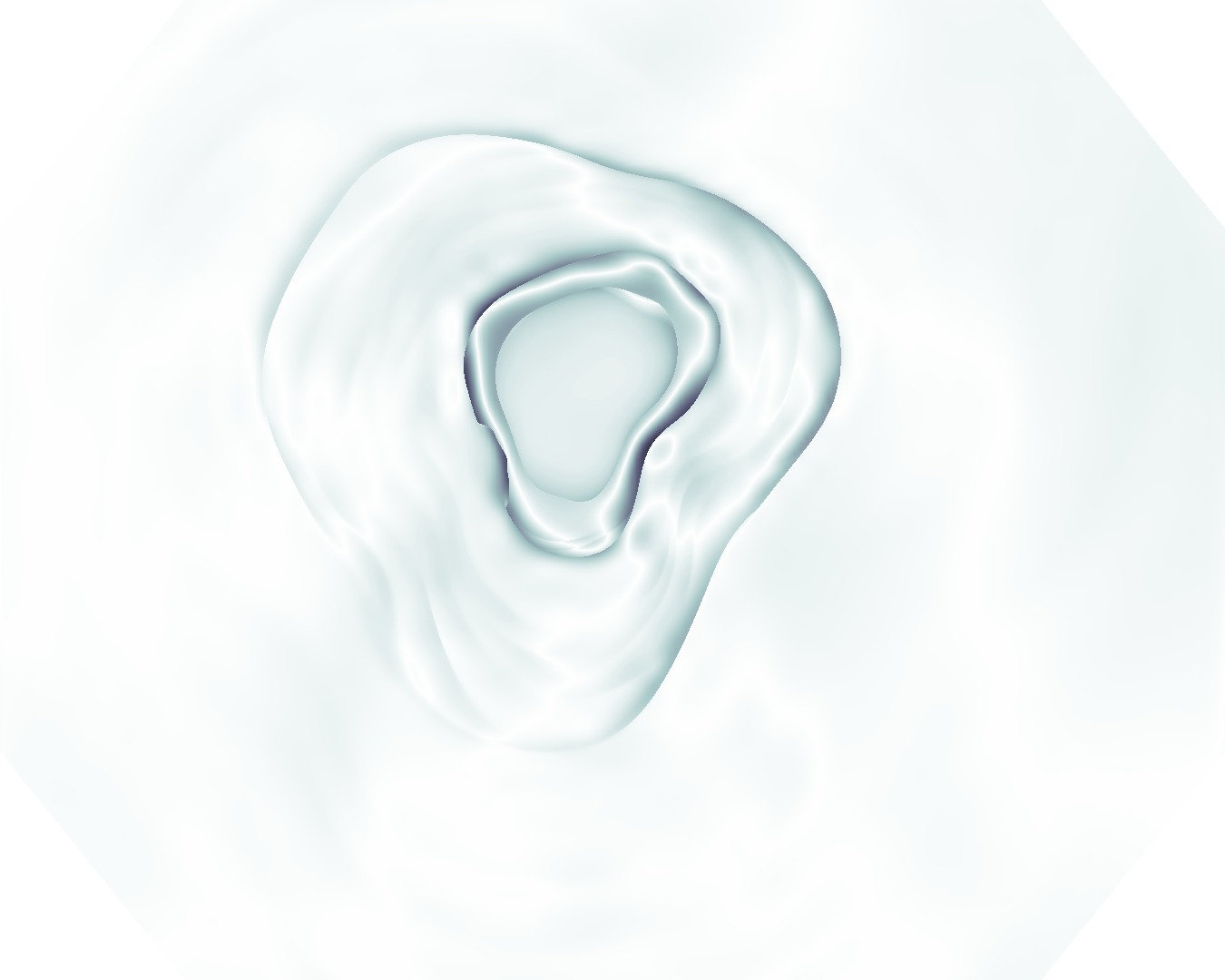} & \includegraphics[width=2cm]{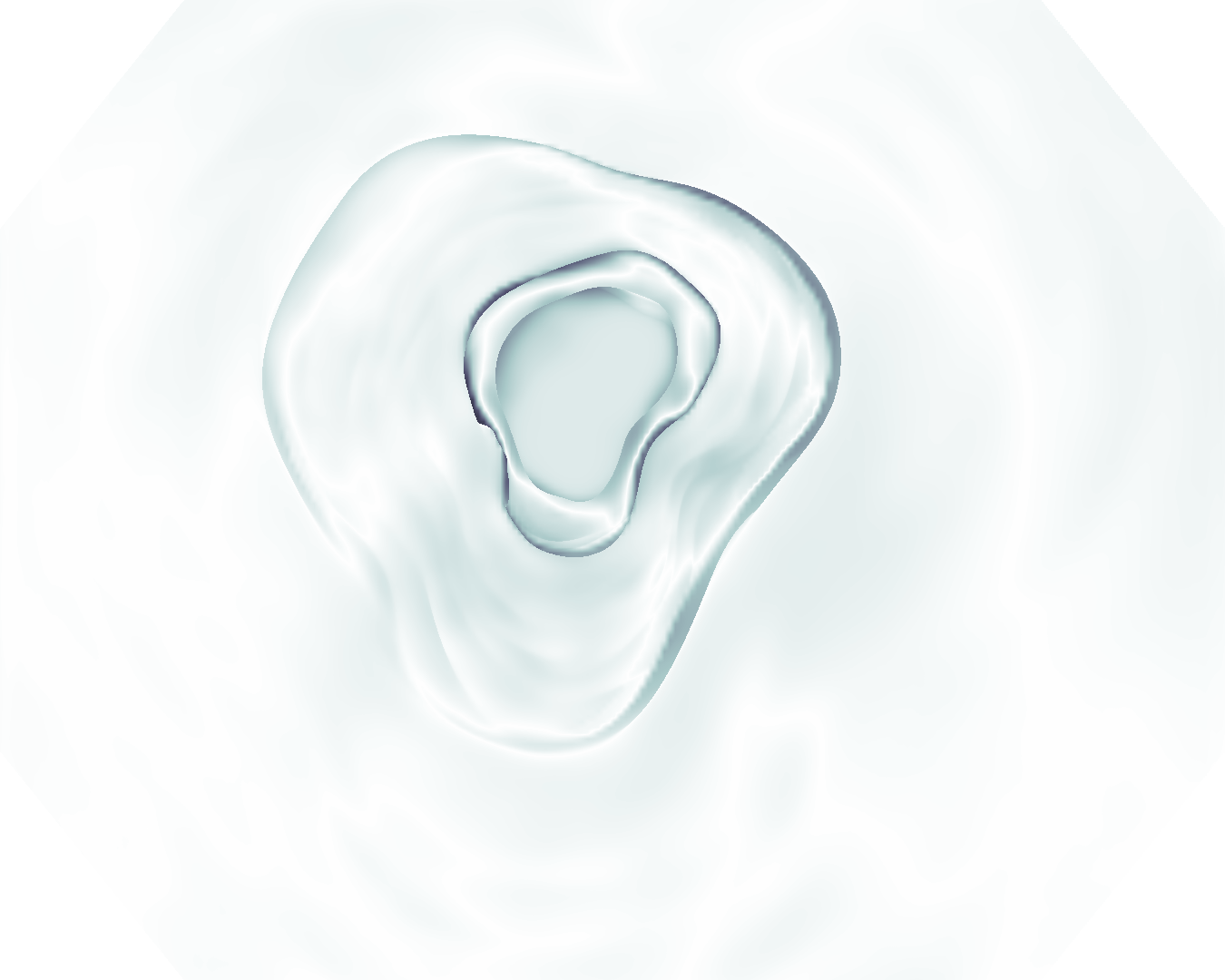} & \includegraphics[width=2cm]{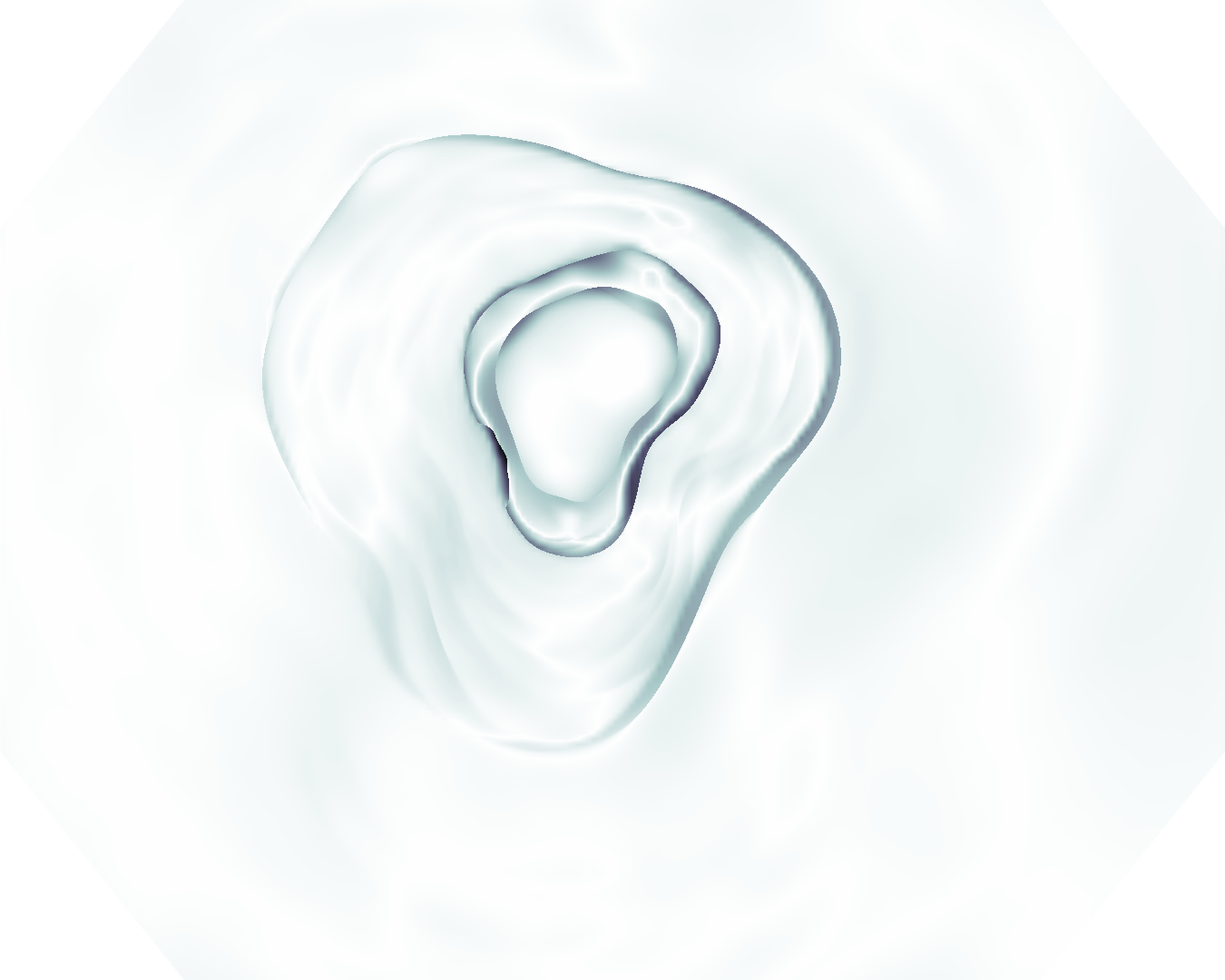} & \includegraphics[width=2cm]{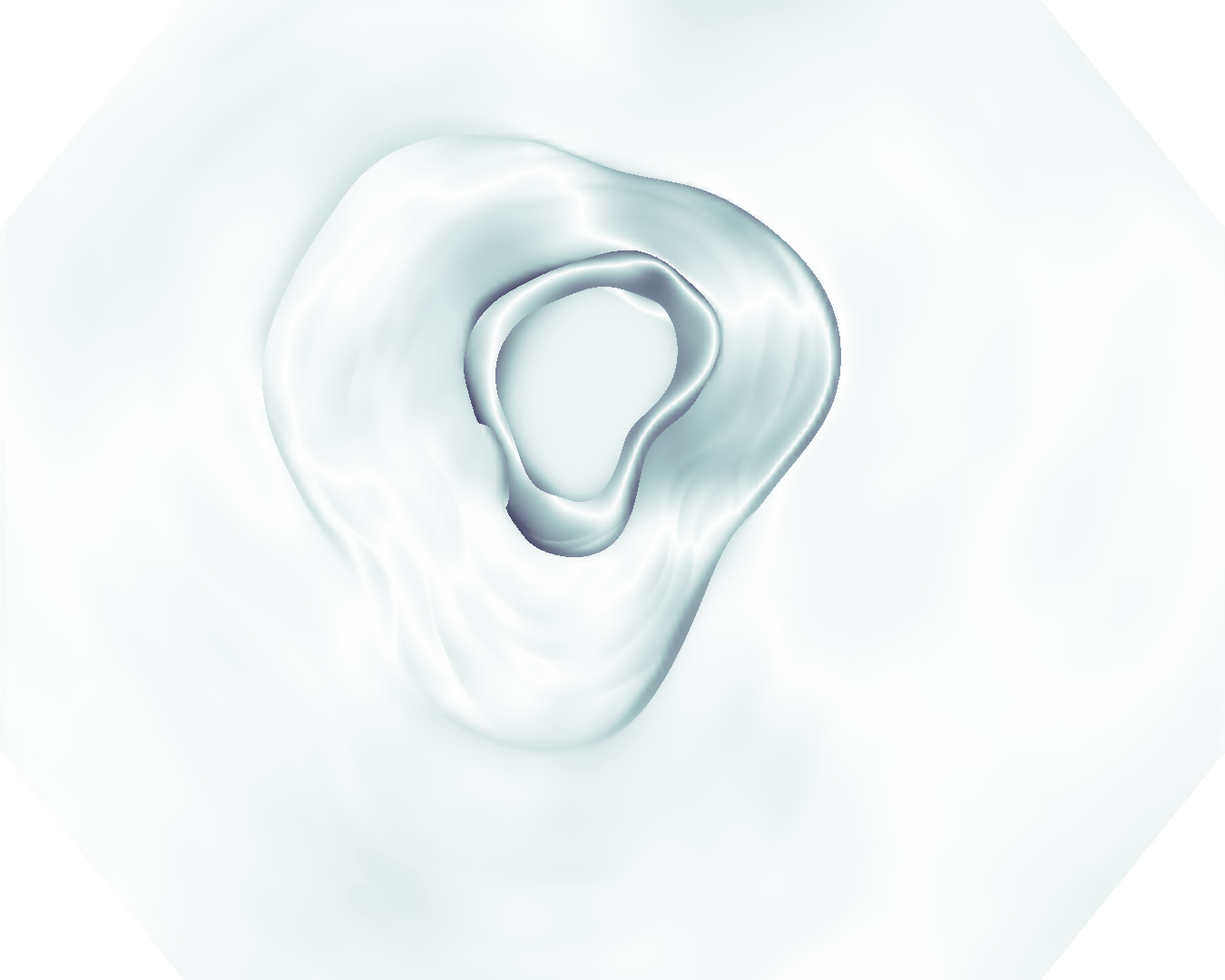} \\
   \includegraphics[width=2cm]{365_color.png}  &\includegraphics[width=2cm]{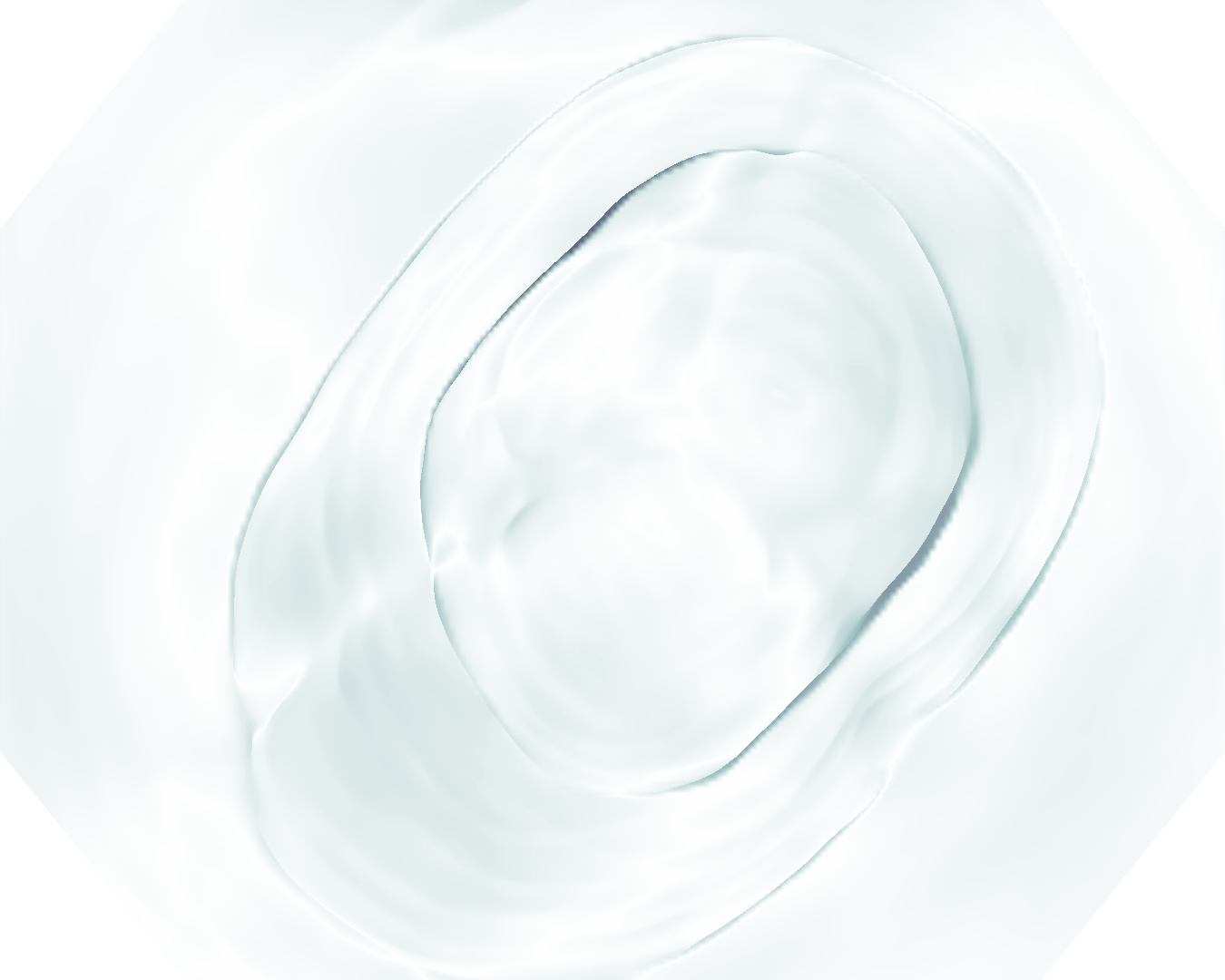} & \includegraphics[width=2cm]{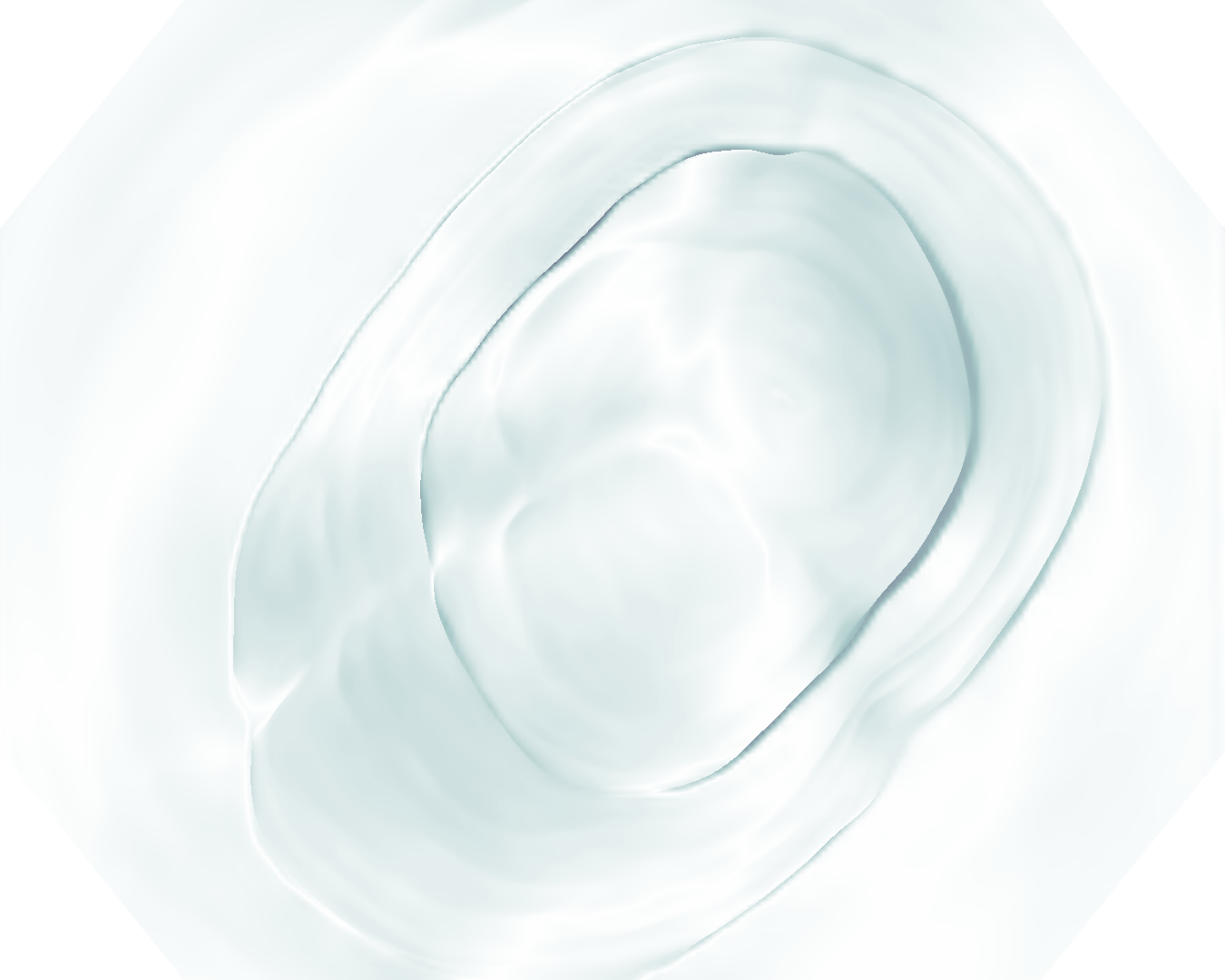} & \includegraphics[width=2cm]{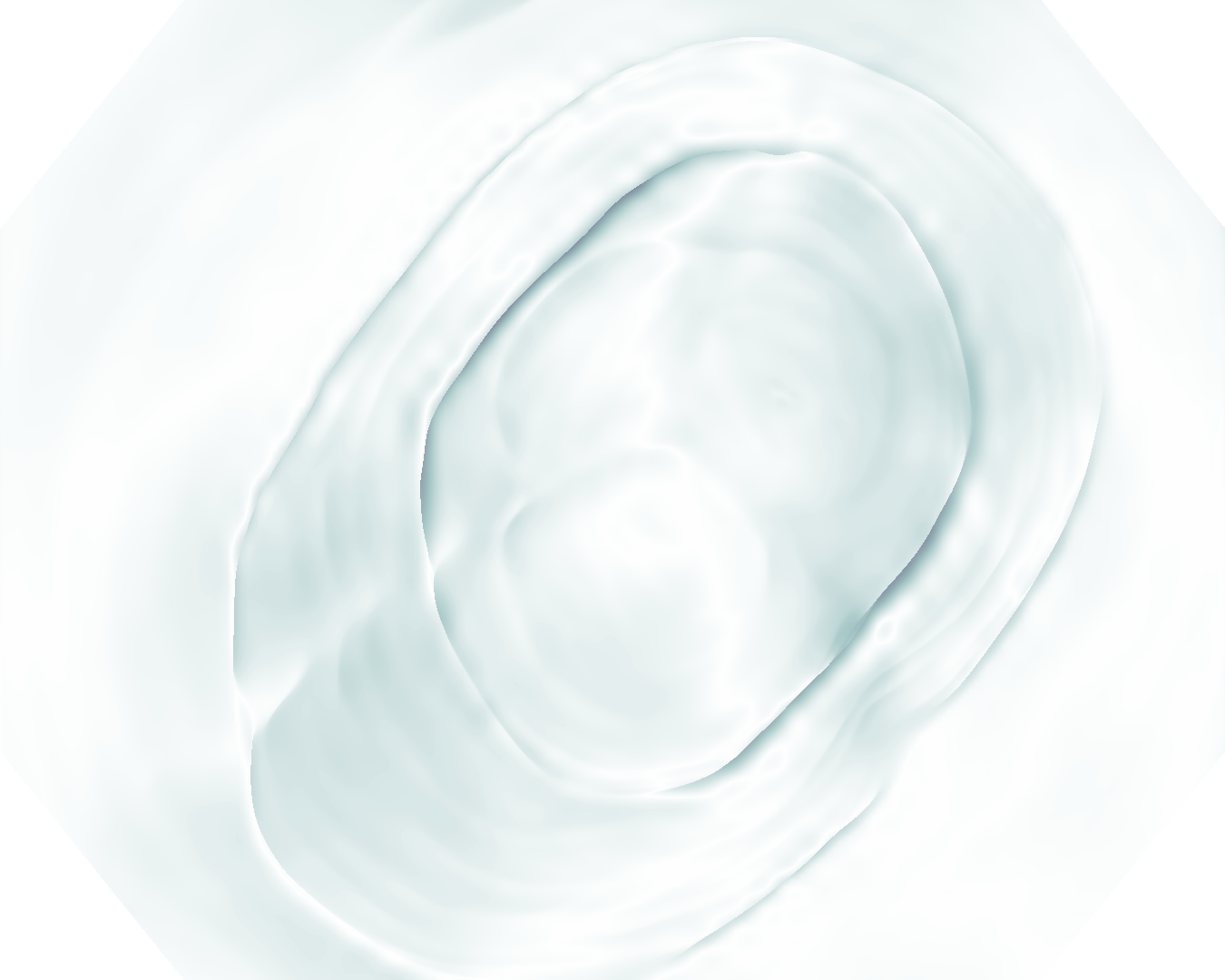} & \includegraphics[width=2cm]{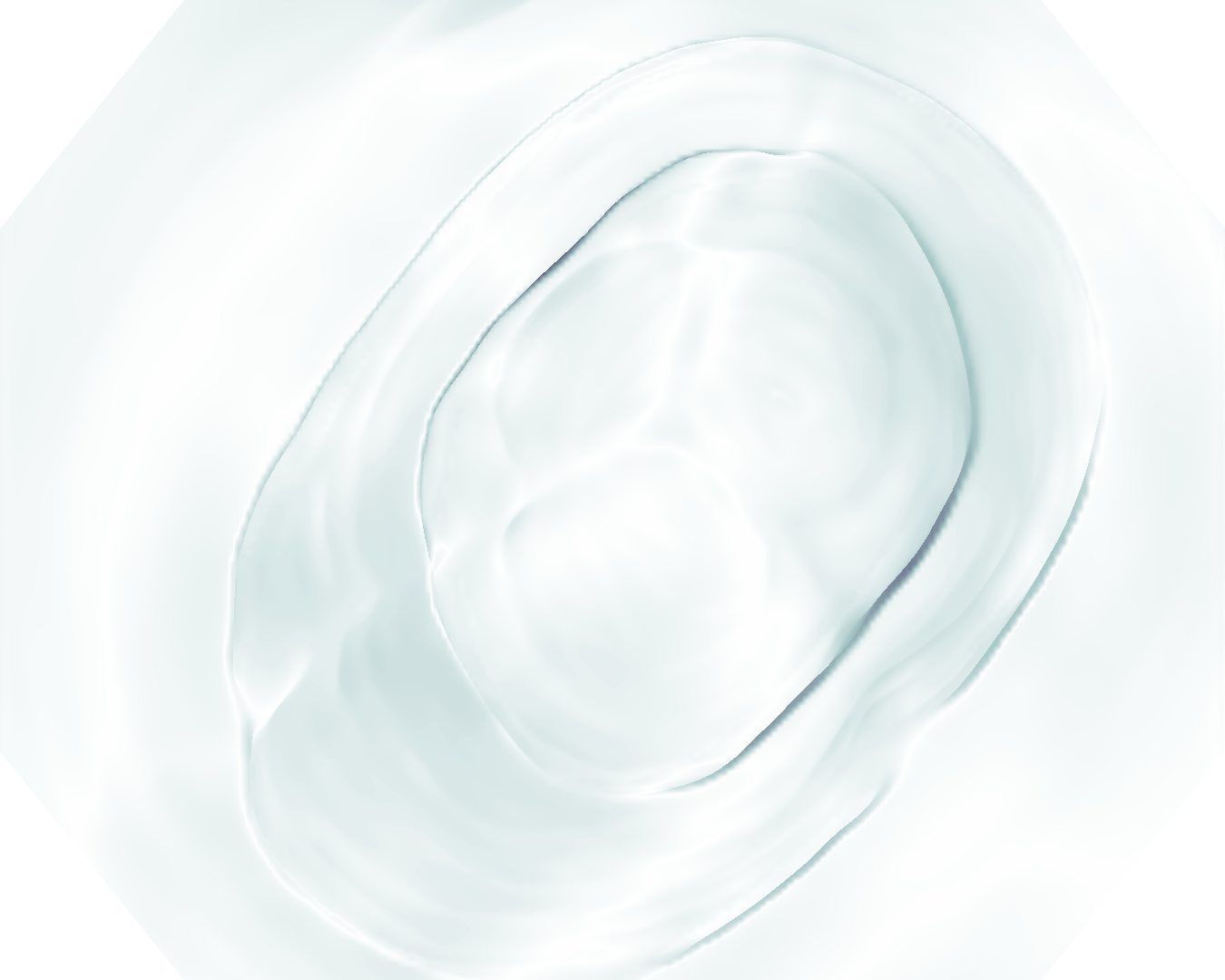} & \includegraphics[width=2cm]{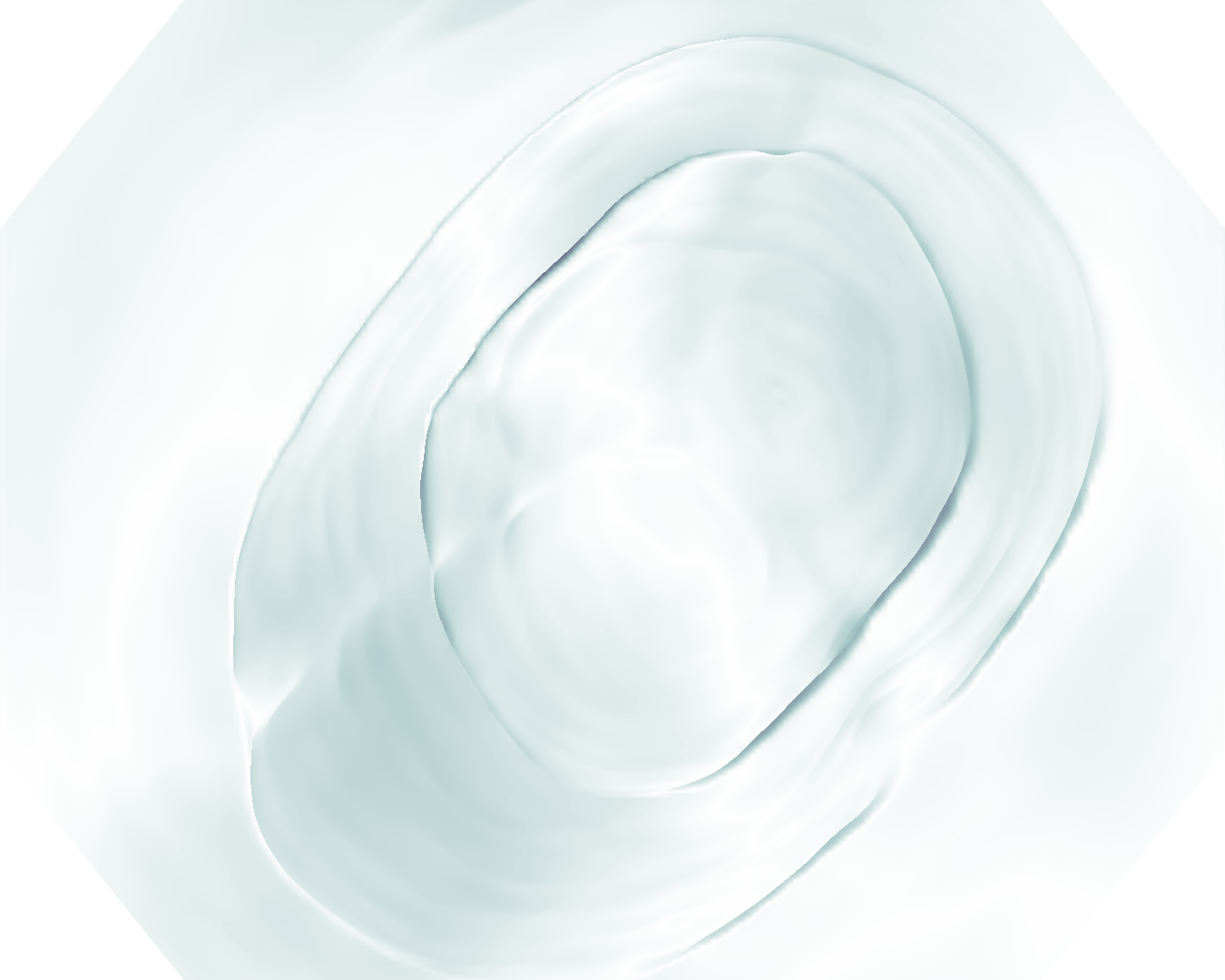}& \includegraphics[width=2cm]{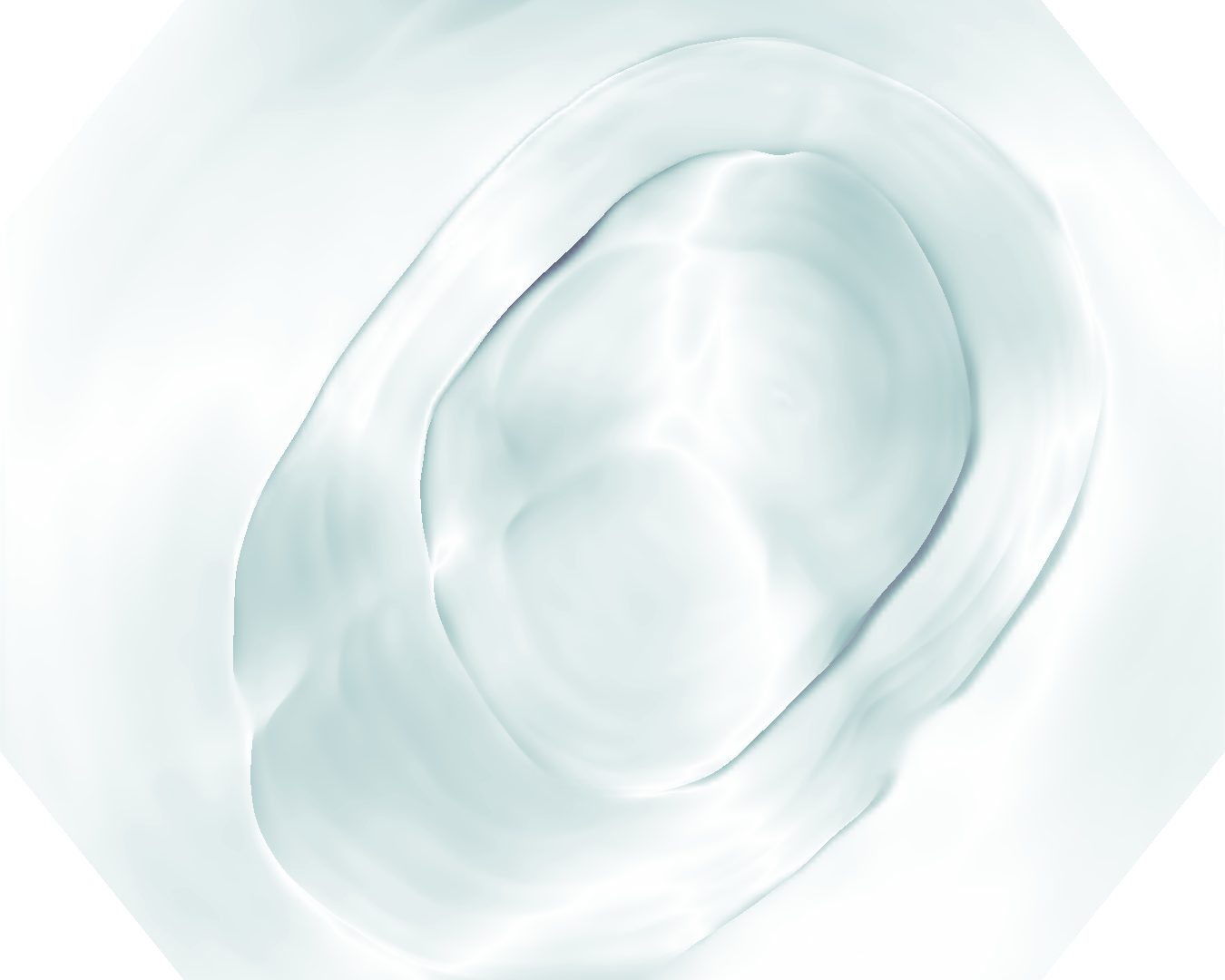}
\end{tabular}}
    \caption{Error maps for the post-processed predictions shown in Fig. \ref{fig:example_depth}, illustrating the absolute error with a larger value represented by a darker shade. For conciseness, we denote ResNet50s with \textit{RN}, ViT-Bs with \textit{VT}, Hyperkvasir-unlabelled with \textit{HK}, ImageNet-1k with \textit{IN}, MoCo v3 with \textit{MC}, Barlow Twins with \textit{BT}, MAE with \textit{MA}, supervised pretraining with \textit{SL}, and no pretraining with \textit{NA-NA}.}
    \label{fig:example_depth_error}
\end{figure*}

We evaluate the resulting monocular depth estimation models using the test data, where the images are pre-processed in the same manner as the validation images. We load two target depth maps for each image, one which is pre-processed in the same manner as the validation depth maps, for computing the scale and shift for pixels covered by the lens, and one left at the original size and scale ($[0\textrm{cm},10\textrm{cm}]$), for computing the performance. We compute and apply the scale and shift for the prediction, then resize the result to $\mathrm{max}(h,w)\times \mathrm{max}(h,w)$, where $h$ and $w$ are the height and width of the original image, crop to $h\times w$ to remove values for padded pixels, clip values to $[0,1]$, set any values for pixels not covered by the lens to 0, and scale the resulting values to $[0\textrm{cm},10\textrm{cm}]$. We then use the four metrics used the SimCol3D challenge \cite{simcol}, namely the arithmetic mean across the test set of: the root MSE (mRMSE), the median relative absolute error (mMRAE), and the mean absolute error (mMAE), which are only applied to pixels covered by the lens:

\begin{equation}
    \text{mRMSE}=\frac{1}{N_e}\sum_{i=1}^{N_e}\sqrt{\frac{1}{N_V}\sum_{j=1}^{N_V}\left(\hat{y}_{i,j}-y_{i,j}\right)^2}
\end{equation}

\begin{equation}
    \text{mMRAE}=\frac{1}{N_e}\sum_{i=1}^{N_e}\median_{j=1,\ldots,N_V}\left(\left\vert\frac{\hat{y}_{i,j}-y_{i,j}}{y_{i,j}}\right\vert\right)
\end{equation}

\begin{equation}
    \text{mMAE}=\frac{1}{N_eN_V}\sum_{i=1}^{N_e}\sum_{j=1}^{N_V}\left\vert\hat{y}_{i,j}-y_{i,j}\right\vert
\end{equation}

\noindent where $N_V$ is the number of pixels covered by the lens in an image at its original size, $\hat{y}_{i,j}$ is the value in the post-processed prediction for the $j^{th}$ pixel covered by the lens in the $i^{th}$ image at its original size, and $y_{i,j}$ is the corresponding target value. For all metrics, a lower value indicates better performance. The results are presented in Table \ref{tab:depth}, and some examples are shown in Fig. \ref{fig:example_depth} with corresponding error maps shown in Fig. \ref{fig:example_depth_error} to help visualise the differences.

\begin{table*}[ht]
\caption{\label{tab:depth}Performance in monocular depth estimation in colonoscopy. The best results for each architecture are highlighted as bold, and the best results overall are underlined.}
\centering
\begin{tabular}{cccccc}
\toprule
Backbone arch.            & Pretraining data                      & Pretraining algo. & mRMSE (cm)                                                    & mMRAE                                                          & mMAE (cm)                                                   \\ \hline
\multirow{6}{*}{ResNet50} & \multirow{2}{*}{Hyperkvasir-unlabel.} & MoCo v3           & \textbf{0.177}                              & \textbf{0.0345}                              & \textbf{0.131}                              \\
                          &                                       & Barlow Twins      & 0.178                                                        & 0.0362                                                        & 0.134                                                        \\ \cmidrule{2-6} 
                          & \multirow{3}{*}{ImageNet-1k}          & MoCo v3           & 0.207                                                        & 0.0428                                                        & 0.159                                                        \\ 
                          &                                       & Barlow Twins      & 0.203                                                        & 0.0428                                                        & 0.156                                                        \\ \cmidrule{3-6}
                          &                                       & Supervised        & 0.208                                                        & 0.0452                                                        & 0.159                                                        \\ \cmidrule{2-6} 
                          & None                                  & None              & 0.241                                                        & 0.0515                                                        & 0.188                                                        \\ \cmidrule{1-6}
\multirow{6}{*}{ViT-B}    & \multirow{2}{*}{Hyperkvasir-unlabel.} & MoCo v3           & 0.151                                                        & 0.0259                                                        & 0.104                                                        \\
                          &                                       & MAE               & \textbf{\underline{0.143}} & \textbf{\underline{0.0246}} & \textbf{\underline{0.098}} \\ \cmidrule{2-6} 
                          & \multirow{3}{*}{ImageNet-1k}          & MoCo v3           & 0.166                                                        & 0.0274                                                        & 0.116                                                        \\
                          &                                       & MAE               & 0.160                                                        & 0.0298                                                        & 0.115                                                        \\ \cmidrule{3-6}
                          &                                       & Supervised        & 0.168                                                        & 0.0322                                                        & 0.121                                                        \\ \cmidrule{2-6} 
                          & None                                  & None              & 0.226                                                        & 0.0436                                                        & 0.167                                                        \\ \bottomrule
\end{tabular}
\end{table*}

\section{Analysis}

\begin{figure}[ht]
\centering
\includegraphics[width=\linewidth]{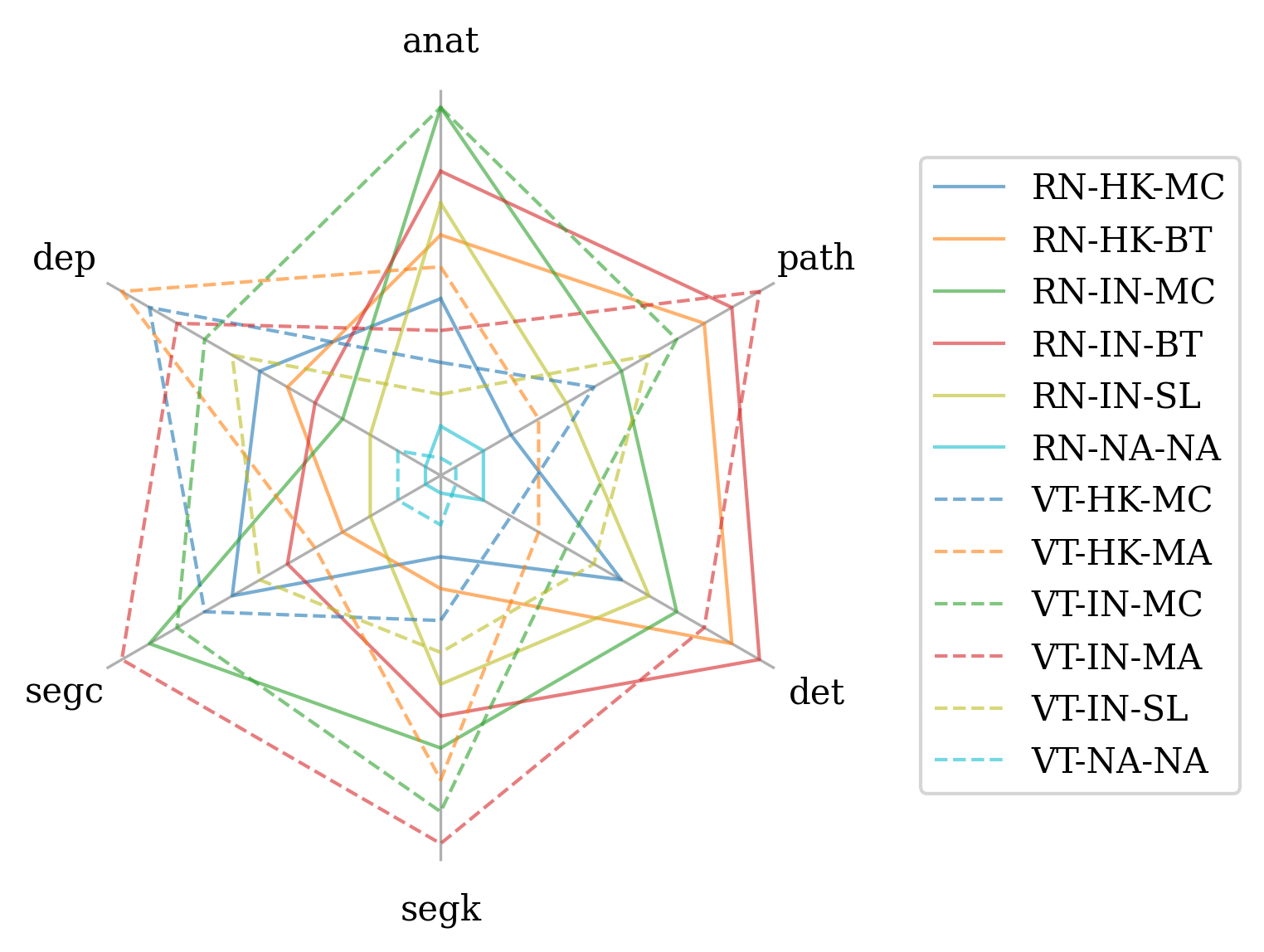}
\caption{Ranking of the performance of each model on each task, as measured by mF1 (anatomical landmark recognition and pathological finding characterisation), AP (polyp detection), mDice (polyp segmentation), and mRMSE (monocular depth estimation in colonoscopy), where a better rank is represented by a greater distance from the centre. For conciseness, we denote ResNet50s with \textit{RN}, ViT-Bs with \textit{VT}, Hyperkvasir-unlabelled with \textit{HK}, ImageNet-1k with \textit{IN}, MoCo v3 with \textit{MC}, Barlow Twins with \textit{BT}, MAE with \textit{MA}, supervised pretraining with \textit{SL}, and no pretraining with \textit{NA-NA}. Additionally, we refer to anatomical landmark recognition as \textit{anat}, pathological finding characterisation as \textit{path}, polyp detection as \textit{det}, polyp segmentation with Kvasir-SEG as \textit{segk}, polyp segmentation with CVC-ClinicDB as \textit{segc}, and monocular depth estimation in colonoscopy as \textit{dep}.}
\label{fig:rank}
\end{figure}

The results presented in the previous sections primarily provide an indication of the ranking of the pretraining pipelines for each considered GIE vision task. Notably, there is some variation in this ranking, as illustrated in Fig. \ref{fig:rank}, however the ViT-B encoder pretrained with MAE and ImageNet-1k most consistently allows for either the best, or highly competitive, downstream performance. Beyond this identification, however, these results provide evidence for more general principles regarding the pretraining of encoders for use as backbones in solutions to GIE vision tasks, which we reveal through an analysis presented in this section.

\begin{figure*}[ht]
\centering
\includegraphics[width=450pt]{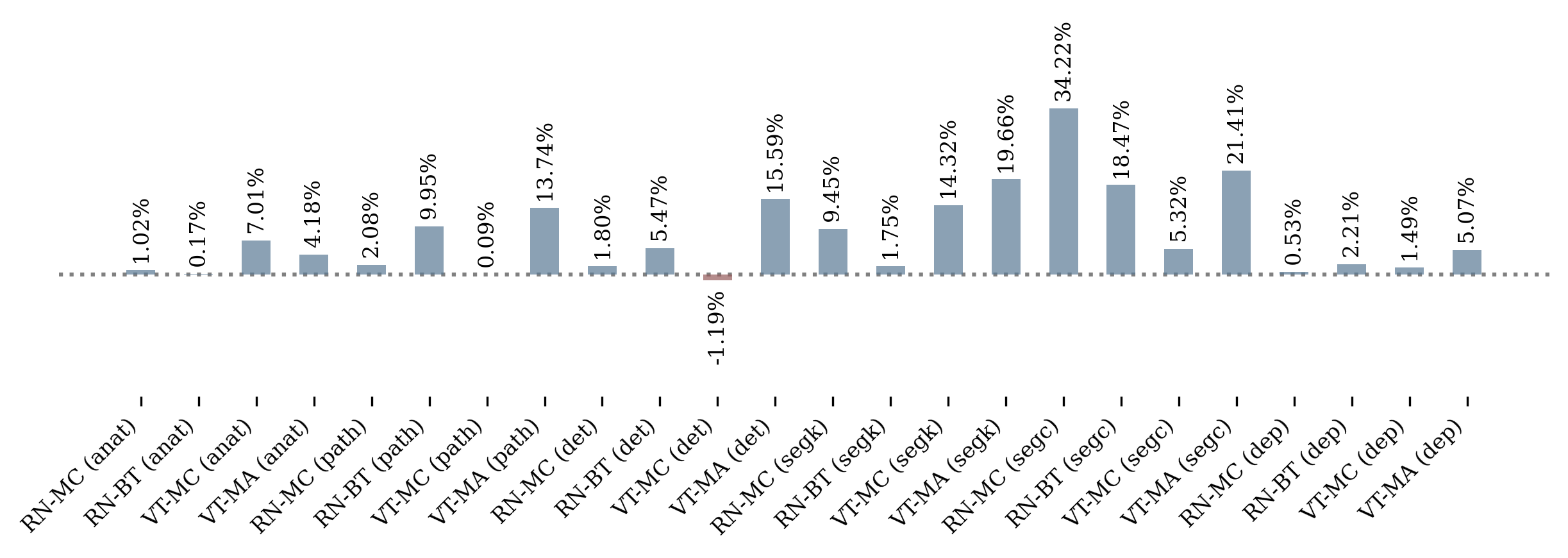}
\caption{Improvement of self-supervised pretraining \textit{vs.} supervised pretraining for same architecture and pretraining data (ImageNet-1k). For conciseness, we denote ResNet50s with \textit{RN}, ViT-Bs with \textit{VT}, MoCo v3 with \textit{MC}, Barlow Twins with \textit{BT}, and MAE with \textit{MA}. Additionally, we refer to anatomical landmark recognition as \textit{anat}, pathological finding characterisation as \textit{path}, polyp detection as \textit{det}, polyp segmentation with Kvasir-SEG as \textit{segk}, polyp segmentation with CVC-ClinicDB as \textit{segc}, and monocular depth estimation in colonoscopy as \textit{dep}.}
\label{fig:sl_vs_ssl}
\end{figure*}

First, we demonstrate that self-supervised pretraining is generally more suitable than supervised pretraining. To assess this, we evaluate the \textit{relative} improvement of each model that uses a backbone pretrained in a self-supervised manner with ImageNet-1k \textit{vs.} the equivalent model (same architecture and task) that uses a backbone pretrained in a supervised manner with ImageNet-1k. To compute the relative improvement, we consider the primary metric for each task as mF1 (image classification), AP (object detection), mDice (semantic segmentation), and mRMSE (depth estimation), as defined in the discussion of each task. Then, for all but mRMSE, we take the absolute difference between the result and a perfect score of 1, in order to convert each \textit{score} (higher is better) to a measure of \textit{error} (lower is better). We do not do this for mRMSE since it is already a measure of error. We then compute the relative improvement using:

\begin{equation}
    \%\mathrm{Improvement}_{SL\rightarrow SSL}=100\frac{\delta_{SL}-\delta_{SSL}}{\delta_{SL}}
\end{equation}

\noindent where $\delta_{SSL}$ is the error for a model with a backbone pretrained in a self-supervised manner and $\delta_{SL}$ is the error for an equivalent model (same architecture, pretraining data, and task) with a backbone pretrained in a supervised manner. Note that this analysis omits any results for pretraining with Hyperkvasir-unlabelled or no pretraining. We visualise the results of this analysis in Fig. \ref{fig:sl_vs_ssl}, where it can be seen that self-supervised pretraining overwhelmingly provides improvements over supervised pretraining in our experiments, with only a single marginal exception to this observed across all self-supervised pretraining algorithms, architectures, and tasks. We can therefore confidently conclude that self-supervised pretraining with ImageNet-1k generally provides better backbones than supervised pretraining with ImageNet-1k. Since supervised pretraining with ImageNet-1k is still the conventional pretraining pipeline for backbones used in solutions to vision tasks in GIE, including the state-of-the-art, this is a crucial finding.

We also demonstrate that self-supervised pretraining with ImageNet-1k is generally more suitable than self-supervised pretraining with Hyperkvasir-unlabelled in the considered downstream tasks, with the notable exception of monocular depth estimation in colonoscopy. To assess this, we use the same measures of error used in the previous analysis and evaluate the relative improvement from pretraining with Hyperkvasir-unlabelled \textit{vs.} ImageNet-1k using:

\begin{equation}
    \%\mathrm{Improvement}_{IN\rightarrow HK}=100\frac{\delta_{IN}-\delta_{HK}}{\delta_{IN}}
\end{equation}

\noindent where $\delta_{HK}$ is the error for a model with a backbone pretrained with Hyperkvasir-unlabelled and $\delta_{IN}$ is the error for an equivalent model (same architecture, pretraining algorithm, and task) with a backbone pretrained with ImageNet-1k.

\begin{figure*}[ht]
\centering
\includegraphics[width=450pt]{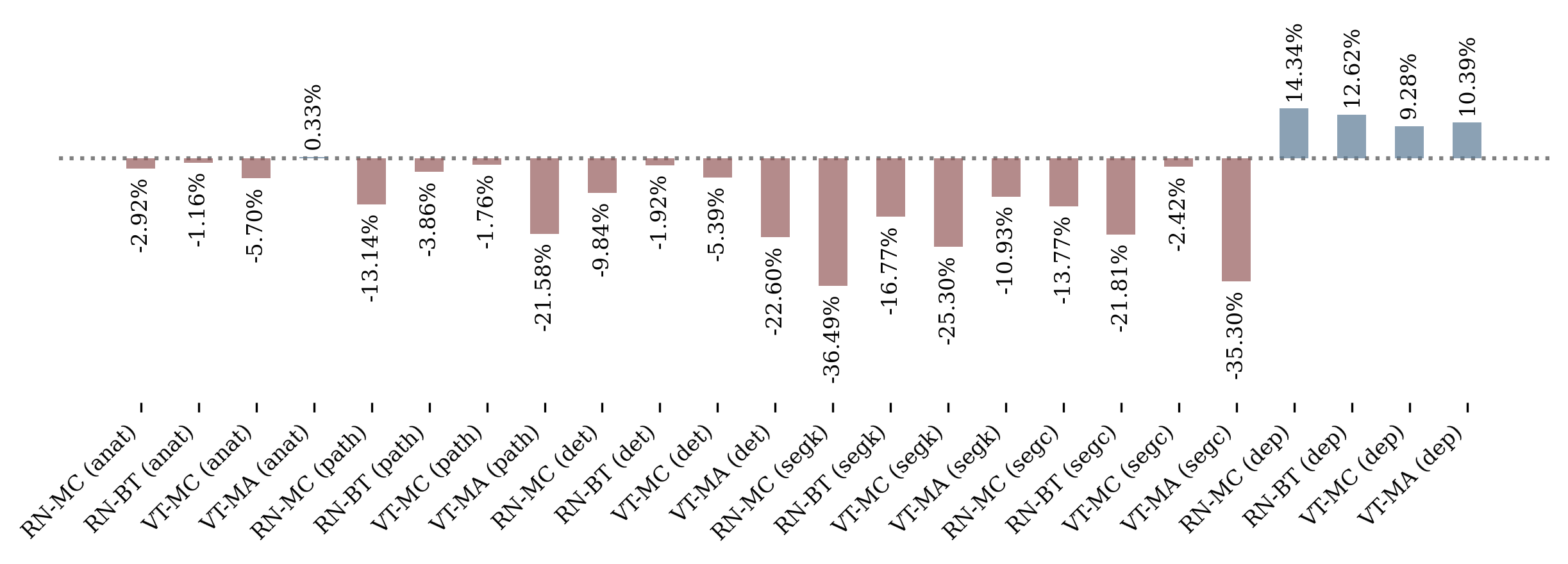}
\caption{Improvement of pretraining with Hyperkvasir-unlabelled \textit{vs.} pretraining with ImageNet-1k for same architecture and self-supervised pretraining algorithm. For conciseness, we denote ResNet50s with \textit{RN}, ViT-Bs with \textit{VT}, MoCo v3 with \textit{MC}, Barlow Twins with \textit{BT}, and MAE with \textit{MA}. Additionally, we refer to anatomical landmark recognition as \textit{anat}, pathological finding characterisation as \textit{path}, polyp detection as \textit{det}, polyp segmentation with Kvasir-SEG as \textit{segk}, polyp segmentation with CVC-ClinicDB as \textit{segc}, and monocular depth estimation in colonoscopy as \textit{dep}.}
\label{fig:hk_vs_in}
\end{figure*}

Note that this analysis omits any results for supervised pretraining or no pretraining. We visualise the results of this analysis in Fig. \ref{fig:hk_vs_in}, where it can be seen that self-supervised pretraining with ImageNet-1k generally provides better performance than self-supervised pretraining with Hyperkvasir-unlabelled, with exceptions including the anatomical landmark recognition models with MAE pretrained backbones, as well as all monocular depth estimation models. While the result for the anatomical landmark recognition models with MAE pretrained backbones shows only a marginal improvement for pretraining with Hyperkvasir-unlabelled \textit{vs.} ImageNet-1k, the results for the depth estimation models are more significant. This implies that the similarity of the pretraining data to the data used in the depth estimation experiments is much more critical than the amount of pretraining data, in comparison to other tasks. While this finding is significant for the development of solutions to vision tasks in GIE, it may have broader implications and further work may find this to be true for monocular depth estimation in general.

Finally, we demonstrate that models with a ViT-B backbone are generally better than models with a ResNet50 backbone in polyp segmentation and monocular depth estimation in colonoscopy, generally worse in polyp detection, and generally similar in image classification. To assess this, we use the same measures of error used in the previous analyses and evaluate the relative improvement from using a ViT-B \textit{vs.} a ResNet50 using:

\begin{equation}
    \%\mathrm{Improvement}_{RN\rightarrow VT}=100\frac{\delta_{RN}-\delta_{VT}}{\delta_{RN}}
\end{equation}

\noindent where $\delta_{VT}$ is the error for a model with a ViT-B backbone and $\delta_{RN}$ is the error for an equivalent model (same pretraining pipeline and task) with a ResNet50 backbone.

\begin{figure*}[ht]
\centering
\includegraphics[width=450pt]{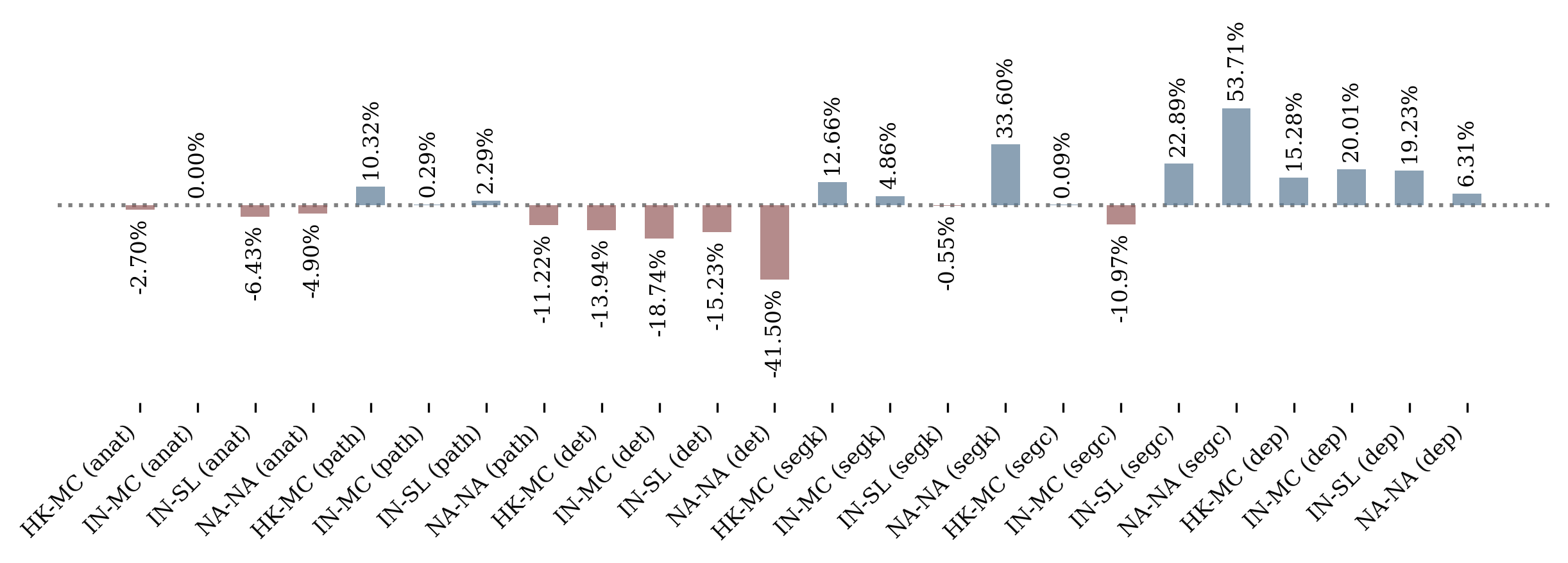}
\caption{Improvement of ViT-B over ResNet50 for same pretraining pipeline (data and algorithm). For conciseness, we denote Hyperkvasir-unlabelled with \textit{HK}, ImageNet-1k with \textit{IN}, MoCo v3 with \textit{MC}, supervised pretraining with \textit{SL}, and no pretraining with \textit{NA-NA}. Additionally, we refer to anatomical landmark recognition as \textit{anat}, pathological finding characterisation as \textit{path}, polyp detection as \textit{det}, polyp segmentation with Kvasir-SEG as \textit{segk}, polyp segmentation with CVC-ClinicDB as \textit{segc}, and monocular depth estimation in colonoscopy as \textit{dep}.}
\label{fig:vt_vs_rn}
\end{figure*}

\begin{figure*}[ht]
\centering
\includegraphics[width=400pt]{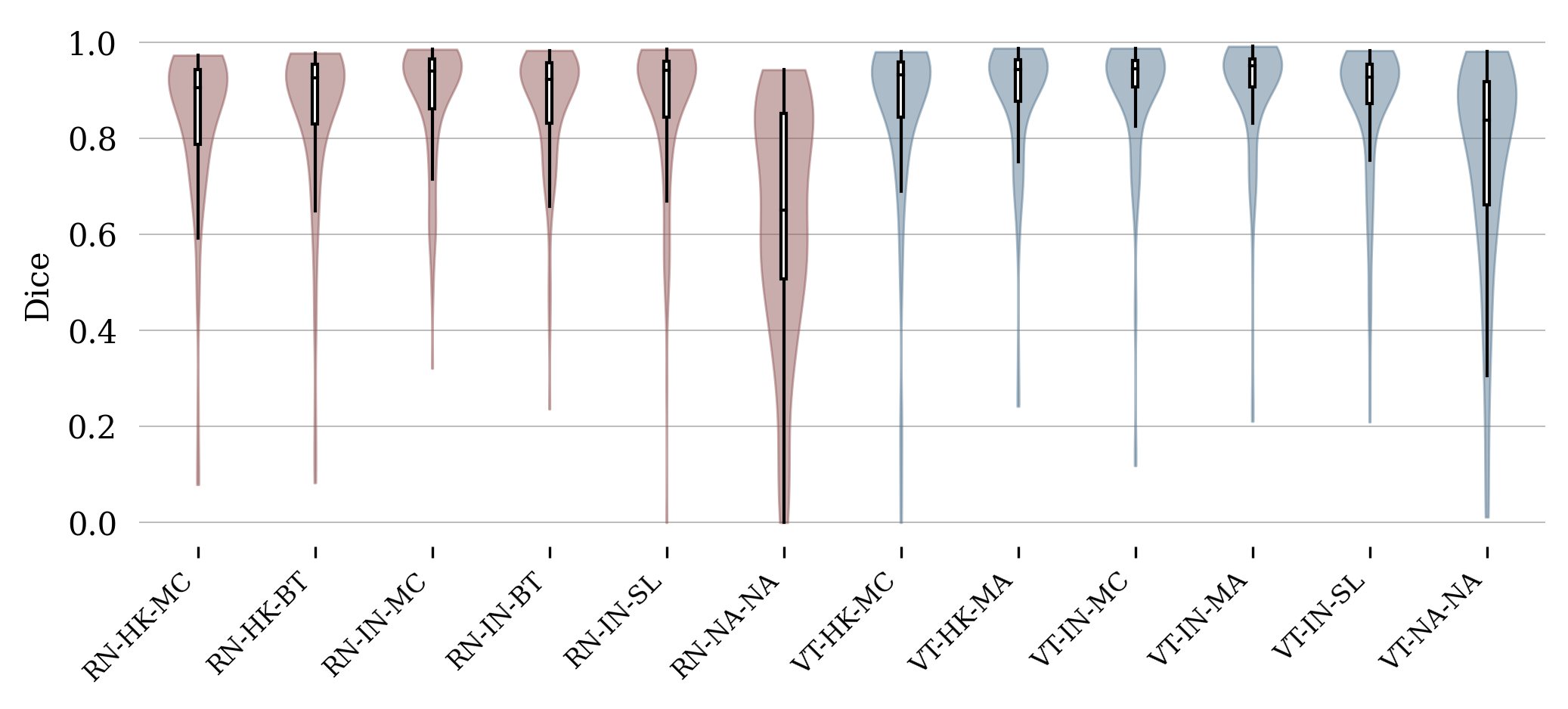}
\caption{Distribution of Dice score (higher is better) across the test set for each Kvasir-SEG polyp segmentation model, visualised as box and violin plots. For conciseness, we denote ResNet50s with \textit{RN}, ViT-Bs with \textit{VT}, Hyperkvasir-unlabelled with \textit{HK}, ImageNet-1k with \textit{IN}, MoCo v3 with \textit{MC}, Barlow Twins with \textit{BT}, MAE with \textit{MA}, supervised pretraining with \textit{SL}, and no pretraining with \textit{NA-NA}. For clarity, the violin plots for ResNet50 models are coloured red and the violin plots for ViT-B models are coloured blue.}
\label{fig:kvasir_dist}
\end{figure*}

\begin{figure*}[ht]
\centering
\includegraphics[width=400pt]{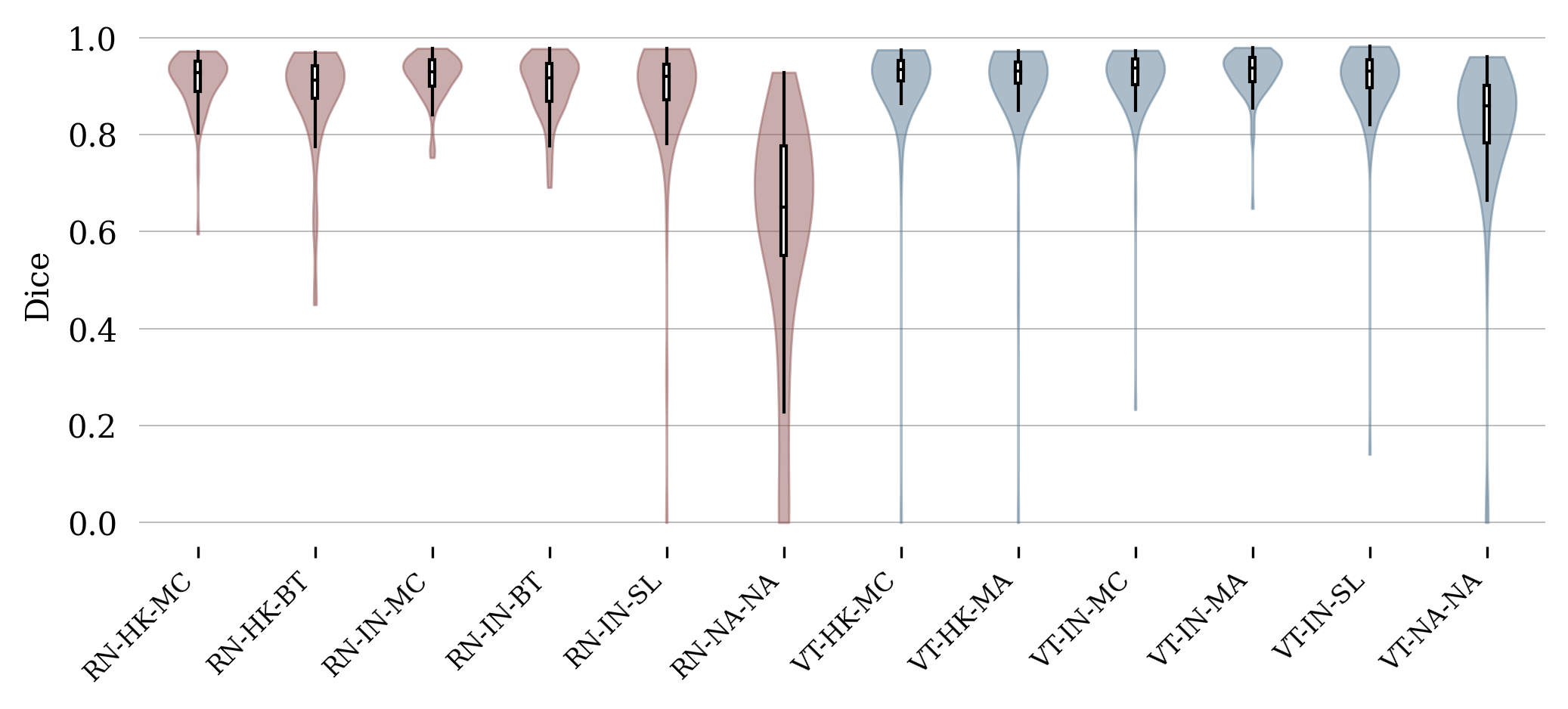}
\caption{Distribution of Dice score (higher is better) across the test set for each CVC-ClinicDB polyp segmentation model, visualised as box and violin plots. For conciseness, we denote ResNet50s with \textit{RN}, ViT-Bs with \textit{VT}, Hyperkvasir-unlabelled with \textit{HK}, ImageNet-1k with \textit{IN}, MoCo v3 with \textit{MC}, Barlow Twins with \textit{BT}, MAE with \textit{MA}, supervised pretraining with \textit{SL}, and no pretraining with \textit{NA-NA}. For clarity, the violin plots for ResNet50 models are coloured red and the violin plots for ViT-B models are coloured blue.}
\label{fig:cvc_dist}
\end{figure*}

\begin{figure*}[ht]
\centering
\includegraphics[width=400pt]{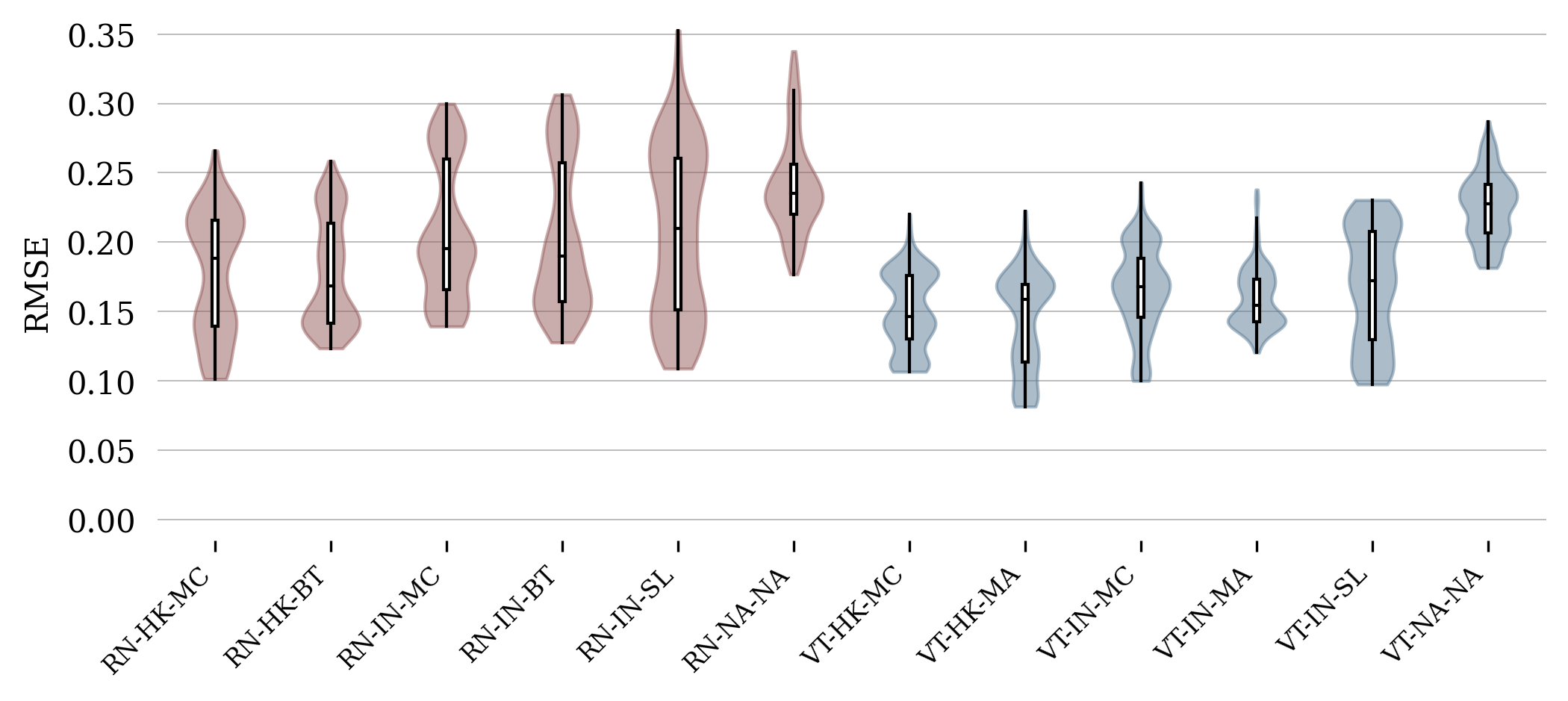}
\caption{Distribution of RMSE (lower is better) across the test set for each C3VD monocular depth estimation model, visualised as box and violin plots. For conciseness, we denote ResNet50s with \textit{RN}, ViT-Bs with \textit{VT}, Hyperkvasir-unlabelled with \textit{HK}, ImageNet-1k with \textit{IN}, MoCo v3 with \textit{MC}, Barlow Twins with \textit{BT}, MAE with \textit{MA}, supervised pretraining with \textit{SL}, and no pretraining with \textit{NA-NA}. For clarity, the violin plots for ResNet50 models are coloured red and the violin plots for ViT-B models are coloured blue.}
\label{fig:c3vd_dist}
\end{figure*}

Note that this analysis omits any results for pretraining with Barlow Twins or MAE. We visualise the results of this analysis in Fig. \ref{fig:vt_vs_rn}, where it can be seen that the ResNet50 and ViT-B models perform similarly in anatomical landmark recognition and pathological finding characterisation, that the ResNet50 models perform better than the ViT-B models perform in polyp detection, and that the ViT-B models generally perform better in the dense prediction tasks of polyp segmentation and monocular depth estimation colonoscopy. We further demonstrate the advantage of the ViT-B models over the ResNet50 models in dense prediction by visualising the distribution of performance across the Kvasir-SEG, CVC-ClinicDB, and C3VD test sets in Fig. \ref{fig:kvasir_dist}, Fig. \ref{fig:cvc_dist}, and Fig. \ref{fig:c3vd_dist}, respectively. Such visualisations are only suitable for these experiments since the metrics measure the performance on each instance prior to averaging, which is not the case for our image classification or object detection experiments. While we observe that ResNet50 models are typically better on polyp detection, we note that the polyp detection model with an MAE pretrained backbone with ImageNet-1k performs better than all but two models with ResNet50 backbones with respect to AP, and performs best with respect to AP$_{50}$, further emphasising the particular robustness of this pretraining pipeline. There is still much to understand about the relative strengths and weaknesses of these architectures, particularly in the context of domains where the availability of data is much lower than that of everyday images, such as GIE. However, these results provide useful insights into which architecture may be better suited to each considered task.

One final note we make is that, as expected, pretraining with any of the considered pipelines consistently leads to better fine-tuned performance than training on the downstream task from random initialisation.

\section{Conclusion}
In this work, we studied the pretraining of image encoders for use as backbones in solutions to vision tasks in GIE, considering variation in encoder architecture, pretraining pipeline (data and algorithm), and downstream task. This was motivated by recent opportunities to improve on the convention of supervised pretraining backbones on image classification with ImageNet-1k, namely modern self-supervised pretraining algorithms and Hyperkvasir-unlabelled --- a relatively large dataset of unlabelled GIE images. We primarily identified the best pretraining pipeline and architecture, out of those considered, for each considered task by adapting the encoders to the tasks with state-of-the-art decoders, fine-tuning the resulting models on datasets that include suitable annotations for the tasks, and evaluating the performance on test sets with well-established metrics. Overall, we found that a ViT-B backbone pretrained using the MAE algorithm and ImageNet-1k was most robust. Additionally, our findings suggest three general principles regarding the pretraining of encoders for use as backbones in solutions to vision tasks in GIE, which we revealed through an analysis of the downstream performance. These include:

\begin{itemize}
    \item Self-supervised pretraining generally produces more suitable backbones than supervised pretraining. This result is significant as it is still the convention to use backbones that have been pretrained on ImageNet-1k in a supervised manner --- this implies that the current state-of-the-art could be improved upon through self-supervised pretraining. Additionally, this result contrasts with the results observed for tasks involving everyday images, where supervised pretraining typically leads to better performance.
    \item Self-supervised pretraining with ImageNet-1k generally produces more suitable backbones than self-supervised pretraining with Hyperkvasir-unlabelled, with the notable exception of monocular depth estimation in colonoscopy where the similarity of the pretraining data to the downstream data appears to be more critical than the amount of pretraining data. While this is a useful insight for the development of monocular depth estimation models for GIE, this finding may also be true for monocular depth estimation solutions in other domains.
    \item That ResNet50 backbones are generally better for polyp detection, whereas ViT-B backbones are generally better for polyp segmentation and monocular depth estimation in colonoscopy, and both architectures perform similarly in anatomical landmark recognition and pathological finding characterisation.
\end{itemize}

We hope that this paper encourages further work on the topic of pretraining image encoders for use as backbones in solutions to vision tasks in GIE. Firstly, the scope of this work could be extended to more tasks and datasets, as well as decoder architectures and fine-tuning procedures. For example, we considered the Faster R-CNN object detection pipeline, which is a 2-stage detector, and it is worth investigating whether our findings are also true for 1-stage detectors. Additionally, we considered supervised fine-tuning for monocular depth estimation in colonoscopy, while self-supervised fine-tuning for monocular depth estimation is also a promising research avenue and may benefit from an investigation into pretraining. Also, the impact of existing pretraining pipelines on the hybrid architectures that combine both convolutional and transformer components and that have found success in polyp segmentation can be investigated. We believe that such research should lay the groundwork for the development of backbones that are better suited to tasks in GIE, which should allow for significant advancement in the state-of-the-art. Beyond extending the scope of this study and the further investigation of existing pretraining algorithms, we suggest that future work also studies the development of pretraining algorithms specifically for this domain, as well as for other encoder architectures.

\section*{Acknowledgment}
Data Access Statement: this publication is supported by multiple datasets which are openly available as cited in the 'References' section of this paper. A persistent record of the software developed as part of the reported research is openly available from https://doi.org/10.17030/uclan.data.00000447.

\bibliographystyle{unsrt}
\bibliography{sample}

\begin{IEEEbiography}[{\includegraphics[width=1in,height=1.25in,clip,keepaspectratio]{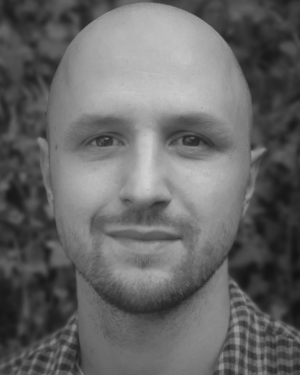}}]{Edward Sanderson} received the Ph.D. degree in data science from the University of Central Lancashire in 2022, Preston, U.K.. He has since worked as a Postdoctoral Researcher in the Computer Vision and Machine Learning (CVML) Group, University of Central Lancashire, on the Science and Technology Facilities Council (STFC) CDN+ funded ‘‘Machine Learning System for Decision Support and Computational Automation of Early Cancer Detection and Categorization in Colonoscopy (AIdDeCo)'' project. He has won several awards for his research, including first place in the Depth Estimation task of the SimCol3D EndoVis Challenge at MICCAI 2022, second place in the Poster Competition at the STFC CDN+ Celebration Event 2023, and third place in the Best Paper Award at MIUA 2022. His research interests include machine learning, computer vision, and medical image processing.
\end{IEEEbiography}

\begin{IEEEbiography}[{\includegraphics[width=1in,height=1.25in,clip,keepaspectratio]{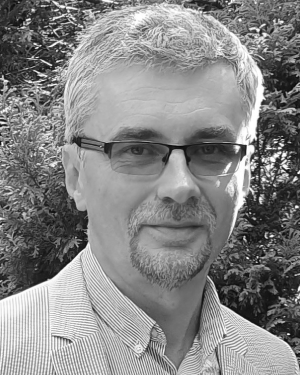}}]{BOGDAN J. MATUSZEWSKI} (Member, IEEE) received the Ph.D. degree in electronic engineering from the Wrocław University of Science and Technology, Poland, in 1996. He is currently a Professor in computer vision, the Deputy Director of the Research Centre in Engineering, and the Head of the Computer Vision and Machine Learning (CVML) Group, University of Central Lancashire, Preston, U.K. He has participated in 24 research projects, leading 11 of them, with funding secured from the U.K. Research Councils, EU, and industry. Most recently, he is also the Principal Investigator of the Science and Technology Facilities Council CDN+ funded ‘‘Machine Learning System for Decision Support and Computational Automation of Early Cancer Detection and Categorization in Colonoscopy (AIdDeCo)’’ Project. He has published over 150 research papers, won several awards for his research, and supervised 18 Ph.D.’s to successful completion. His research interests include computer vision, data science, and AI (machine learning) within areas of imaging, healthcare technologies, digital engineering, deformation modeling, segmentation, registration, object characterization, and 3D scene reconstruction and understanding
\end{IEEEbiography}

\EOD

\end{document}